\pdfoutput=1
\RequirePackage{fix-cm}
\documentclass[natbib, extended, 16pt]{svjour3}       
\smartqed  
\usepackage{graphicx}

\usepackage{mathptmx}      

\usepackage{amsmath, amssymb}
\usepackage{bm}
\usepackage{mathrsfs}
\usepackage{graphicx, fancyhdr}
\usepackage{placeins}  
\usepackage{floatrow}

\usepackage[intoc]{nomencl}

\usepackage{hyperref}

\let\origfontsize\fontsize
\def\fontsize#1#2{\origfontsize{12}{14.5}}

\hypersetup{
 colorlinks=true,
 linkcolor=blue,
 citecolor=blue,
 urlcolor=blue
 }

\usepackage{epstopdf}
\usepackage{color}
\usepackage[boxruled,vlined,linesnumbered]{algorithm2e}
\usepackage{ifpdf}
\usepackage{multirow}
\usepackage{caption}
\usepackage{subcaption}

\newdimen\figrasterwd
\figrasterwd\textwidth

\usepackage{setspace}
\usepackage{xcolor}

\usepackage[margin=1in]{geometry}

\newlength\myindent
\journalname{my journal}
\begin{document}
\title{A Deep-Learning-Based Geological Parameterization for History Matching Complex Models
}

\titlerunning{Deep-learning-based parameterization for history matching}        

\author{Yimin Liu, Wenyue Sun, Louis J.~Durlofsky}


\institute{Y. Liu \at
              Department of Energy Resources Engineering, Stanford University, \\
              Stanford, CA 94305-2220, USA \\
              \email{yiminliu@stanford.edu}           
           \and
           W. Sun \at
              Department of Energy Resources Engineering, Stanford University, \\
              Stanford, CA 94305-2220, USA \\
              \email{wenyue@stanford.edu}           
           \and
           L.J. Durlofsky \at
              Department of Energy Resources Engineering, Stanford University, \\
              Stanford, CA 94305-2220, USA \\
              \email{lou@stanford.edu}           
}

\date{Received: date / Accepted: date}
\maketitle

\begin{abstract}

A new low-dimensional parameterization based on principal component analysis (PCA) and convolutional neural networks (CNN) is developed to represent complex geological models. The CNN-PCA method is inspired by recent developments in computer vision using deep learning. CNN-PCA can be viewed as a generalization of an existing optimization-based PCA (O-PCA) method. Both CNN-PCA and O-PCA entail post-processing a PCA model to better honor complex geological features. In CNN-PCA, rather than use a histogram-based regularization as in O-PCA, a new regularization involving a set of metrics for multipoint statistics is introduced. The metrics are based on summary statistics of the nonlinear filter responses of geological models to a pre-trained deep CNN. In addition, in the CNN-PCA formulation presented here, a convolutional neural network is trained as an explicit transform function that can post-process PCA models quickly. CNN-PCA is shown to provide both unconditional and conditional realizations that honor the geological features present in reference SGeMS geostatistical realizations for a binary channelized system. Flow statistics obtained through simulation of random CNN-PCA models closely match results for random SGeMS models for a demanding case in which O-PCA models lead to significant discrepancies. Results for history matching are also presented. In this assessment CNN-PCA is applied with derivative-free optimization, and a subspace randomized maximum likelihood method is used to provide multiple posterior models. Data assimilation and significant uncertainty reduction are achieved for existing wells, and physically reasonable predictions are also obtained for new wells. Finally, the CNN-PCA method is extended to a more complex non-stationary bimodal deltaic fan system, and is shown to provide high-quality realizations for this challenging example.

\keywords{Geological Parameterization \and History Matching \and Deep Learning \and Principal Component Analysis}
\end{abstract}

\newcommand{\Ud}{\mathrm{d}} 
\newcommand{\Bx}{\mathbf{x}}
\newcommand{\By}{\mathbf{y}} 
\newcommand{\Bh}{\mathbf{h}} 
\newcommand{\Ba}{\mathbf{a}}
\newcommand{\Bb}{\mathbf{b}}
\newcommand{\Bxstar}{\mathbf{x}^{*}}
\newcommand{\Bz}{\mathbf{z}}
\newcommand{\Bxi}{\boldsymbol{\xi}}
\newcommand{\BXi}{\boldsymbol{\Xi}}
\newcommand{\Bximap}{\boldsymbol{\xi}_\text{map}}
\newcommand{\Bxirml}{\boldsymbol{\xi}_\text{rml}}
\newcommand{\Bxiuc}{\boldsymbol{\xi}^*}
\newcommand{\Bm}{\mathbf{m}}
\newcommand{\Bw}{\mathbf{w}}
\newcommand{\BM}{\mathbf{M}}
\newcommand{\Bg}{\mathbf{g}}
\newcommand{\Bmref}{\Bm_{\text{ref}}}

\newcommand{\Bmbar}{\bar{\mathbf{m}}}
\newcommand{\Bmuc}{\mathbf{m}_\text{uc}}
\newcommand{\Bmpca}{\mathbf{m}_\text{pca}}
\newcommand{\Bmopca}{\mathbf{m}_\text{opca}}
\newcommand{\Bmcnnpca}{\mathbf{m}_\text{cnn}}
\newcommand{\Bu}{\mathbf{u}}
\newcommand{\argmin}[1]{\underset{#1}{\text{argmin}}}
\newcommand{\erf}[1]{\text{erf}(#1)}
\newcommand{\Bd}{\mathbf{d}} 
\newcommand{\Bdobs}{\mathbf{d}_{\text{obs}}}
\newcommand{\Cd}{C_{\text{d}}}
\newcommand{\Cdinv}{C^{-1}_{\text{d}}}
\newcommand{\Cm}{C_{\text{m}}}
\newcommand{\Cminv}{C^{-1}_{\text{m}}}
\newcommand{\Nr}{N_{\text{r}}}
\newcommand{\Nx}{N_x}
\newcommand{\Ny}{N_y}
\newcommand{\Nb}{N_{\text{b}}}
\newcommand{\Nc}{N_{\text{c}}}
\newcommand{\Nt}{N_{\text{t}}}
\newcommand{\Nh}{N_{\text{h}}}
\newcommand{\Nep}{N_\text{ep}}
\newcommand{\Nz}[1]{N_{z,#1}}
\newcommand{\Nct}{N_{\text{c}}^{\text{t}}}
\newcommand{\td}{\text{d}}
\newcommand{\Bnu}{\boldsymbol{\nu}}
\newcommand{\COtwo}{\text{CO}_2}
\newcommand{\Smap}{S_\text{map}}
\newcommand{\Srml}{S_\text{rml}}
\newcommand{\Sd}{S_\text{d}}
\newcommand{\Sm}{S_\text{m}}
\newcommand{\lr}{l_\text{r}}

\newcommand{\Swir}{S_\text{wir}}
\newcommand{\Sor}{S_\text{or}}
\newcommand{\nw}{{n_\text{w}}}
\newcommand{\no}{{n_\text{o}}}
\newcommand{\Bmkr}{\mathbf{m}_\text{kr}}
\newcommand{\Bmkrbar}{\bar{\mathbf{m}}_\text{kr}}
\newcommand{\Bmkruc}{\mathbf{m}_\text{kr,uc}}
\newcommand{\Ckr}{C_\text{kr}}
\newcommand{\Ckrinv}{C_\text{kr}^{-1}}
\newcommand{\cmpd}{\text{m}^3/\text{day}}
\newcommand{\StdGauss}{N(\boldsymbol{0},I)}
\newcommand{\Nrml}{N_\text{rml}}
\newcommand{\Niter}{N_\text{iter}}
\newcommand{\Nd}{N_\text{d}}
\newcommand{\tsim}{t_\text{sim}}
\newcommand{\textapprox}{\raisebox{0.5ex}{\texttildelow}}
\newcommand{\BBR}{\mathbb{R}}

\newcommand{\mref}{M_\text{ref}}

\newcommand{\hp}{\hspace{8px}}
\newcommand{\hps}{\hspace{4px}}

\newcommand{\Lt}{L_\text{t}}
\newcommand{\Lc}{L_\text{c}}
\newcommand{\Ls}{L_\text{s}}
\newcommand{\Lh}{L_\text{h}}
\newcommand{\al}{\alpha_l}
\newcommand{\bl}{\beta_l}
\newcommand{\Nl}{N_l}
\newcommand{\Dl}{D_l}
\newcommand{\FO}{F_l[O]}
\newcommand{\FI}{F_l[I]}
\newcommand{\Fd}{F_l[\cdot]}
\newcommand{\GO}{G_l[O]}
\newcommand{\Gd}{G_l[\cdot]}
\newcommand{\GS}{G_l[S]}

\newcommand{\TI}{\text{TI}}

\section{Introduction}

History matching in the context of oil and gas reservoir engineering involves calibrating uncertain model parameters such that flow simulation results match observed data to within some tolerance. The uncertain variables most commonly considered are porosity and permeability in every grid block of the model. History matching algorithms are often most effective when combined with procedures for parameterizing these geological variables. Such parameterizations, based on, e.g., principal component analysis (PCA), serve two key purposes. Namely, they reduce the number of variables that must be determined through data assimilation, and they preserve, to varying degrees, the spatial correlation structure that exists in prior geological models.

A number of geological parameterizations have been presented and assessed. However, as discussed below, existing methods have some important limitations. In this work, a new low-dimensional parameterization technique for complex geological models, involving a recently developed deep-learning-based algorithm known as fast neural style transfer \citep{Johnson2016}, is introduced. This new technique is referred to as CNN-PCA since it combines convolutional neural network (CNN) approaches from the deep-learning domain and PCA. The CNN-PCA representation can be viewed as an extension of an existing parameterization procedure, optimization-based PCA (O-PCA) \citep{Vo2014,Vo2015}. CNN-PCA shows advantages over O-PCA for complex systems, especially when conditioning data are scarce.

Previous work on low-dimensional model parameterization has involved the direct use of PCA representations \citep{Oliver1996,Reynolds1996,Sarma2006}, discrete wavelet transform (DWT)~\citep{LuPengbo2013}, discrete cosine transform (DCT)~\citep{Jafarpour2010}, level-set methods~\citep{Chang2010,PingJing2013}, K-SVD~\citep{Tavakoli2010}, and tensor decomposition~\citep{Insuasty2017} procedures. These methods address the parameterization problem in different manners and each has advantages and limitations. For example, the DWT and DCT procedures transform model parameters based on wavelet functions or cosine functions, though they do not incorporate the prior covariance matrix of model parameters when constructing the basis. Therefore, conditioning DCT and DWT parameterizations to prior geological information is not straightforward. PCA is based on the eigen-decomposition of the covariance matrix of prior models. As such, it honors only the two-point spatial statistics of the prior models, and is thus not directly applicable to non-Gaussian systems. 

Several methods have been proposed to extend PCA for non-Gaussian systems. Pluri-PCA, developed by \cite{Chen2016}, combines PCA and truncated-pluri-Gaussian \citep{Astrakova2015} representations to provide a low-dimensional model for multi-facies systems. This procedure entails truncation of the underlying PCA models and is thus nondifferentiable. \cite{Hakim-Elahi2017} introduced a distance transformation to map a discrete facies model into a continuous distance model that can be parameterized with PCA. This technique, like the pluri-PCA procedure, is however currently limited to discrete systems. Kernel PCA (KPCA) generalizes PCA through a kernel formulation that acts to preserve some multipoint spatial statistics \citep{Sarma2008}. The optimization-based PCA (O-PCA) procedure essentially extends PCA with a post-processing step based on the single-point statistics (histogram) of the prior models \citep{Vo2014,Vo2015}. Regularized KPCA (R-KPCA) \citep{Vo2016} can be viewed as a combination of KPCA and O-PCA. \cite{Emerick2016} investigated the performance of PCA, KPCA, and O-PCA in the history matching of channelized reservoirs using an ensemble algorithm. In that study, O-PCA performed the best in terms of data assimilation and maintaining key geological features. In addition, although KPCA and R-KPCA have some theoretical advantages over O-PCA, it is not clear if these more complex methods outperform O-PCA in problems of practical interest.

However, as illustrated later in this work, O-PCA does display limitations when applied to complex geomodels with little or no conditioning data. This is because O-PCA entails a post-processing of the underlying PCA model using what is essentially a point-wise histogram transformation. The ability of O-PCA to provide models consistent with the training image depends on the quality of the underlying PCA model, which in turn depends on the complexity of the training image and the amount of hard data available. The point-wise post-processing step does not substantially improve the ability of O-PCA to capture multipoint statistics. The CNN-PCA formulation introduced here employs features for capturing multipoint spatial correlations, which provides enhancement relative to O-PCA for non-Gaussian systems.

Recent successes in the application of deep learning for image processing have inspired the development of geological parameterization techniques based on algorithms that use deep neural networks. Among the many deep-learning procedures, the so-called generative models, which are used for constructing new images consistent with a set of training images, are of particular interest. \cite{Mosser2017} applied a type of generative adversarial network (GAN) for generating binary porous media models for pore-scale flow simulation. The use of deep neural networks for lower-dimensional geological parameterization has been proposed by \cite{Laloy2017b,Laloy2017}. They applied generative deep neural network models, based on a variational autoencoder (VAE) and a spatial generative adversarial network (SGAN), to obtain lower-dimensional representations of binary facies models. Both procedures have the ability to sample a lower-dimensional variable from a simple distribution to obtain models consistent with the training image. The quality of the new models (in both 2D and 3D) was shown to be superior to those based on existing algorithms, including PCA, O-PCA, and DCT. The two methods have also been successfully applied for data assimilation for flow simulation problems. In other work, \cite{Canchumuni2017} developed an autoencoder-based parameterization for representing facies models. This approached achieved comparable results to those from a method that combined truncated Gaussian simulation and PCA. The methods noted above are innovative in their use of recent developments in deep learning. They still have limitations, however, including substantial computational demands for training the neural networks and imperfect conditioning to hard data. 

In this study, an alternative deep-learning-based geological parameterization method, inspired by the `fast neural style transfer' algorithm of \cite{Johnson2016}, is developed and applied. The new method, referred to as CNN-PCA, can be viewed as a generalized O-PCA that includes features for capturing multipoint correlations. The metrics that enter the formulation are based on statistical quantities extracted from deep convolutional neural networks. In addition, the (somewhat) computationally demanding optimization component of the method is performed in an offline pre-processing step, so the online generation of models (as required in history matching) is very fast. In addition, although not guaranteed (as in PCA and O-PCA), the CNN-PCA approach developed here honors hard data for the cases considered. Extensions to handle bimodal models (as opposed to strictly binary models) are also presented.

This paper proceeds as follows. In Sect.~\ref{sec-methodology}, the CNN-PCA procedure is presented as a generalization of O-PCA, in which convolutional neural networks are applied. Next, in Sect.~\ref{sec-model-gen}, both unconditional and conditional realizations generated using CNN-PCA are presented for a binary channelized system. In Sect.~\ref{sec-flow-stats}, the accuracy of flow statistics obtained from CNN-PCA realizations are assessed by comparing with reference geostatistical (SGeMS) realizations as well as to O-PCA models. Then, in Sect.~\ref{sec-hm}, history matching is performed to assess the ability of posterior CNN-PCA models to provide production forecasts for both existing wells and new wells. Next, in Sect.~\ref{sec-bimodal-df}, CNN-PCA is extended to model a non-stationary bimodal deltaic fan system. Concluding remarks are provided in Sect.~\ref{sec-concl}. The detailed architecture for the convolutional neural networks used in CNN-PCA is presented in the Appendix.

\section{Convolutional Neural Network-based Principal Component Analysis (CNN-PCA)}
\label{sec-methodology}

In this section the new geological parameterization, CNN-PCA, is developed. The method is presented as a generalized O-PCA procedure coupled with new treatments based on deep convolutional neural networks.

\subsection{Optimization-based Principal Component Analysis (O-PCA)}
\label{subsec-opca}
The uncertain rock properties that characterize a reservoir model, such as permeability and porosity, are often represented as random fields. In this paper the geological model is denoted by $\Bm \in \BBR^{\Nc\times 1}$, where $\Nc$ is the number of grid blocks (computational cells). The spatial correlation structure to be captured in $\Bm$ is defined by the geological features in the reservoir. 

The basic idea of PCA, also referred to as Karhunen-Lo\`eve expansion, is to represent the random field $\Bm$ using a set of variables $\Bxi$ in a lower-dimensional subspace spanned by orthogonal basis components. PCA is highly efficient for dimension reduction and, as discussed below, enables the generation of new realizations of $\Bm$ by sampling $\Bxi$ from a standard normal distribution. Here a brief description of the PCA procedure is provided. Refer to \cite{Sarma2008} and \cite{Vo2014} for further details.  

The first step in PCA is to generate a set of $\Nr$ realizations of $\Bm$ using a geostatistical toolbox such as SGeMS~\citep{Remy2009}. Note that throughout this paper, realizations generated using geostatistical simulation algorithms within SGeMS will be referred to as `SGeMS realizations' or `SGeMS models.' Next the SGeMS realizations are assembled into a centered data matrix 
\begin{equation}
Y=\dfrac{1}{\sqrt{\Nr-1}}[\Bm_1 - \Bmbar \hp \Bm_2 - \Bmbar \hp ...\hp \Bm_{\Nr} - \Bmbar],
\label{eq-pca1}
\end{equation}
where $Y \in \BBR^{\Nc  \times \Nr}$, $\Bm_i\in \BBR^{\Nc\times 1}$ denotes realization $i$, and $\Bmbar\in \BBR^{\Nc\times 1}$ is the mean of all $\Nr$ realizations. Then a singular value decomposition is performed on $Y$, which gives $Y=U\Sigma V^T$, where $U \in \BBR^{\Nc \times \Nr}$ is the left singular matrix whose columns are the eigenvectors of $YY^T$, $\Sigma \in \BBR^{\Nr \times \Nr}$ is a diagonal matrix whose elements are the square root of the corresponding eigenvalues, and $V \in \BBR^{\Nr \times \Nr}$ is the right singular matrix. 

Given $U$ and $\Sigma$, new PCA realizations ($\Bm_\text{pca}$) can be generated by
\begin{equation}
\Bm_\text{pca}=U_l\Sigma _l\Bxi+\Bmbar,
\label{eq-pca3}
\end{equation}
where $U_l \in \BBR^{\Nc \times l}$ contains the $l$ columns in $U$ that are associated with the largest eigenvalues, $\Sigma _l\in \BBR^{l \times l}$ is a diagonal matrix containing the square roots of the corresponding eigenvalues in $U_l$, and $\Bxi \in \BBR^{l \times 1}$ is a vector with elements drawn independently from the standard normal distribution. In practice, the variability of the random field $\Bm$ can often be captured using relatively few PCA components (columns in $U$), which leads to $l \ll \Nc$. The choice of $l$ can be determined by applying an `energy' criterion \citep{Sarma2008} or through visual inspection.

PCA performs well for Gaussian random fields, whose spatial correlation structure can be fully characterized by two-point statistics (i.e., by the covariance matrix). However, for non-Gaussian models with spatial correlation characterized by multipoint statistics, PCA realizations constructed using Eq.~\ref{eq-pca3} do not fully honor the geological features. O-PCA was developed by \citet{Vo2014} to provide improved performance for non-Gaussian models. The O-PCA formulation for binary systems will now be described. In such systems, the value of $\Bm$ is either 0 or 1, where, e.g., 0 represents mud (shale) facies and 1 represents sand (channel) facies. The detailed formulation for O-PCA for bimodal systems is presented in \cite{Vo2015}.

For binary systems, the O-PCA model is constructed through solution of the following minimization problem:
\begin{equation}
\label{eq_opca_1}
\Bmopca(\Bxi)=\argmin{\Bx}{\big\{||U_l\Sigma _l\Bxi + \Bmbar -\Bx||_2^2+\gamma \Bx^T(\mathbf{1}-\Bx)\big\}},\hp x_i \in [0,1].
\end{equation}
Here $\Bx\in\BBR^{\Nc \times 1}$, $x_i$ represents the value in grid block $i$, and $\mathbf{1}\in\BBR^{\Nc \times 1}$ is a vector with all components equal to one. Equation~\ref{eq_opca_1} defines a minimization problem for finding the $\Bx$ vector that minimizes a loss function, consisting of two terms with a weight factor $\gamma$. The first term, $||U_l\Sigma _l\Bxi + \Bmbar -\Bx||_2^2$, represents the difference between $\Bx$ and the underlying PCA model (Eq.~\ref{eq-pca3}), which ensures that the solution resembles the PCA description. The second term, $\Bx^T(\mathbf{1}-\Bx)$, is a regularization that is a minimum (zero) when $x_i$ is either 0 or 1. It thus acts to shift the solution towards a binary distribution. The separable (block-by-block) analytical solution of Eq.~\ref{eq_opca_1} is provided by \cite{Vo2014}. As discussed there, O-PCA realizations can be viewed as a post-processing of the corresponding PCA realization through application of a block-wise histogram transformation.

\begin{figure}[htb!]
    \centering
        \floatbox[{\capbeside\thisfloatsetup{capbesideposition={left,top},capbesidewidth=7cm}}]{figure}[\FBwidth]
{\caption{Training image for the binary facies model of a 2D channelized system.}\label{fig-ti}{\hspace{1\baselineskip}}}
    {\includegraphics[width=0.5\textwidth]{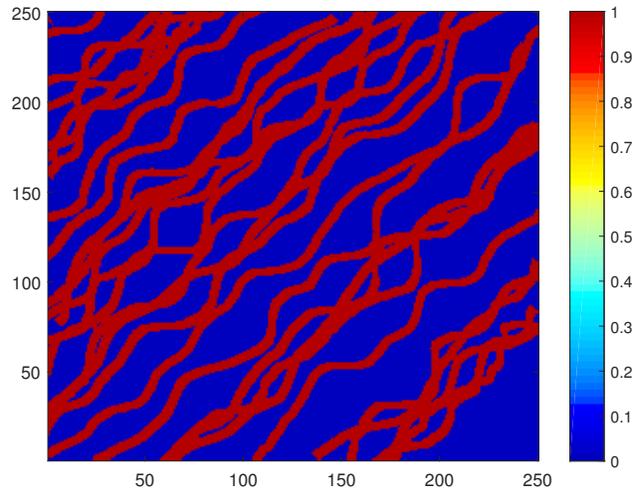}}
\end{figure}

\begin{figure}[!htb]
    \centering
    \begin{subfigure}[b]{0.32\textwidth}
        \includegraphics[width=1\textwidth]{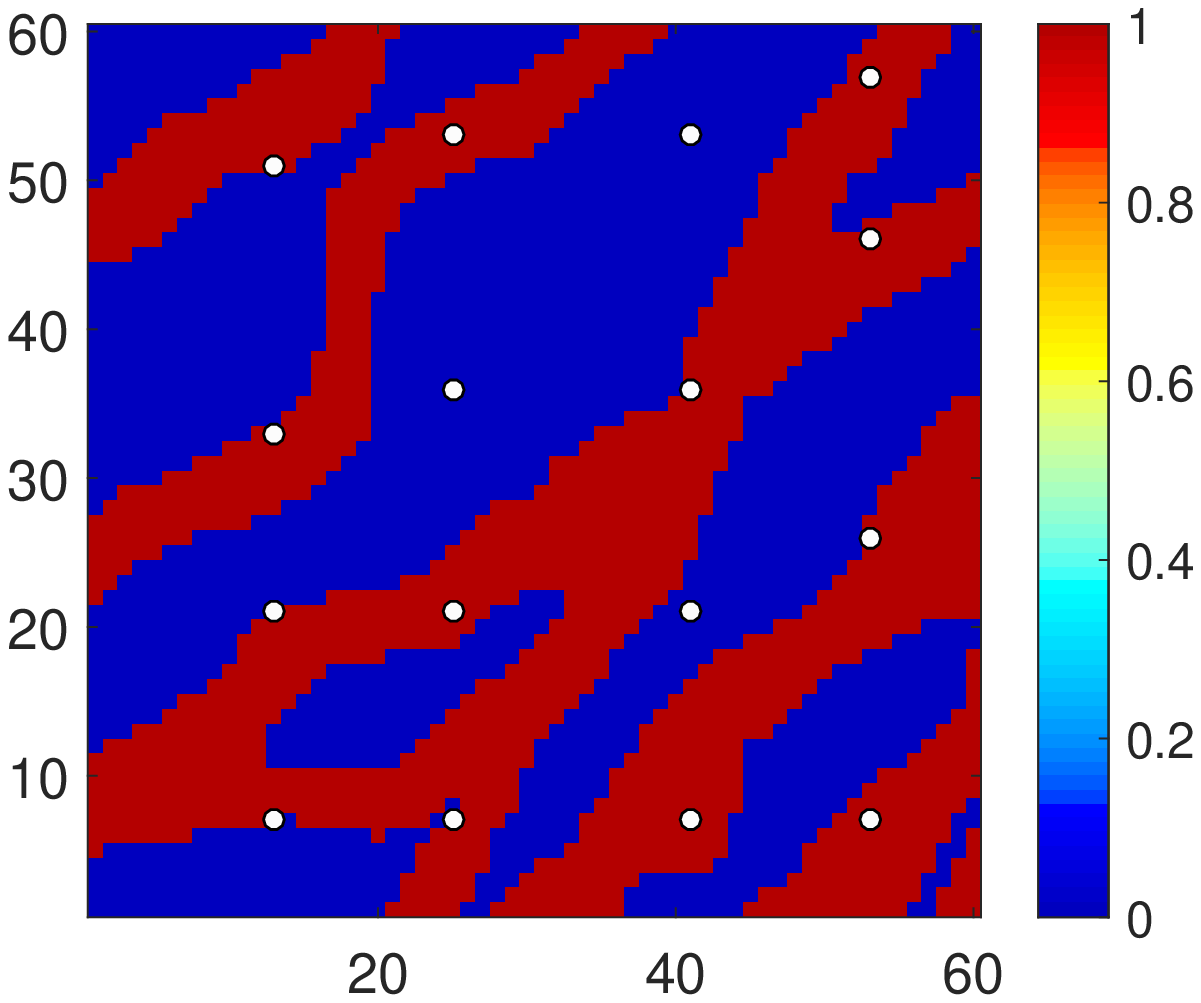}
        \caption{}
    \end{subfigure}%
    ~ 
    \hspace{0.5em}
    \begin{subfigure}[b]{0.32\textwidth}
        \includegraphics[width=1\textwidth]{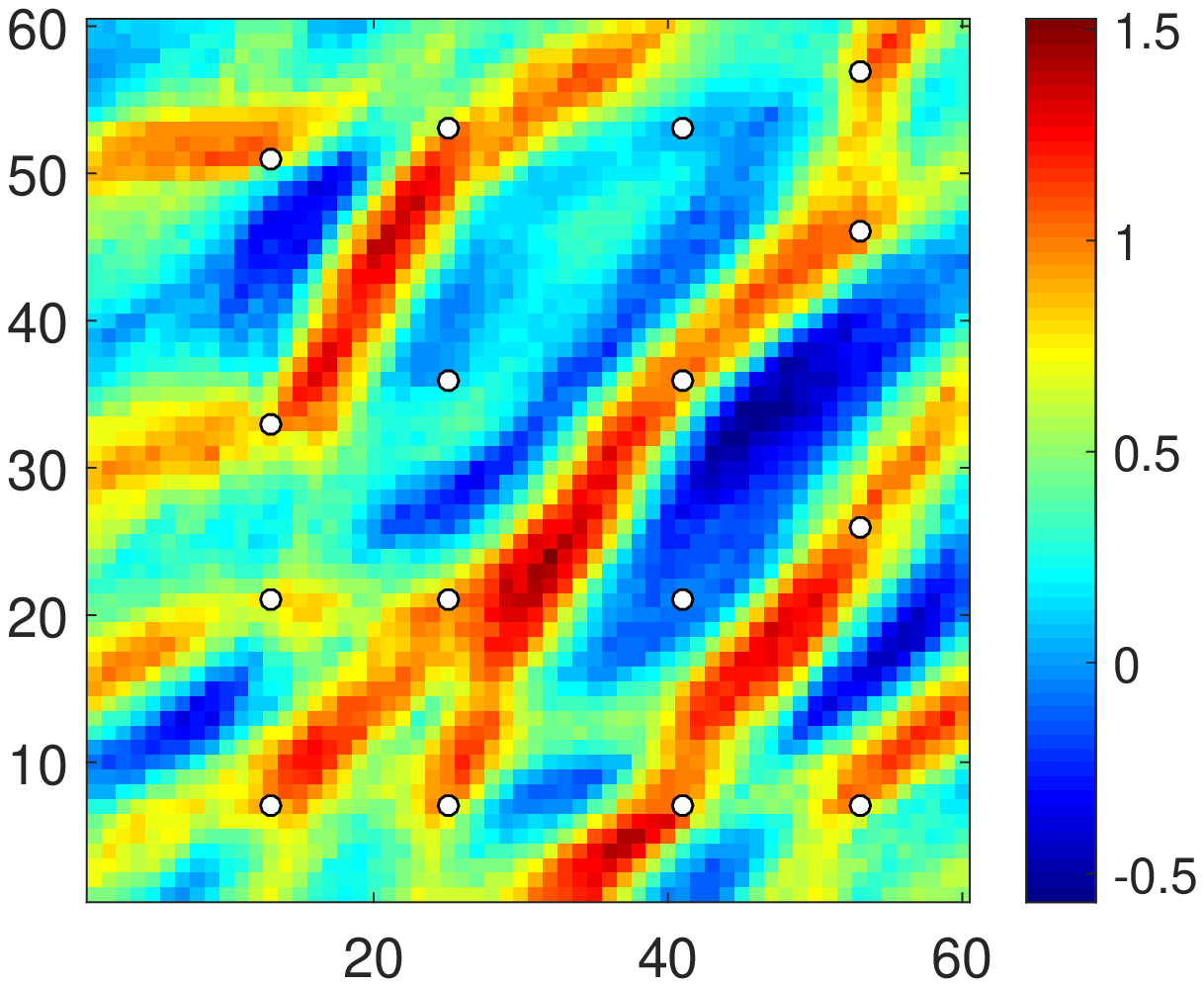}
        \caption{}
    \end{subfigure}%
    ~
    \hspace{0.5em}
    \begin{subfigure}[b]{0.32\textwidth}
        \includegraphics[width=1\textwidth]{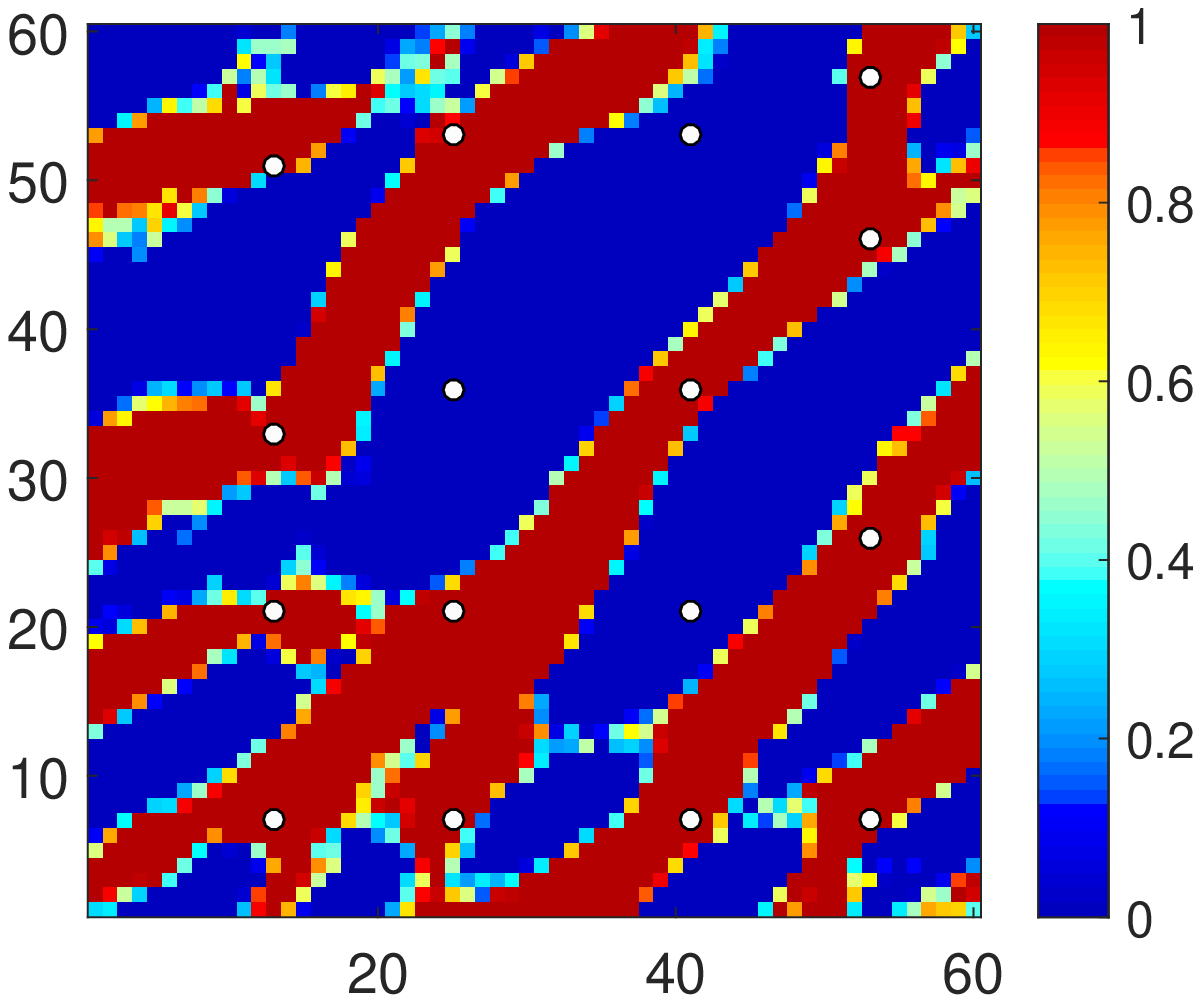}
        \caption{}
    \end{subfigure}%
    
    \begin{subfigure}[b]{0.31\textwidth}
        \includegraphics[width=1\textwidth]{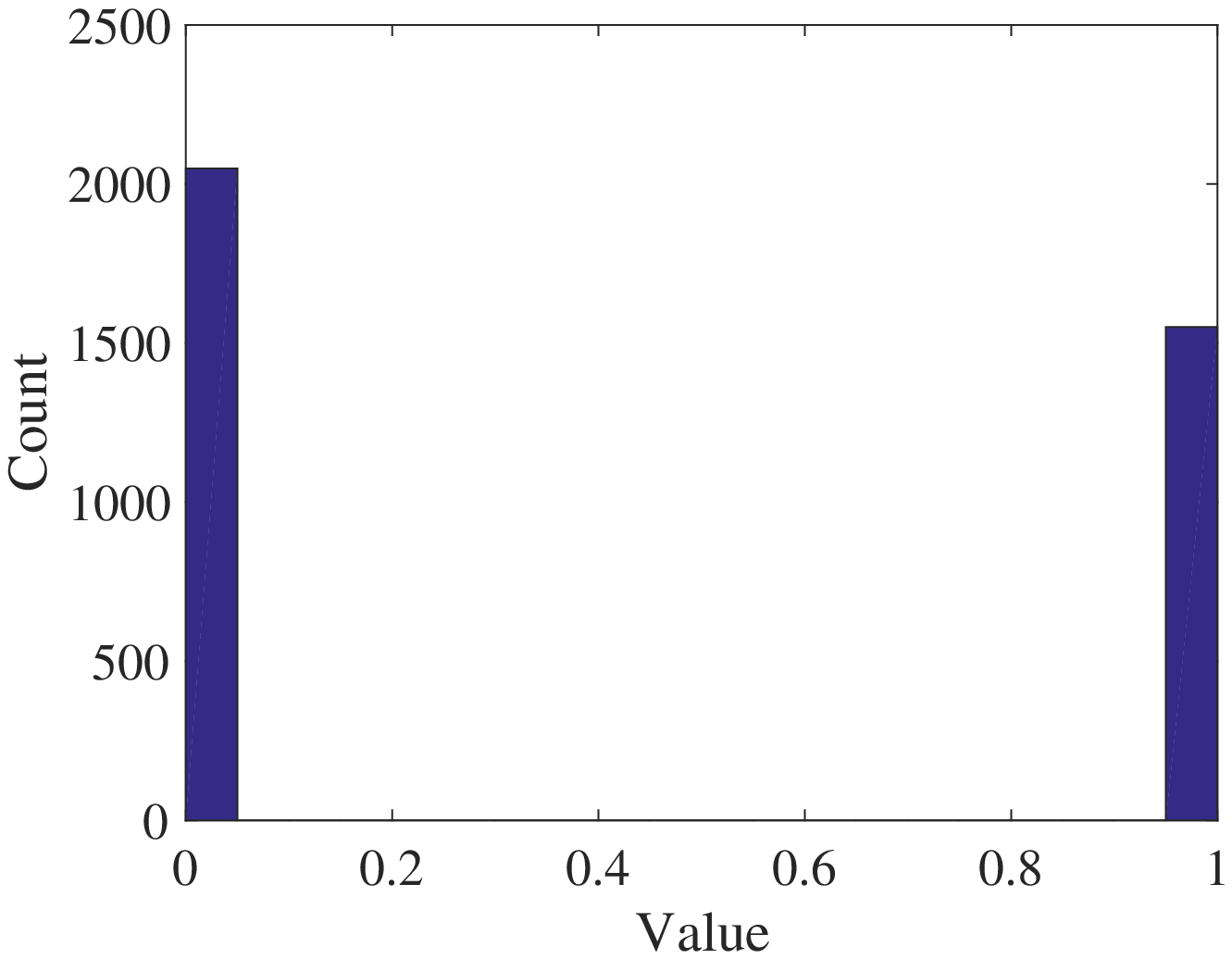}
        \caption{}
    \end{subfigure}%
    ~
    \hspace{0.5em}
    \begin{subfigure}[b]{0.31\textwidth}
        \includegraphics[width=1\textwidth]{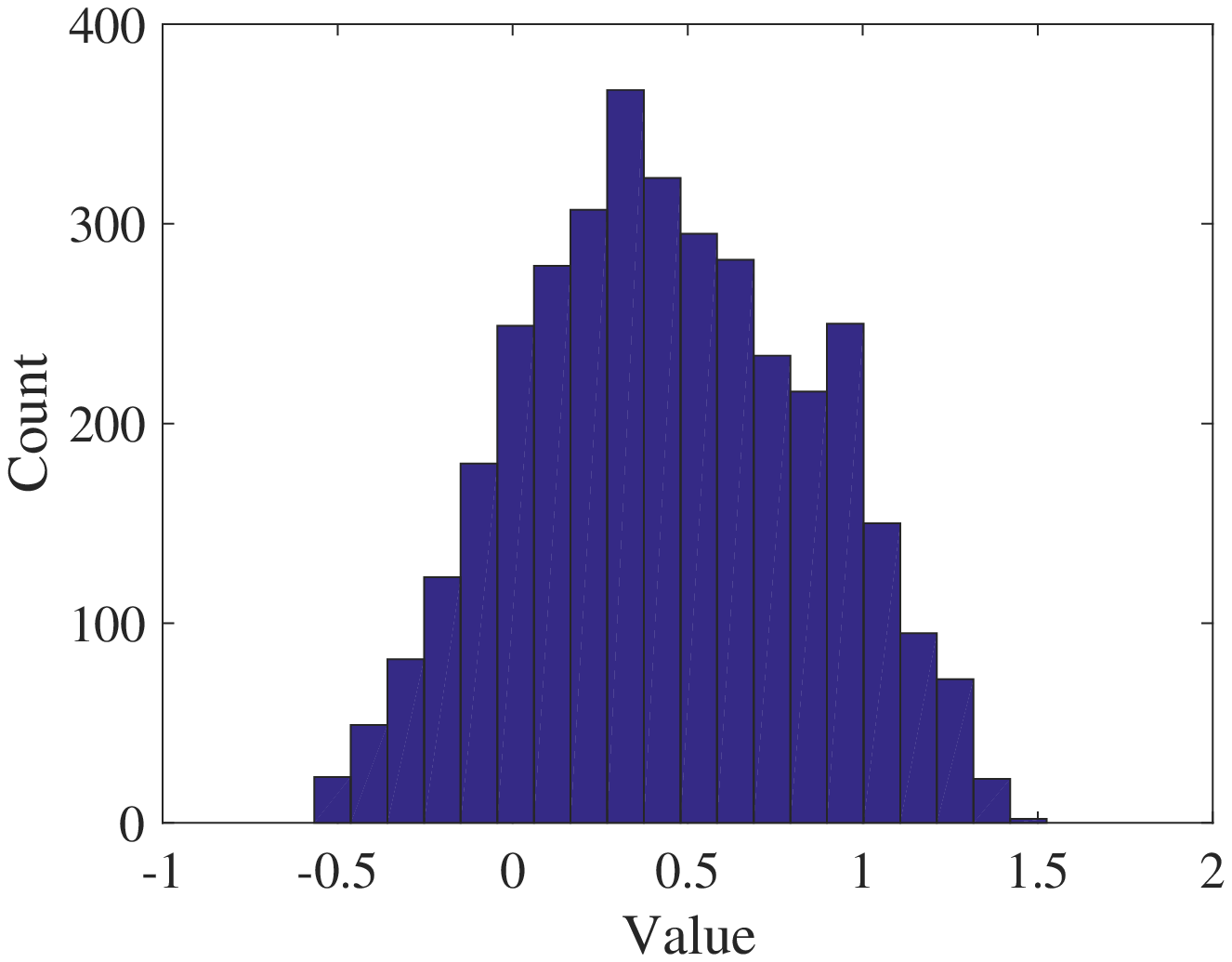}
        \caption{}
    \end{subfigure}%
    ~ 
    \hspace{0.5em}
    \begin{subfigure}[b]{0.31\textwidth}
        \includegraphics[width=1\textwidth]{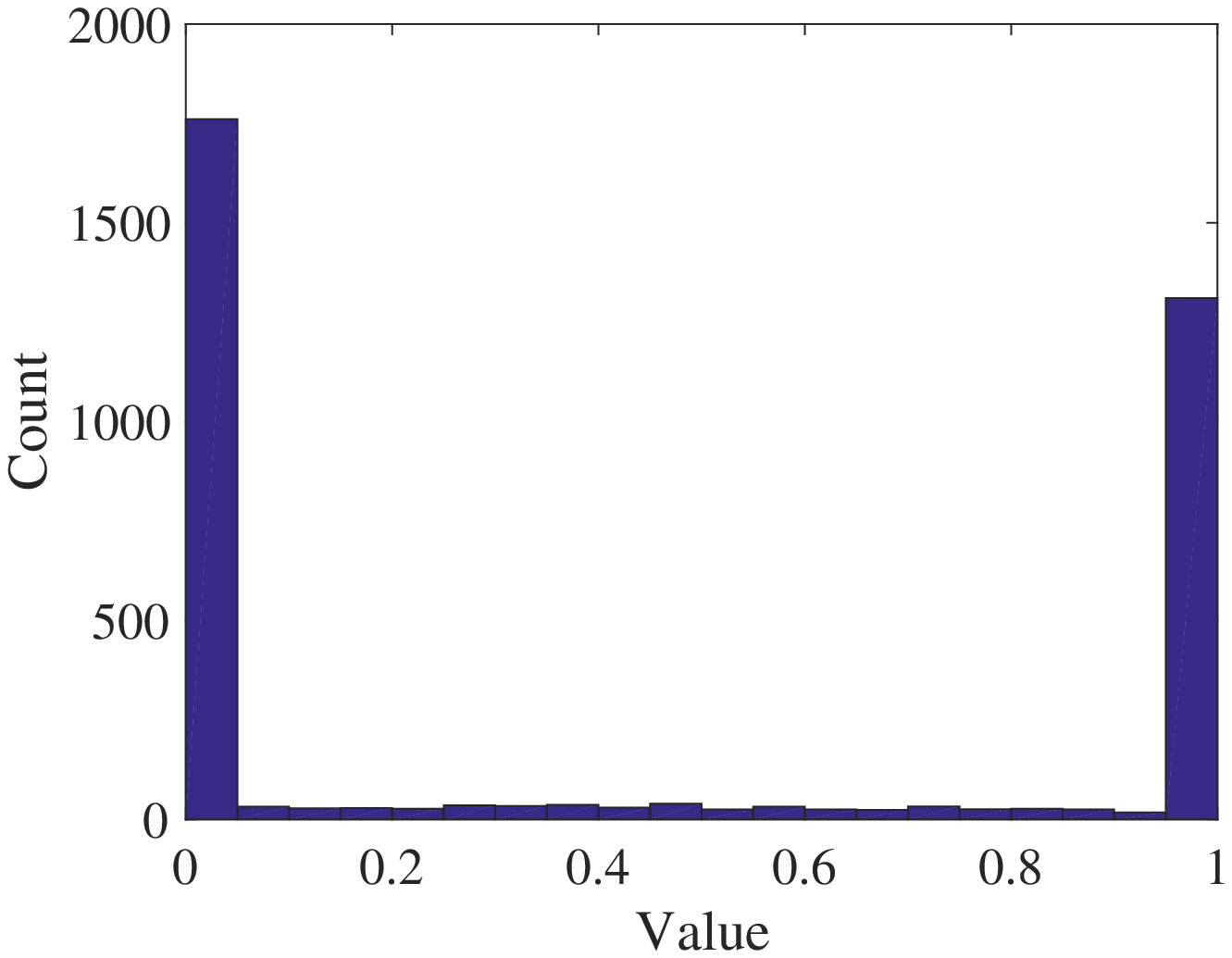}
        \caption{}
    \end{subfigure}%
    \caption{Random conditional realizations obtained from SGeMS, PCA and O-PCA for a binary channelized system with hard data at 16 wells. \textbf{a} SGeMS realization, \textbf{b} PCA realization, \textbf{c} O-PCA realization, \textbf{d} SGeMS realization histogram, \textbf{e} PCA realization histogram, \textbf{f} O-PCA realization histogram.}
    \label{fig-binary-cond}
\end{figure}

The application of PCA and O-PCA to generate new realizations of a binary facies model will now be demonstrated. The conditional channelized system, described in \cite{Vo2014}, is considered. Figure~\ref{fig-ti} displays the training image, which characterizes the geological features (spatial correlation structure) and thus the multipoint statistics. The training image is defined on a $250 \times 250$ grid.  In the first step, a total of $\Nr=1000$ conditional SGeMS realizations are generated using the `snesim' geostatistical algorithm \citep{Strebelle2002a}. The model contains $60 \times 60$ grid blocks, therefore $\Nc=3600$. Figure~\ref{fig-binary-cond}a,~b shows one SGeMS realization and the corresponding histogram of grid-block facies type. The red and blue colors in Fig.~\ref{fig-binary-cond}a correspond to sand and mud facies, respectively. The 16 white points indicate the well locations. There are three wells in mud facies and 13 wells in sand facies. All SGeMS realizations are conditioned to hard data at these well locations. The training image contains overlapping sinuous sand channels, with general orientation of about $45^\circ$. The SGeMS realization (Fig.~\ref{fig-binary-cond}a) is seen to capture these features.

The PCA and O-PCA representations are constructed using a reduced dimension $l=70$ for the PCA components (Eqs.~\ref{eq-pca3} and~\ref{eq_opca_1}). In O-PCA, the weighting factor $\gamma$ is set to 0.8. These values correspond to those used by \cite{Vo2014}. Realizations obtained from PCA and O-PCA, and the associated histograms, are shown in Fig.~\ref{fig-binary-cond}. Both the PCA and O-PCA models honor all hard data. However, the PCA model displays continuous values for the facies type and does not fully capture channel connectivity. In addition, the corresponding histogram of grid-block facies type (Fig.~\ref{fig-binary-cond}e) is essentially Gaussian, rather than near-binary. The O-PCA model, by contrast, displays a much sharper contrast between sand and mud facies, and the associated histogram is approximately binary (Fig.~\ref{fig-binary-cond}f). In addition, the O-PCA model shows channel geometry and connectivity that are reasonably consistent with that in the SGeMS realization (Fig.~\ref{fig-binary-cond}a).

\subsection{Generalized O-PCA}
\label{Subsect_limitation_opca}
In the example above, the O-PCA procedure gives satisfactory results in terms of reproducing the histogram and the geological features evident in the training image. However, as mentioned earlier, O-PCA essentially performs histogram transformation of the PCA grid-block values, which only ensures reproduction of the histogram. As will be shown later, for more challenging cases when there are fewer or no conditioning (hard) data, O-PCA realizations may not adequately capture the spatial correlation structure inherent in the training image. 

This limitation of O-PCA for non-Gaussian models is expected since Eq.~\ref{eq_opca_1} does not account for higher-order spatial correlations. The first term in the O-PCA equation, $||U_l\Sigma _l\Bxi + \Bmbar -\Bx||_2^2$, penalizes O-PCA realizations that do not closely resemble the underlying PCA realizations, and these only honor two-point correlations. The regularization term, $\Bx^T(\mathbf{1}-\Bx)$, is a metric based only on single-point statistics (histogram). This motivates the introduction of a generalized O-PCA formulation that incorporates multipoint spatial statistics.

This generalized O-PCA formulation is expressed as
\begin{equation}
\label{eq_gen_opca}
\Bm(\Bxi) = \argmin{\Bx}\big\{\Lc(\Bx, \Bmpca(\Bxi)) + \gamma_s \Ls(\Bx, \mref)\big\},
\end{equation}
where $\Bmpca(\Bxi)$ denotes a PCA realization, and $\mref$ denotes a reference model that honors the target spatial correlation structure. The objective function consists of two loss functions. The first loss function, $\Lc(\Bx, \Bmpca(\Bxi))$, referred to as the content loss, quantifies the `closeness' between model $\Bx$ and $\Bmpca(\Bxi)$ in terms of the content. The second loss function, $\Ls(\Bx, \mref)$, referred to as style loss, quantifies the degree to which $\Bx$ resembles the style of $\mref$. `Style' in this case refers to channel continuity and channel width, and the sharp contrast between sand and mud. The reference model $\mref$ can be either a training image or a particular SGeMS realization. When the spatial random field is stationary, the dimension of $\mref$ can be different than the dimension of $\Bm$. Finally, the weighting factor $\gamma_s$ is referred to as the style weight. 

The original O-PCA formulation in Eq.~\ref{eq_opca_1} can be viewed as one particular case of the generalized formulation. In this instance the content loss is based on the Euclidean distance between $\Bx$ and $\Bmpca$, and the style loss is based on the histogram. The style loss in O-PCA does not explicitly contain a reference model $\mref$, but it is essentially a metric for the mismatch between the histogram of $\Bx$ and the histogram of $\mref$.

In this study, a generalized O-PCA formulation with new content and style losses is introduced in order to generate models that better honor $\mref$. The new loss terms are inspired by the neural style transfer algorithm, developed by \cite{Gatys} within the context of computer vision. For the 2D systems considered in this study, the new content and style losses correspond closely to those in the neural style transfer algorithm.

The high-level procedure for generalized O-PCA with CNN-based loss functions is as follows. For the content loss function, let $F(\Bm)$ be an encoded representation of the model $\Bm$ such that, if two models $\Bm_1$ and $\Bm_2$ have the same encoded representation ($F(\Bm_1) = F(\Bm_2)$), then they should be perceived as having the same content. For instance, in the channelized system, if $F(\Bm_1) = F(\Bm_2)$, then $\Bm_1$ and $\Bm_2$ should have approximately the same locations for sand and mud. The content loss is defined as
\begin{equation}
\label{eq-lc-gen}
\Lc(\Bx,\Bmpca(\Bxi)) = \dfrac{1}{N_F}||F(\Bx) - F(\Bmpca(\Bxi))||_{Fr}^2,
\end{equation}
where $N_F$ is the number of elements in $F(\Bx)$ and $F(\Bmpca(\Bxi))$ and the subscript $Fr$ stands for Frobenius norm. Rather than directly using $||\Bx - \Bmpca(\Bxi)||^2_2$ as the content loss, the encoded representations are used because they are less sensitive to the block-by-block match between $\Bx$ and $\Bmpca(\Bxi)$. In addition, this treatment has been shown to generate less noisy models.

To quantify the style loss, let $\{G_k,k=1,...,K\}$ denote a set of statistical metrics that can effectively capture the spatial correlation structure of the random fields. In other words, if $\Bm$ and $\mref$ produce the same outcome for all of these statistical measurements ($G_k(\Bm)=G_k(\mref), \ \forall k=1,..,K$), then $\Bm$ and $\mref$ should be perceived as having the same spatial correlation structure. Specifically, the style loss is defined as
\begin{equation}
\label{eq-ls-gen}
\Ls(\Bx, \mref) = \sum_{k=1}^K\dfrac{1}{N_k}||G_k(\Bx) - G_k(\mref)||_{Fr}^2,
\end{equation}
where $N_k$ is the number of elements in $G_k$. Both the encoded representations $F(\Bm)$ and the statistical metrics $G_k$ are based on deep convolutional neural networks, described in more detail in Sect.~\ref{sec-cnn-metric}.

With the above content and style loss terms, the generalized O-PCA procedure is as follows. Starting with a PCA model $\Bmpca(\Bxi)$ as the initial guess, Eqs.~\ref{eq-lc-gen} and \ref{eq-ls-gen} are evaluated. The content loss for the initial guess is zero since $\Bx = \Bmpca(\Bxi)$. However, the style loss for the initial guess is likely to be large since the PCA model and the reference model $\mref$ have quite different styles. The model $\Bx$ is then updated to reduce the style loss while also limiting the content loss. Gradient-based optimization algorithms can be applied to update $\Bx$ because the gradient of the loss functions with respect to $\Bx$ can be readily obtained, as described below. The iterations continue until a stopping criterion is reached, such as a maximum number of iterations. The model corresponding to the optimal solution preserves the content indicated in the PCA model and matches the style of $\mref$.

\subsection{CNN-based Loss Functions}
\label{sec-cnn-metric}

The detailed treatments for the loss terms are now described, starting with the set of statistical metrics $\{G_k,k=1,...,K\}$ used for the style loss. The purpose of these metrics is to explicitly characterize the spatial correlation structure of a spatial random field. The covariance matrix and experimental variogram are two such metrics for Gaussian models. However, for non-Gaussian models with multipoint correlations, obtaining effective metrics for the spatial correlation is not as straightforward. Note that although there are multipoint statistics simulation algorithms that can generate realistic realizations for non-Gaussian models, these algorithms do not generally provide explicit metrics that quantify how well the spatial correlation is honored. 

Theoretically, the $N^{\rm th}$ order joint histogram of the model can fully capture the correlation of $N$ points. However, even for models of relatively small size, the computational cost to obtain accurate estimates is prohibitive. A more practical approach, proposed by \cite{Dimitrakopoulos2010}, uses high-order experimental anisotropic cumulants calculated with predefined spatial templates. This approach can be viewed as an extension of two-point variograms for multipoint statistics. However, for different geological systems, different spatial templates need to be selected to effectively capture the spatial correlation. 

\begin{figure}[!htb]
    \centering
    \begin{subfigure}[b]{0.24\textwidth}
        \includegraphics[width=1\textwidth]{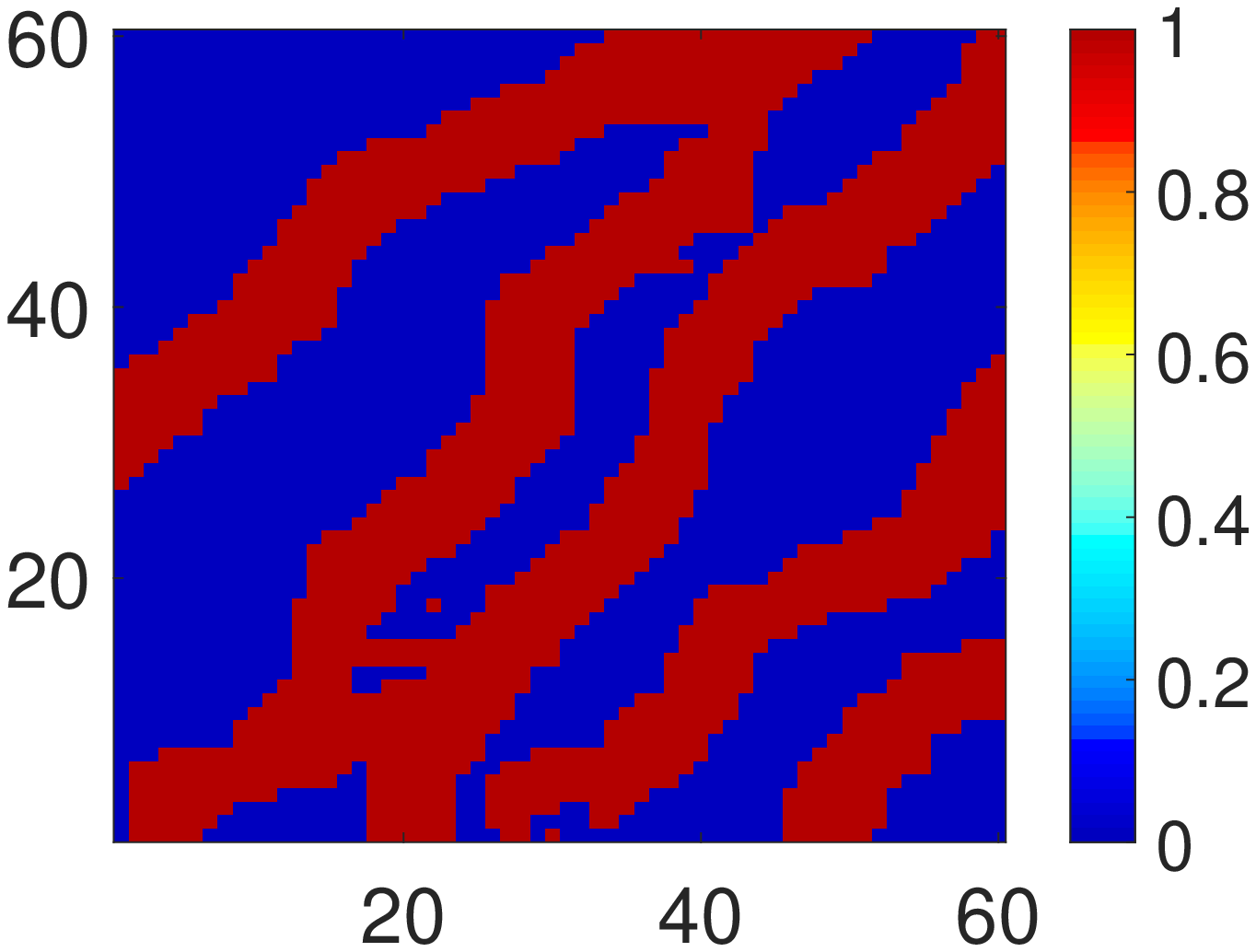}
        \caption{}
    \end{subfigure}%
    ~
    \begin{subfigure}[b]{0.24\textwidth}
        \includegraphics[width=1\textwidth]{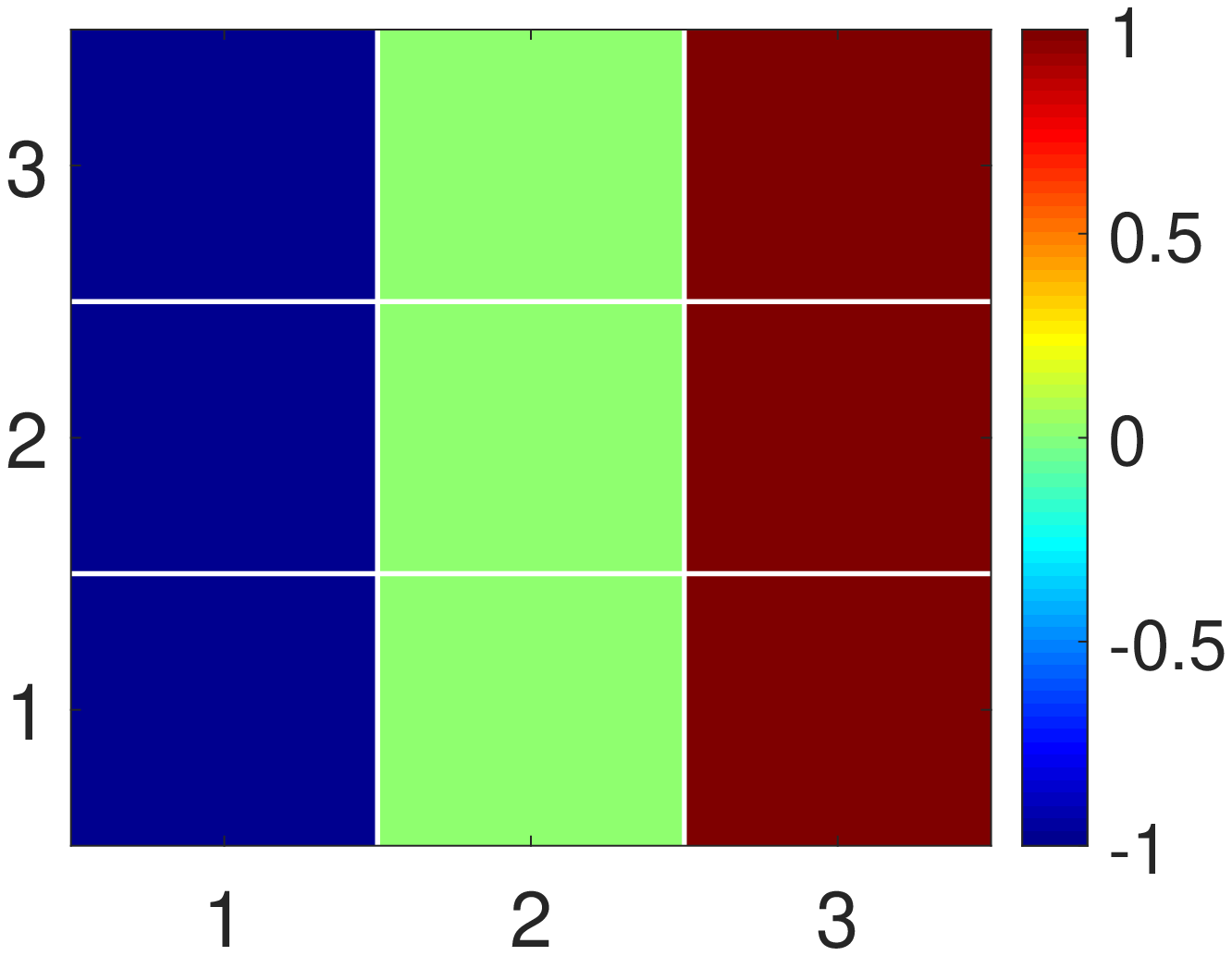}
        \caption{}
    \end{subfigure}%
    ~
    \begin{subfigure}[b]{0.24\textwidth}
        \includegraphics[width=1\textwidth]{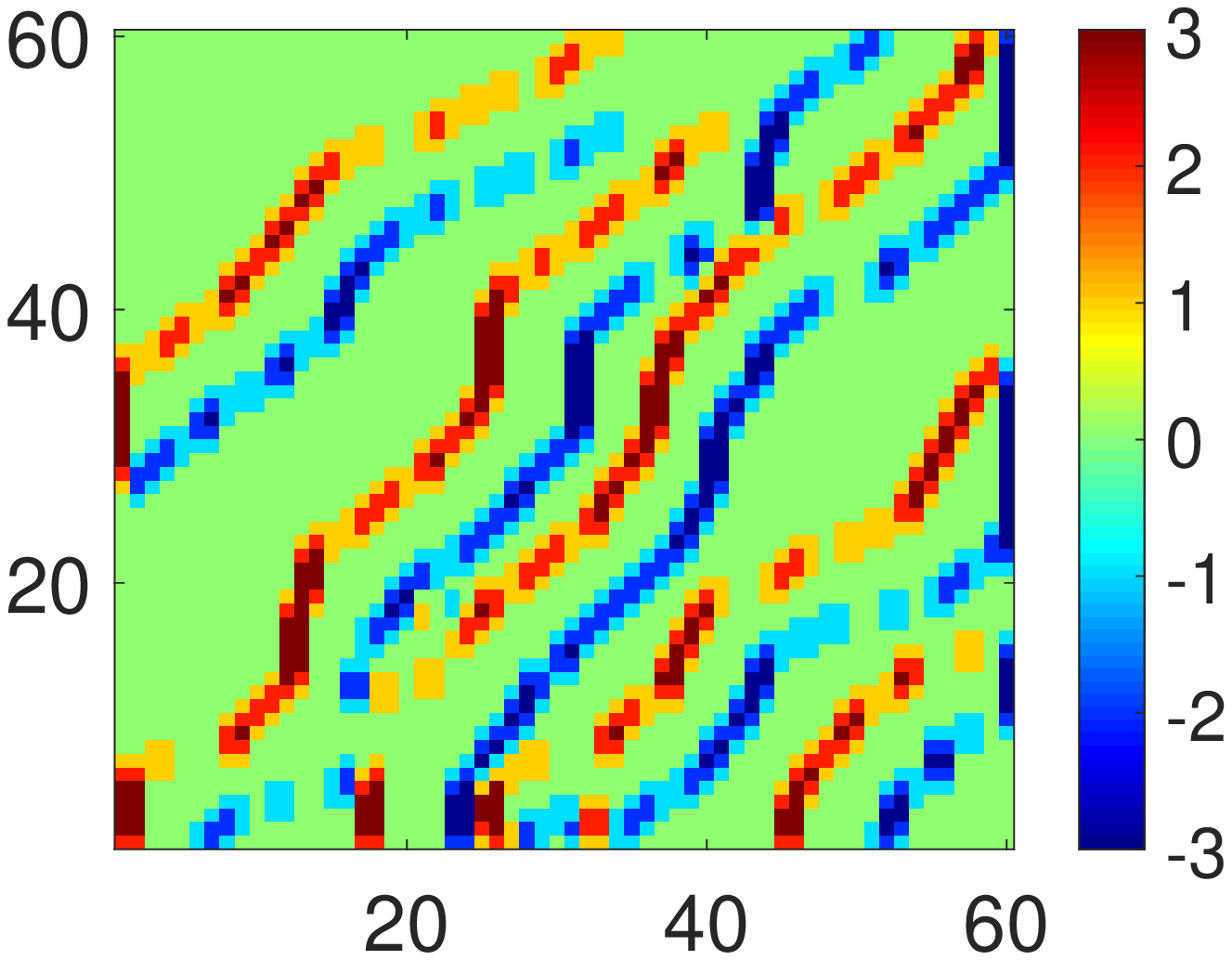}
        \caption{}
    \end{subfigure}%
        ~
    \begin{subfigure}[b]{0.24\textwidth}
        \includegraphics[width=1\textwidth]{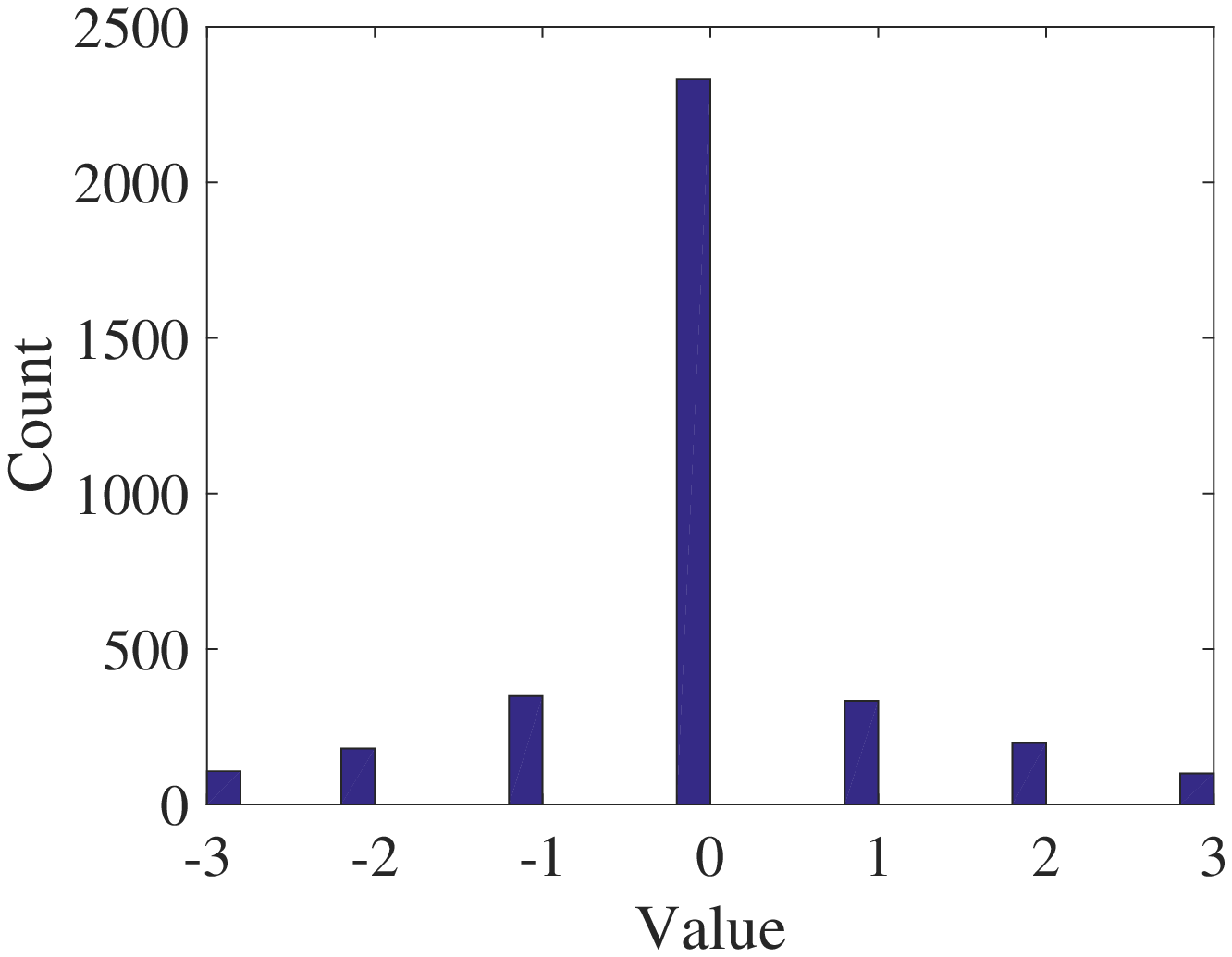}
        \caption{}
    \end{subfigure}
    \caption{Linear filter response of a binary channelized model to a simple $3\times3$ template and the histogram of the filter response. \textbf{a}~Original binary channelized model, \textbf{b} $3\times3$ template, ~\textbf{c}~linear filter response, and \textbf{d} histogram of the filter response.}
    \label{fig-linear-filter}
\end{figure}

Rather than using high-order statistical quantities of the original spatial random field, other treatments instead use low-order summary statistics of filter responses, obtained by scanning `templates' through the spatial random field. In \cite{Zhang2006}, for example, it was shown that the histograms of linear filter responses obtained with simple templates are effective at classifying training images with different patterns. Figure~\ref{fig-linear-filter} displays a binary channelized model $\Bm$ and the linear filter response map, denoted by $F(\Bm)$, obtained with a $3\times3$ template $\Bw$ (also referred to as a filter), and the histogram of $F(\Bm)$. The linear filtering process is accomplished by scanning $\Bm$ with the template $\Bw$. At each location the sum of the element-wise product of $\Bw$ and a local block of $\Bm$ is computed to obtain the value of the corresponding element in $F(\Bm)$. 

Away from the boundary, the mathematical formulation of this linear filtering is
\begin{equation}
F_{i,j}(\Bm) = \sum_{p=-n}^{n}\sum_{q=-n}^{n}\Bw_{i,j}\Bm_{i+p,j+q} + b,
\end{equation}
where $b$ is a constant referred to as bias ($b=0$ in this example), subscripts $i$ and $j$ are $x$ and $y$-direction indices, and the size of the template is $(2n+1)\times(2n+1)$. This scanning process is also referred to as a convolution. For the boundary, zero-padding is used, where one row and one column of zeros are added on each boundary of $\Bm$ such that $F(\Bm)$ is of the same size as $\Bm$. In this case, the simple template $\Bw$ acts to compute the $x$-direction gradient of $\Bm$. The nonzero values in the resulting filter response $F(\Bm)$ indicate the edges (excluding edges aligned with the $x$-axis) between sand and mud. The histogram of $F(\Bm)$ can be used as a metric to characterize the edges in the system. This reflects an important aspect of the spatial correlation in channelized systems.

A recent study by \cite{Gatys2015} in computer vision showed that summary statistics of the filter responses obtained with pre-trained deep convolutional neural networks can effectively characterize multipoint correlations in spatial random fields. In their case the spatial random fields were images and the spatial correlation was depicted by texture images. A convolutional neural network consists of many convolutional layers. Each convolutional layer first performs linear filtering, similar to that described above, on the output from the previous layer (or on the input in the case of the first layer). Then a nonlinear activation function, which in the model transform net described below is `relu' (rectified linear unit, $\max(0,x)$), is applied on the filter response maps to give the final output from the convolutional layer. This output is referred to as the `activation' of the layer. Note that there are typically multiple filters applied at each convolutional layer. The filter response maps associated with the multiple filters are stacked into a third-order tensor. Thus the filters are also often third-order tensors. For a more detailed description of CNN, the reader is referred to \cite{IanGoodfellowYoshuaBengio2016}.

The CNN-based method in \cite{Gatys2015} can be viewed as an extension of the linear filtering method described above. The improvements are mainly two-fold. First, instead of using relatively few hand-crafted templates (filters), a CNN contains a large number of filters obtained automatically through a training process. Second, rather than using linear filtering, each convolutional layer entails a nonlinear filtering process. The activation at different convolutional layers can be viewed as a set of multiscale nonlinear filter responses. In the context of computer vision, the statistical metrics $\{G_k,k=1,...,K\}$ are based on the nonlinear filter responses of the models to a deep CNN pre-trained on image classification. After a CNN is trained for classifying images, the filters at each convolutional layer are capable of capturing different characteristics of the input image. It will be shown that CNNs trained with images are also directly applicable for 2D geological models.

Although not considered in this study, the use of CNNs to characterize multipoint statistics can be extended to 3D models. Potential approaches include layer-by-layer treatments, which may be appropriate when the correlation in the vertical direction is relatively weak (as is often observed in reservoir models). A more general technique is to train a convolutional neural network to capture the full spatial correlation structure of the 3D model. These approaches will be investigated in future work.

For 2D Cartesian models, $\Bm$ can be written as a matrix of dimensions $\Nx \times \Ny$, where $\Nx$ and $\Ny$ denote the number of grid blocks in the $x$ and $y$-directions. Following the RGB format for images, $\Bm$ is converted into a third-order tensor of size $\Nx  \times \Ny \times 3$ by first replicating $\Bm$ three times along the third dimension, and then multiplying by 255 such that, in the binary case, the elements of $\Bm$ are mapped to the range $[0,255]$. The pre-trained CNN for extracting the nonlinear filter responses is the 16-layer VGG network \citep{Simonyan2015a}, trained for image classification over the ImageNet dataset \citep{JiaDeng2009}.

\begin{figure}[!htb]
    \centering
    \includegraphics[width=0.8\textwidth]{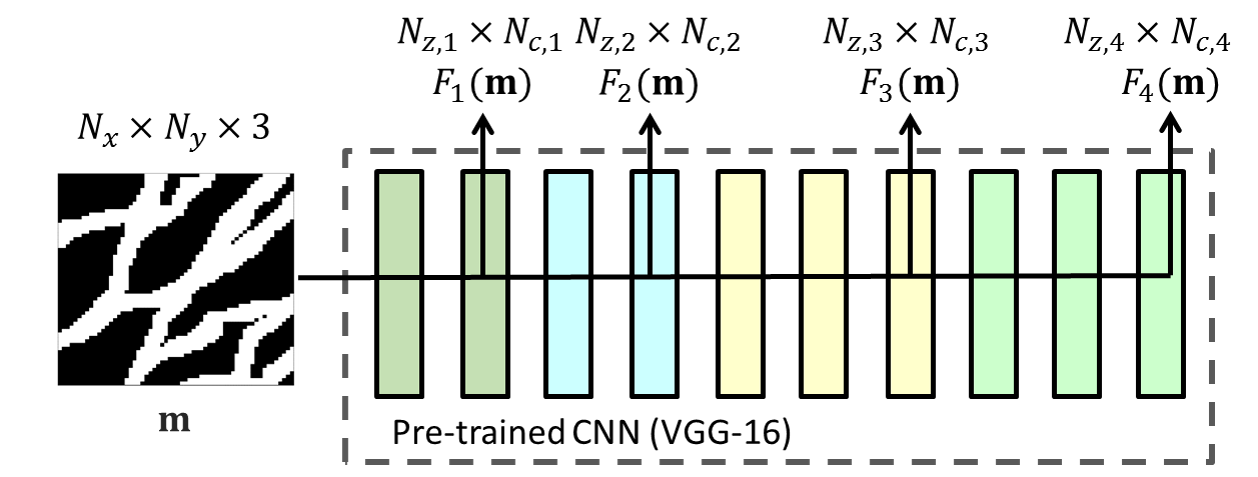}
    \caption{Extracting nonlinear filter responses of $\Bm$ from different layers in a pre-trained convolutional neural network. The dashed outline denotes the first 10 layers of the 16-layer VGG network, trained for image classification. Each box represents a convolutional layer.}
    \label{fig-m-vgg}
\end{figure}

Following suggestions in \citet{Johnson2016}, activations are extracted at four specific layers ($k=1,..,4$), as shown in Fig.~\ref{fig-m-vgg}. The activation at layer $k$, denoted by $\phi_k(\Bm)$, is a third-order tensor of size $N_{x,k}  \times N_{y,k} \times \Nz{k}$, where $N_{x,k}$ and $N_{y,k}$ are the dimensions of $\phi_k(\Bm)$ along the $x$ and $y$-directions, and $\Nz{k}$ is the number of filters in convolutional layer $k$. Next, $\phi_k(\Bm)$ is reshaped into a so-called feature matrix $F_k(\Bm)$ of size $\Nz{k} \times N_{\text{c},k}$, where $N_{\text{c},k}=N_{x,k}\times N_{y,k}$ is the number of elements in each filter response map.

The uncentered covariance matrices of the nonlinear filter responses are referred to as Gram matrices, $G_k(\Bm) \in \BBR^{\Nz{k}\times\Nz{k}}$ \citep{Gatys}. These are defined as
\begin{equation}
G_k(\Bm) = \dfrac{1}{N_{\text{c},k}\Nz{k}}F_k(\Bm)F_k(\Bm)^T .
\end{equation}
Gram matrices thus defined have been shown to represent a set of effective statistical metrics for characterizing the multipoint correlation of $\Bm$. In fact, the style loss is defined in terms of these $G_k$ as follows:
\begin{equation}
\label{eq-l2}
\Ls(\Bx, \mref) = \sum_{k=1}^4\dfrac{1}{\Nz{k}^2}||G_k(\Bx) - G_k(\mref)||_{Fr}^2.
\end{equation}
Note that the dimensions of $\Bx$ and $\mref$ can be different, but those of $G_k(\Bx)$ and $G_k(\mref)$ are both $\Nz{k}\times\Nz{k}$, where $\Nz{k}$ depends only on the architecture of the CNN and is invariant to the input dimension.

For the content loss, the nonlinear filter responses (feature matrices) $F_k(\Bm)$ provide encoded representations of $\Bm$ that capture the content in $\Bm$. As described in \cite{Gatys}, with increasing layer index $k$, the feature matrices $F_k(\Bm)$ encode larger-scale content and become less sensitive to the exact element-wise values in $\Bm$. Therefore, following the treatment of \cite{Johnson2016}, the nonlinear filter response at the second layer is used to construct the content loss. This was shown to provide a reasonable balance between large-scale content and small-scale details. The precise form for the content loss term is thus
\begin{equation}
\label{eq-l1}
\Lc(\Bx,\Bmpca(\Bxi)) = \dfrac{1}{\Nz{2}N_{\text{c},2}}||F_2(\Bx) - F_2(\Bmpca(\Bxi))||_{Fr}^2 .
\end{equation}

Combining Eqs.~\ref{eq_gen_opca}, ~\ref{eq-l2},~and~\ref{eq-l1}, the generalized O-PCA formulation with CNN-based feature representation can be written as
\begin{equation}
\label{eq_gen_opca_cnn}
\Bm(\Bxi) = \argmin{\Bx}\Big\{\dfrac{1}{\Nz{2}N_{\text{c},2}}||F_2(\Bx) - F_2(\Bmpca(\Bxi))||_{Fr}^2 + \gamma_s \sum_{k=1}^4\dfrac{1}{\Nz{k}^2}||G_k(\Bx) - G_k(\mref)||_{Fr}^2\Big\}.
\end{equation}
The goal, for a given realization of $\Bxi$, is to obtain an optimal model $\Bx$ that minimizes the weighted sum of the distance to the underlying PCA model $\Bmpca(\Bxi)$, in terms of the feature matrices, and the distance to the reference model $\mref$, in terms of the Gram matrices. The gradient of the loss function with respect to $\Bx$ can be readily computed using back-propagation through the VGG-16 net, so gradient-based minimization strategies can be used for this problem. The initial guess is typically set to be either a random noise model or the PCA model $\Bmpca(\Bxi)$. 

The optimization in Eq.~\ref{eq_gen_opca_cnn} is still computationally demanding since evaluating the objective function and its gradient requires forward and backward passes through a deep CNN. During history matching, where $\Bxi$ is updated many times, this approach becomes inefficient as it requires solving this optimization problem a large number of times. Therefore, in this study, Eq.~\ref{eq_gen_opca_cnn} is not used to generate realizations. Instead, following the idea proposed in \cite{Johnson2016}, a more efficient algorithm, CNN-PCA, is formulated. In CNN-PCA a separate CNN is trained to form an explicit transform function that post-processes PCA realizations very quickly, thus avoiding the time-consuming optimization in Eq.~\ref{eq_gen_opca_cnn}. The CNN-PCA treatment is now described.

\subsection{CNN-PCA Procedure}
\label{Subsect_cnn_pca}
The optimization in Eq.~\ref{eq_gen_opca_cnn} can be interpreted as a post-processing of the underlying PCA model to match the pattern of the target training image. In other words, Eq.~\ref{eq_gen_opca_cnn} defines an implicit mapping from the PCA model to the post-processed model. To map each PCA model to its corresponding post-processed model requires solving the computationally demanding optimization in Eq.~\ref{eq_gen_opca_cnn}. It would be much more efficient if an explicit mapping function could be obtained such that the post-processing of a PCA model required only an explicit function evaluation.

The idea here is to accomplish such a mapping through use of an alternative CNN, which takes PCA models as input and provides post-processed models as output. This CNN is referred to as the model transform net and is denoted by $f_W$, where the subscript $W$ represents the parameters in the network. Following the well-designed network architecture proposed in \cite{Johnson2016}, the model transform net consists of 16 convolutional layers and includes a total of 1,668,865 parameters. The detailed architecture of the model transform net is specified in the Appendix. 

The training process for the model transform net involves a minimization problem similar to the generalized O-PCA procedure described earlier (Eq.~\ref{eq_gen_opca_cnn}). The difference is that, rather than directly obtaining an optimal post-processed model for each particular PCA realization, the idea here is to obtain optimal model-transform-net parameters such that the resulting models display small content and style losses. The first step of the training process is to construct a training set consisting of $\Nt$ random PCA models $\Bmpca^i$, $i=1,2,..,\Nt$, generated by sampling $\Bxi$ from the standard normal distribution and then applying Eq.~\ref{eq-pca3}. These random PCA models are then fed through the initial model transform net (with random parameters) to obtain the post-processed models $f_W(\Bmpca^i)$, $i=1,2,..,\Nt$. Next, the content and style losses for each pair of (corresponding) PCA and post-processed models are quantified. The parameters in the model transform net can then be modified to minimize the average combined loss over the training set. 

This training process is described by the following minimization:
\begin{equation}
\label{eq_cnnpca}
\begin{split}
\argmin{W}\dfrac{1}{\Nt}\sum_{i=1}^{\Nt}\Big\{&\dfrac{1}{\Nz{2}N_{\text{c},2}}||F_2(f_W(\Bmpca^i)) - F_2(\Bmpca^i)||_{Fr}^2 \\
& + \gamma_s \sum_{k=1}^4\dfrac{1}{\Nz{k}^2}||G_k(f_W(\Bmpca^i)) - G_k(\mref)||_{Fr}^2\Big\},
\end{split}
\end{equation}
where the decision variable $W$ denotes the parameters in the model transform net, and the loss function in Eq.~\ref{eq_cnnpca} is the average over the training set. For each pair of PCA and post-processed models, the loss function is the same as in the generalized O-PCA formulation (Eq.~\ref{eq_gen_opca_cnn}). Note that the training of the model transform net is an unsupervised learning process, as there is no pre-defined `true' or `reference' post-processed model provided for each PCA model in the training set.

\begin{figure}[!htb]
    \centering
    \includegraphics[width=1.0\textwidth]{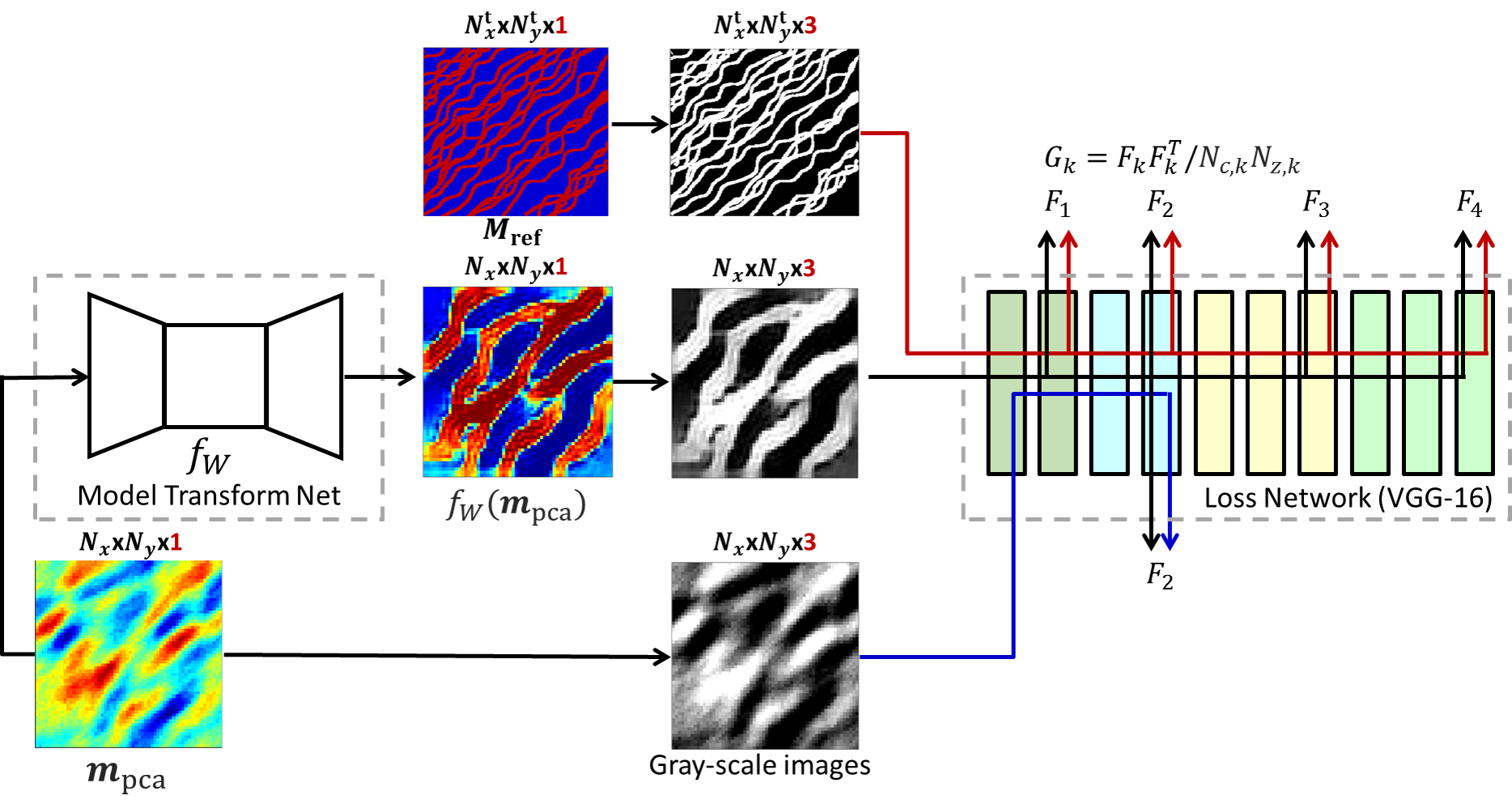}
    \caption{Process for evaluating content and style losses in Eq.~\ref{eq_cnnpca} for one pair of corresponding PCA and post-processed models. Quantifying these losses is required to train the model transform net.}
    \label{fig-cnn-pca}
\end{figure}

Figure~\ref{fig-cnn-pca} illustrates the procedure for evaluating content and style losses for one PCA model in the training set,  for the binary channelized system. The PCA model $\Bmpca$ is fed through the model transform net to obtain the post-processed model $f_W(\Bmpca)$. Next $\Bmpca$, $f_W(\Bmpca)$ and the reference model $\mref$ are converted to gray-scale images and then fed through the `Loss Network' (VGG-16) to extract the feature matrices $F_l$ and then compute the Gram matrices $G_k$, $k=1,..,4$. In this case, the reference model $\mref$ is the training image (Fig.~\ref{fig-ti}) of dimensions $N_x^\text{t} \times N_y^\text{t}$, which is different than the size of the models. Note that the post-processed model $f_W(\Bmpca)$, shown in Fig.~\ref{fig-cnn-pca}, is obtained with the optimal model transform net.

The gradient of the objective function with respect to $W$ can be readily computed using back-propagation through the CNN. The adaptive moment estimation (ADAM) algorithm, which has been proven effective for optimizing various deep neural networks, is used as the optimizer \citep{Kingma2014}. ADAM is an extension of stochastic gradient descent where the gradient at each iteration is approximated using mini-batches of the training set rather than the entire training set at once. In other words, at each iteration, the average loss in Eq.~\ref{eq_cnnpca} is evaluated over a small batch of $\Nb$ PCA models selected from the training set, where $\Nb$ is the batch size. The rate at which decision variables are updated at each iteration is controlled by the learning late $\lr$.

\begin{figure}[!htb]
    \centering
    \includegraphics[width=1.0\textwidth]{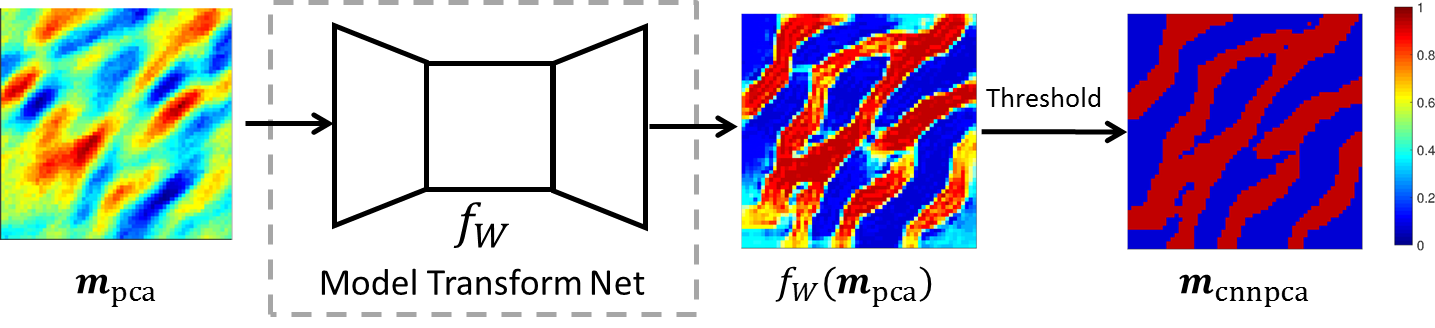}
    \caption{Post-processing a PCA model with the model transform net and a final hard-thresholding step.}
    \label{fig-model-trans-net}
\end{figure}

After model-transform-net training, given a PCA model $\Bmpca$, the post-processed model $f_W(\Bmpca)$ can be obtained very quickly from a forward pass of $\Bmpca$ through the model transform net, as shown in Fig.~\ref{fig-model-trans-net}. Note that the transformation through the model transform net is differentiable. The derivative $\partial f_W(\Bmpca) / \partial \Bmpca$ can be obtained using back-propagation through the model transform net. This enables the direct calculation of $\partial f_W(\Bmpca) / \partial \Bxi$ (using the chain rule and the fact that $\partial \Bmpca / \partial \Bxi = U_l\Sigma _l$). Although in this study gradient-free history matching is applied (as described in Sect.~\ref{sec-hm}), for more efficient gradient-based methods $\partial f_W(\Bmpca) / \partial \Bxi$ is required. A significant advantage of this approach is that it provides the necessary derivative.

It can be seen in Fig.~\ref{fig-model-trans-net} that $f_W(\Bmpca)$ is not strictly binary. A final histogram transformation can be applied, if desired, to ensure the post-processed models are binary or very close to binary. Since gradient-free history matching is used in this work, a simple non-differentiable hard-thresholding step, with a cutoff value of 0.5, is applied here. Application of a differentiable histogram transformation will preserve the differentiability of the CNN-PCA procedure. As discussed in the extension of CNN-PCA to bimodal systems in Sect.~\ref{sec-bimodal-df}, one option is to use O-PCA as a final post-processing step.

The overall CNN-PCA procedure is summarized in Algorithm~\ref{alg-cnn-pca}. The CNN-PCA algorithm used here is modified from an open source implementation \citep{Kadian2018} of the procedure in \cite{Johnson2016} within the PyTorch Python library \citep{Paszke2017}. In PyTorch and many other deep-learning frameworks, back propagation through neural networks for computing gradients can be readily accomplished using auto-differentiation.

\begin{algorithm}[!htb]{
\caption{CNN-PCA Procedure}\label{alg-cnn-pca}
\SetEndCharOfAlgoLine{}
\setstretch{1.2}
\textbf{Construct PCA} \\
\Indp 
Generate $\Nr$ realizations of $\Bm$ based on a training image using the `snesim' algorithm within SGeMS\\
Construct data matrix $Y$ in Eq.~\ref{eq-pca1}, perform singular value decomposition of $Y$, and determine the reduced dimension $l$ to obtain matrices $U_l$ and $\Sigma_l$ in Eq.~\ref{eq-pca3}\\
\Indm 
\textbf{Train the model transform net} \\
\Indp 
Sample $\Nt$ random realizations of $\Bxi$  from $N(0,I_l)$ and obtain $\Nt$ random PCA models by applying Eq.~\ref{eq-pca3}\\
Train the model transform net $f_W$ with Eq.~\ref{eq_cnnpca} or Eq.~\ref{eq_cnnpca_hd} (if hard data are included) using the ADAM algorithm with batch size $\Nb$ and learning rate $\lr$\\
\Indm 
\textbf{Generate new random realizations using CNN-PCA model} \\
\Indp 
Sample $\Bxi$  from $N(0,I_l)$ and obtain the PCA model $\Bmpca(\Bxi)$ by applying Eq.~\ref{eq-pca3} \\
Feed $\Bmpca(\Bxi)$ through the trained model transform net to obtain $f_W(\Bmpca(\Bxi))$, and apply a final histogram transformation to obtain the CNN-PCA model $\Bmcnnpca(\Bxi)$\\

\Indm 
} 
\end{algorithm}

\subsection{Hard Data Loss}
\label{sec-hd}
Hard data are measurement data of static geological properties, such as the permeability values, at particular locations (typically at exploration and production wells). PCA models will always honor hard data if the $\Nr$ SGeMS realizations used to construct the PCA basis honor the hard data. Although the CNN-PCA models $\Bmcnnpca$ are post-processed from $\Bmpca$, training the model transform net with Eq.~\ref{eq_cnnpca} does not guarantee that hard data will be honored. To address this issue, which would be of concern in practical applications, an additional hard data loss term is introduced. The CNN-PCA minimization for conditional systems is now given by
\begin{equation}
\label{eq_cnnpca_hd}
\begin{split}
\argmin{W}\dfrac{1}{\Nt}\sum_{i=1}^{\Nt}\Big\{&\dfrac{1}{\Nz{2}N_{\text{c},2}}||F_2(f_W(\Bmpca^i)) - F_2(\Bmpca^i)||_{Fr}^2 \\
&+ \gamma_s \sum_{k=1}^4\dfrac{1}{\Nz{k}^2}||G_k(f_W(\Bmpca^i)) - G_k(\mref)||_{Fr}^2 + \gamma_h \Lh \Big\}.
\end{split}
\end{equation}
Here $\gamma_h$ is a weighting factor, referred to as the hard-data weight, for the new loss $\Lh$. This hard-data loss is given by
\begin{equation}
\label{eq_h_def}
\Lh = \dfrac{1}{\Nh}\left[\Bh^T(\Bm_{\text{pca}}^i-f_W(\Bmpca^i))^2\right],
\end{equation}
where $\Bh \in \BBR ^ {\Nc \times 1}$ is a hard-data indicator vector, with $h_{j}=1$ meaning there are hard data for grid block $j$ and $h_{j}=0$ meaning there are not hard data for grid block $j$, the square difference $(\Bm_{\text{pca}}^i-f_W(\Bmpca^i))^2$ is evaluated component by component (to give a vector), and $\Nh$ is the number of hard-data values specified. The value of $\gamma_h$ is found through numerical experimentation. Specifically, in this study, $\gamma_h$ is determined such that all hard data are correctly honored over a large set of CNN-PCA models.

\section{Model Generation Using CNN-PCA}
\label{sec-model-gen}
In this and the following sections, CNN-PCA performance is assessed through a variety of evaluations. These assessments follow those presented in \cite{Vo2014}, in that the (visual) quality of random CNN-PCA realizations is first considered, then the flow responses for an ensemble of models are assessed, and finally history matching results are evaluated. In the assessment in this section, CNN-PCA will be applied to generate new models for the binary system considered in Sect.~\ref{subsec-opca}. Both unconditional and conditional models will be compared with O-PCA realizations.

\subsection{Unconditional Realizations}
\label{sec-res-uc}
\begin{figure}[!htb]
    \centering
    \floatbox[{\capbeside\thisfloatsetup{capbesideposition={left,top},capbesidewidth=8.5cm}}]{figure}[\FBwidth]
{\caption{One unconditional SGeMS realization of the binary facies model for the 2D channelized reservoir.}\label{fig-uc-sgems}{\hspace{2.2\baselineskip}}}
    {\includegraphics[width=0.35\textwidth]{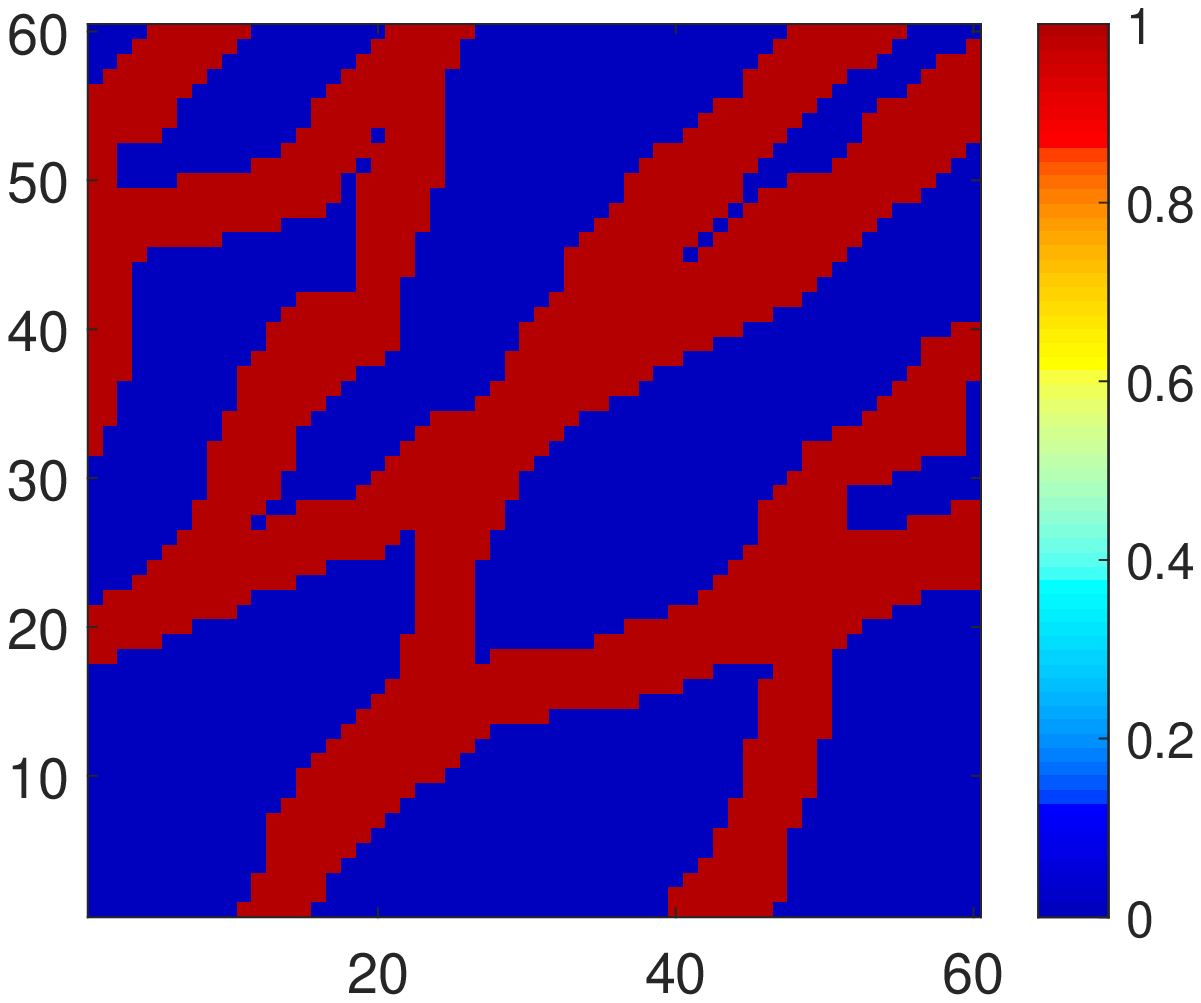}}
\end{figure}

Conditional realizations generated from SGeMS, PCA, and O-PCA for a binary system (Fig.~\ref{fig-binary-cond}) were presented in Sect.~\ref{subsec-opca}. The focus now is on the generation of unconditional realizations. The training image and model setup are the same as described in Sect.~\ref{subsec-opca}. A total of $\Nr=1000$ SGeMS unconditional realizations (Fig.~\ref{fig-uc-sgems}) are generated and used to construct the PCA representation (Eq.~\ref{eq-pca3}), with a reduced dimension $l=70$. For the O-PCA representation (Eq.~\ref{eq_opca_1}), a weighting factor of $\gamma=0.8$ is used \citep{Vo2014}. Note that the same set of SGeMS realizations is used for PCA, O-PCA, and CNN-PCA model construction.

\begin{figure}[!htb]
    \centering
    \begin{subfigure}[b]{0.325\textwidth}
        \includegraphics[width=1\textwidth]{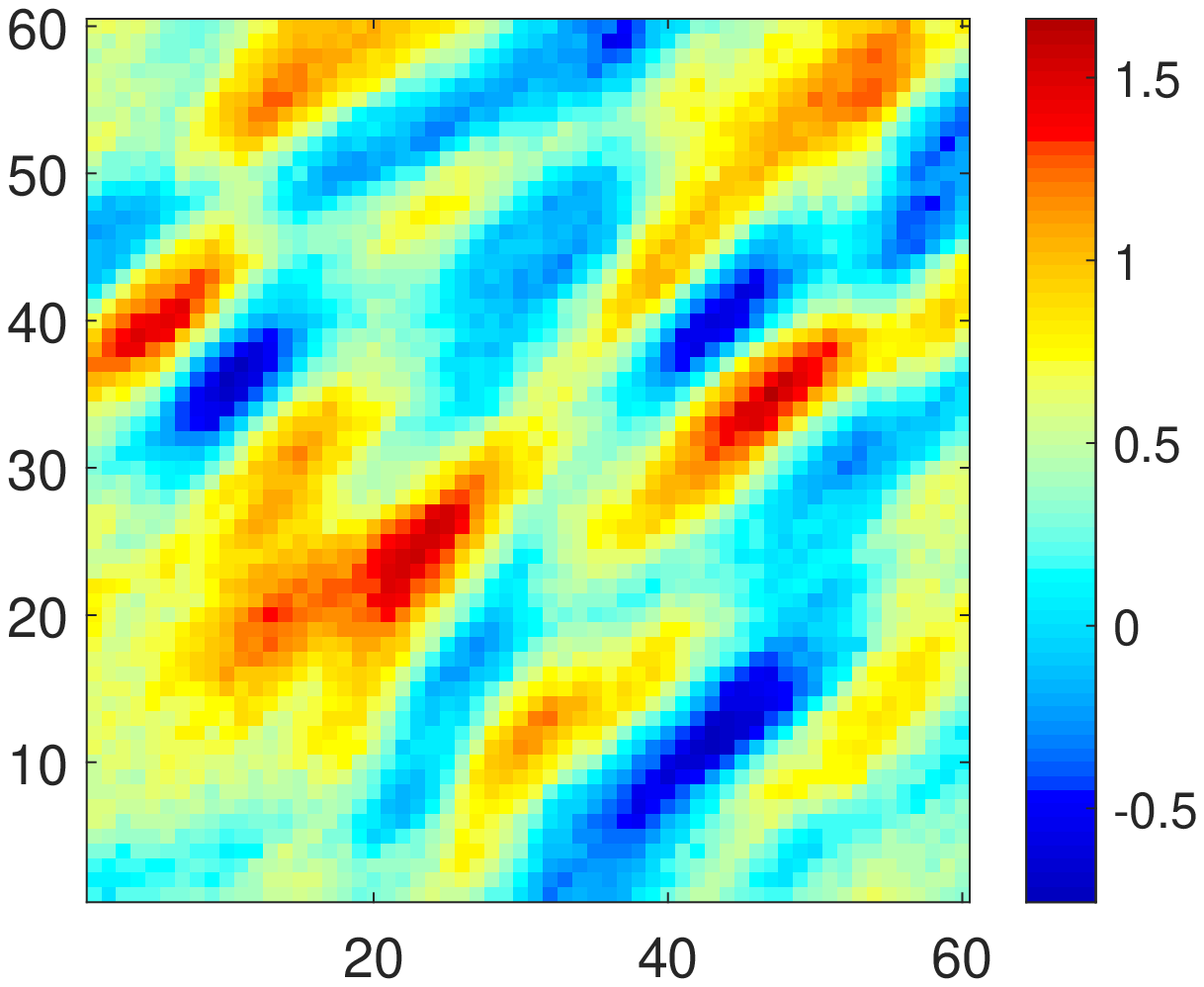}
        \caption{}
    \end{subfigure}%
    ~ 
    \begin{subfigure}[b]{0.32\textwidth}
        \includegraphics[width=1\textwidth]{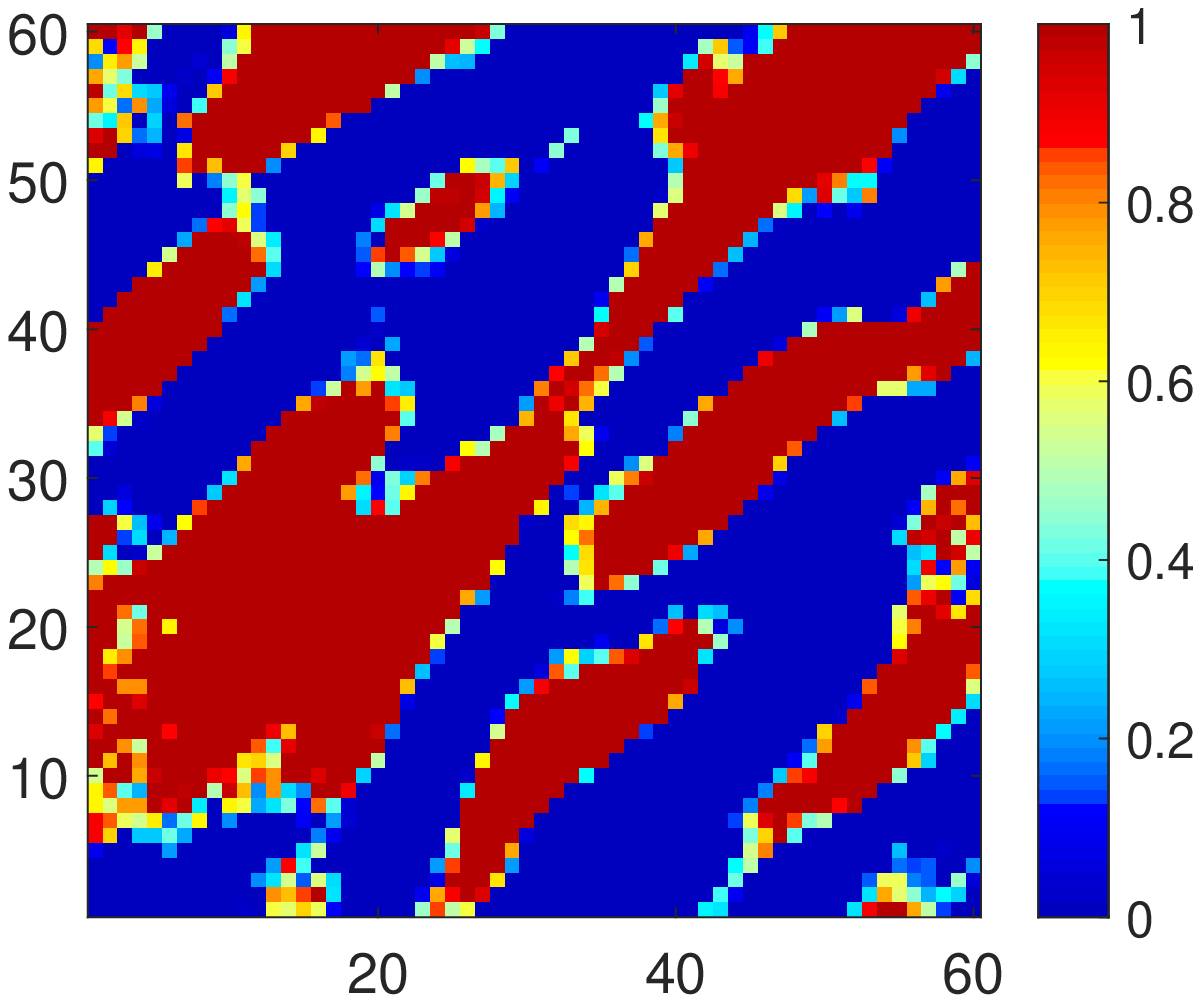}
        \caption{}
    \end{subfigure}%
    ~
    \begin{subfigure}[b]{0.32\textwidth}
        \includegraphics[width=1\textwidth]{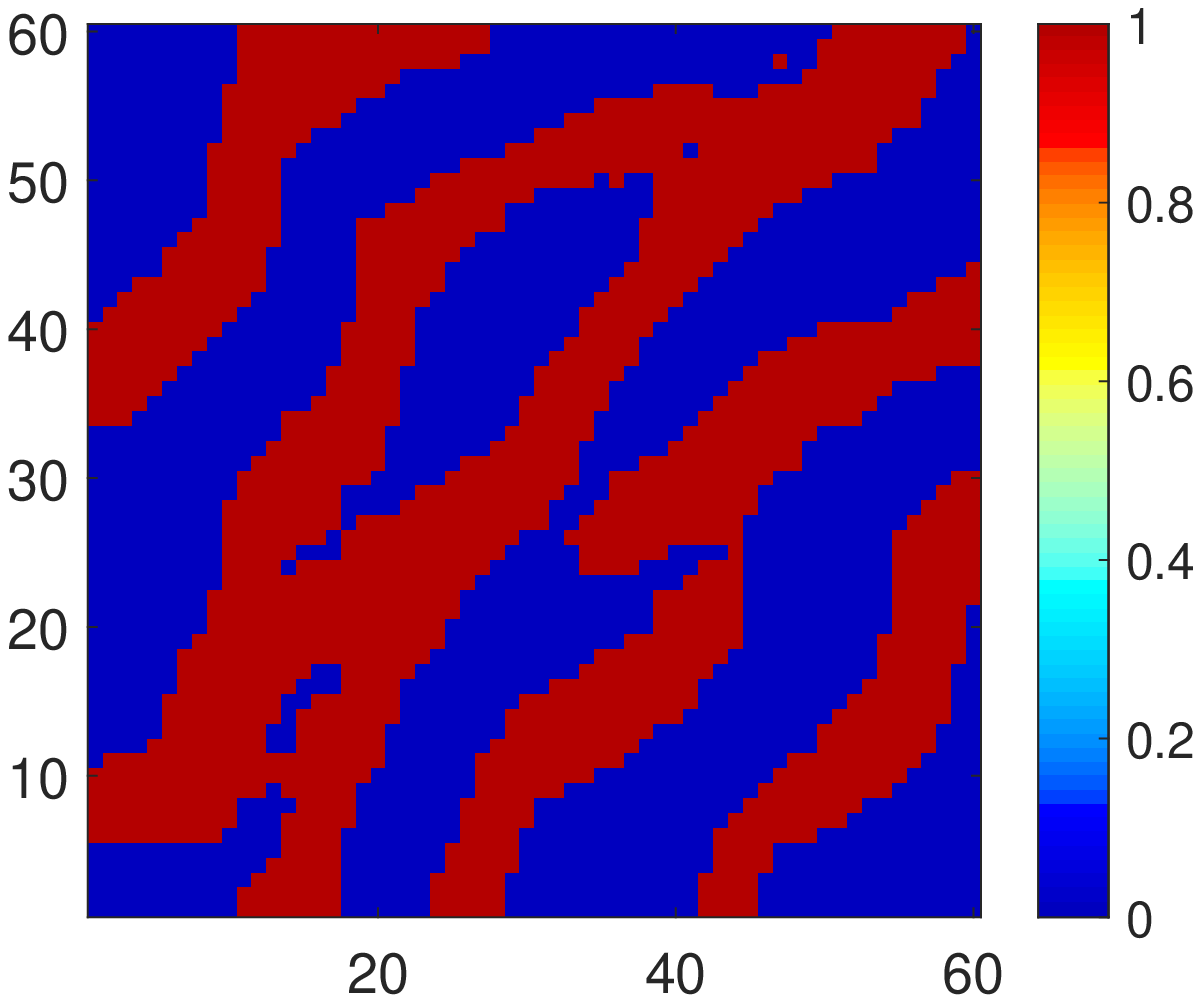}
        \caption{}
    \end{subfigure}%

    \hspace{2\baselineskip}
    \begin{subfigure}[b]{0.325\textwidth}
        \includegraphics[width=1\textwidth]{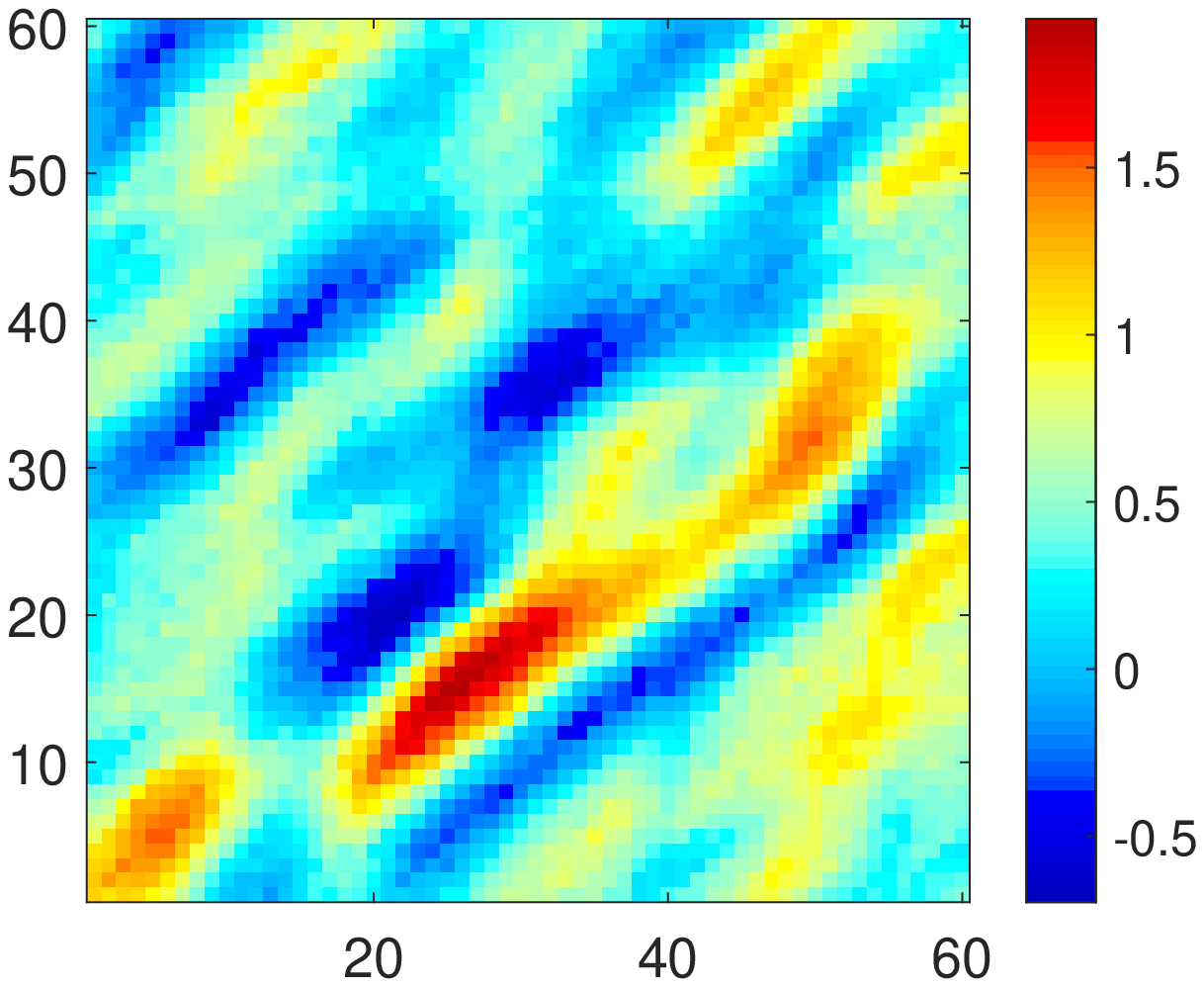}
        \caption{}
    \end{subfigure}%
    ~
    \begin{subfigure}[b]{0.32\textwidth}
        \includegraphics[width=1\textwidth]{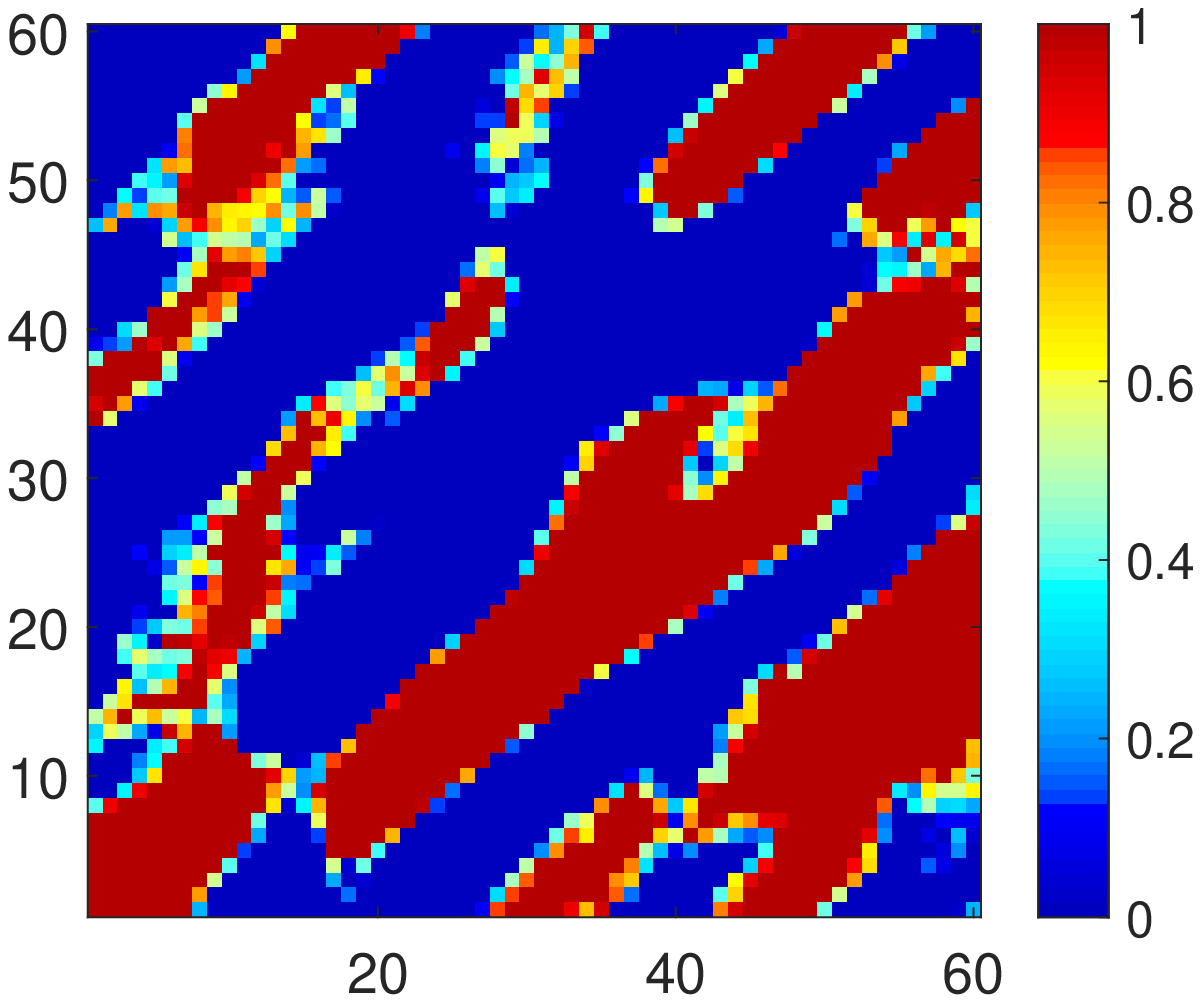}
        \caption{}
    \end{subfigure}%
    ~ 
    \begin{subfigure}[b]{0.32\textwidth}
        \includegraphics[width=1\textwidth]{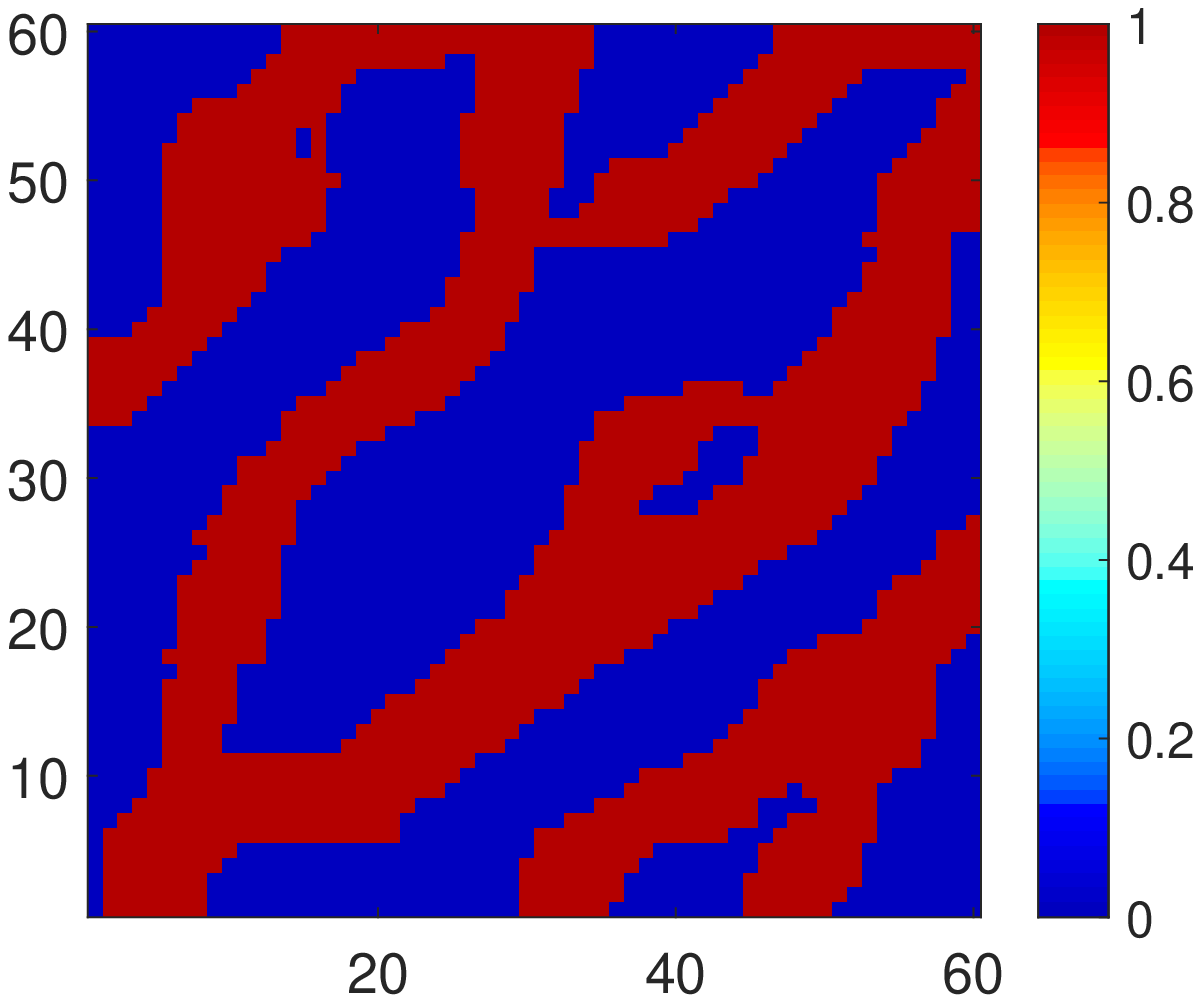}
        \caption{}
    \end{subfigure}%
    \caption{Unconditional realizations for the 2D channelized facies model. \textbf{a, d} Two PCA realizations, \textbf{b, e} two corresponding O-PCA realizations, and \textbf{c, f} two corresponding CNN-PCA realizations.}
    \label{fig-uc-res}
\end{figure}

Figure~\ref{fig-uc-res} displays two random PCA model realizations and the corresponding O-PCA and CNN-PCA models. It is evident that the O-PCA models (Fig.~\ref{fig-uc-res}b, e) display sharper features in comparison to the corresponding PCA models (Fig.~\ref{fig-uc-res}a, d). However, the channels in the O-PCA models lack the degree of connectivity evident in the training image (Fig.~\ref{fig-ti}) and in the SGeMS realization (Fig.~\ref{fig-uc-sgems}). In addition, the channel width in the training image is very consistent, while in the O-PCA realizations large variations are evident in Fig.~\ref{fig-uc-res}c, e. These behaviors were also observed by \cite{Vo2014}.

CNN-PCA realizations are generated following Algorithm~\ref{alg-cnn-pca}. The first step is to train the model transform net. The training set consists of $\Nt=3000$ random PCA realizations (Eq.~\ref{eq-pca3}). The value of the style weight $\gamma_s$ was investigated over a range from 0.05 to 3. Based on visual inspection, a value of $\gamma_s = 0.3$ was found to provide CNN-PCA models with the best balance between resemblance to the PCA models and consistency with the training image (additional values of $\gamma_s$ are considered below). The model transform net was then trained with the ADAM algorithm for a total of 2250 iterations, with a batch size of $\Nb=4$ and a learning rate of $\lr=0.001$. The values of these parameters were obtained through numerical experiments. The training process requires around 3~minutes with one GPU (NVIDIA Tesla K80) for this case. With the trained model transform net, new CNN-PCA models are generated following Step~8 and 9 in Algorithm~\ref{alg-cnn-pca}. 

Figure~\ref{fig-uc-res}c, f shows the CNN-PCA realizations corresponding to the PCA realizations in Fig.~\ref{fig-uc-res}a, d. It is clear that the binary features and the channel width and connectivity show significant improvement relative to that in the PCA and O-PCA realizations. In addition, the relative locations of sand and mud in the CNN-PCA realizations are consistent with the trends in the corresponding PCA realizations. These comparison results demonstrate that CNN-PCA realizations can better preserve the geological features displayed in the  training image (Fig.~\ref{fig-ti}). This suggests that the CNN-PCA formulation described in Sect.~\ref{Subsect_cnn_pca} is indeed able to capture higher-order spatial statistics.

\begin{figure}[!htb]
    \centering
    \begin{subfigure}[b]{0.24\textwidth}
        \includegraphics[width=1\textwidth]{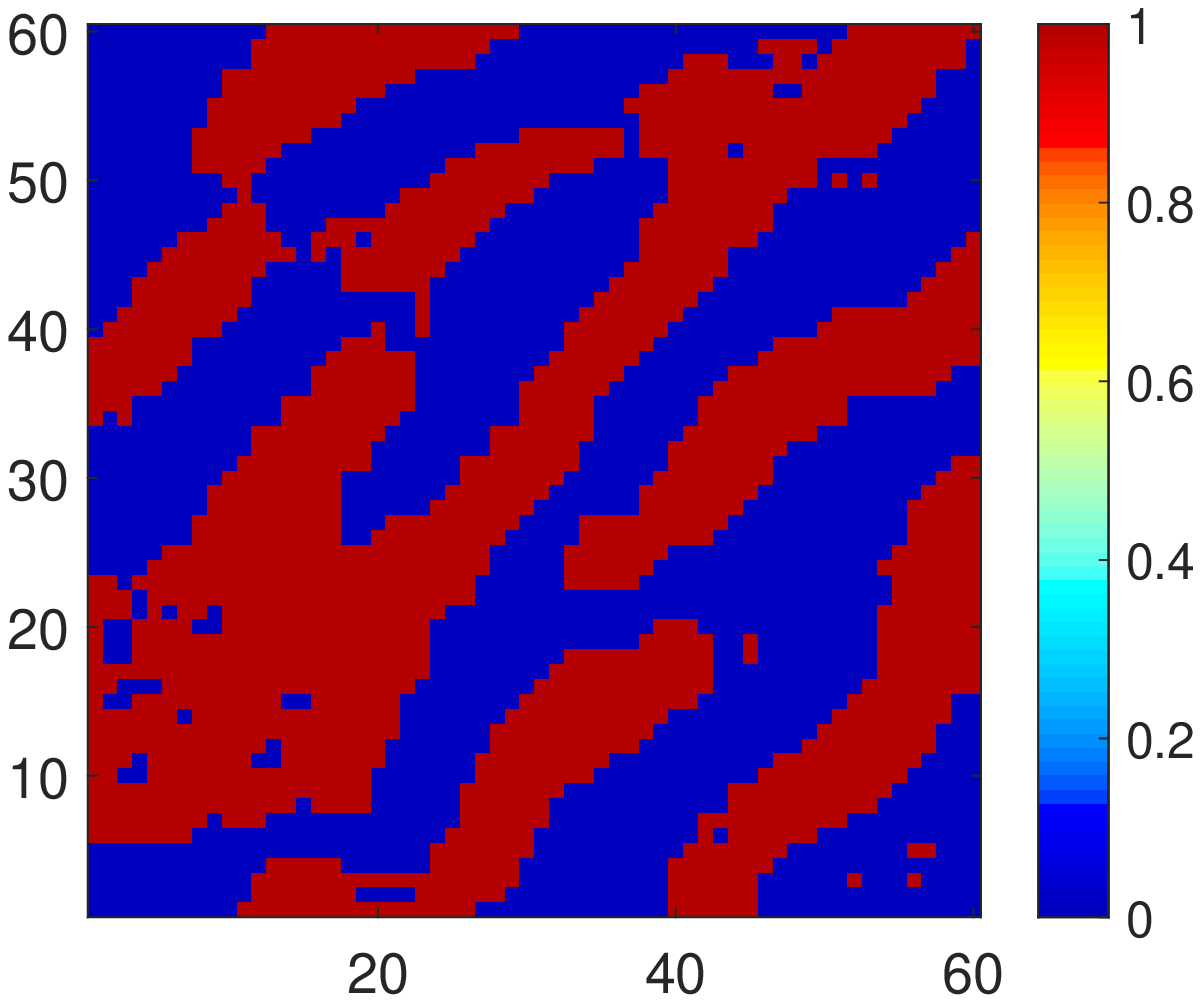}
        \caption{}
    \end{subfigure}%
    ~
    \begin{subfigure}[b]{0.24\textwidth}
        \includegraphics[width=1\textwidth]{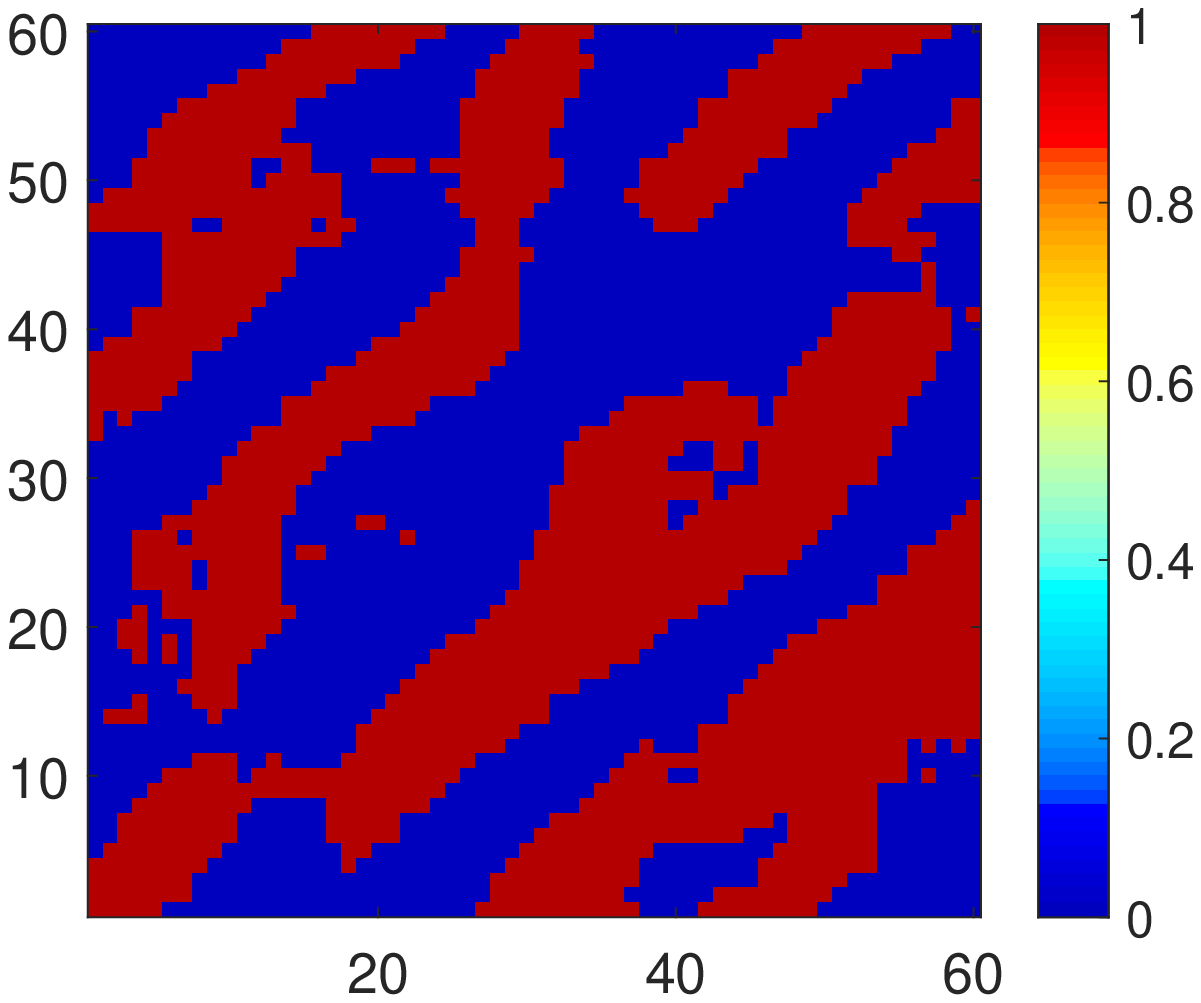}
        \caption{}
    \end{subfigure}%
    ~
    \begin{subfigure}[b]{0.24\textwidth}
        \includegraphics[width=1\textwidth]{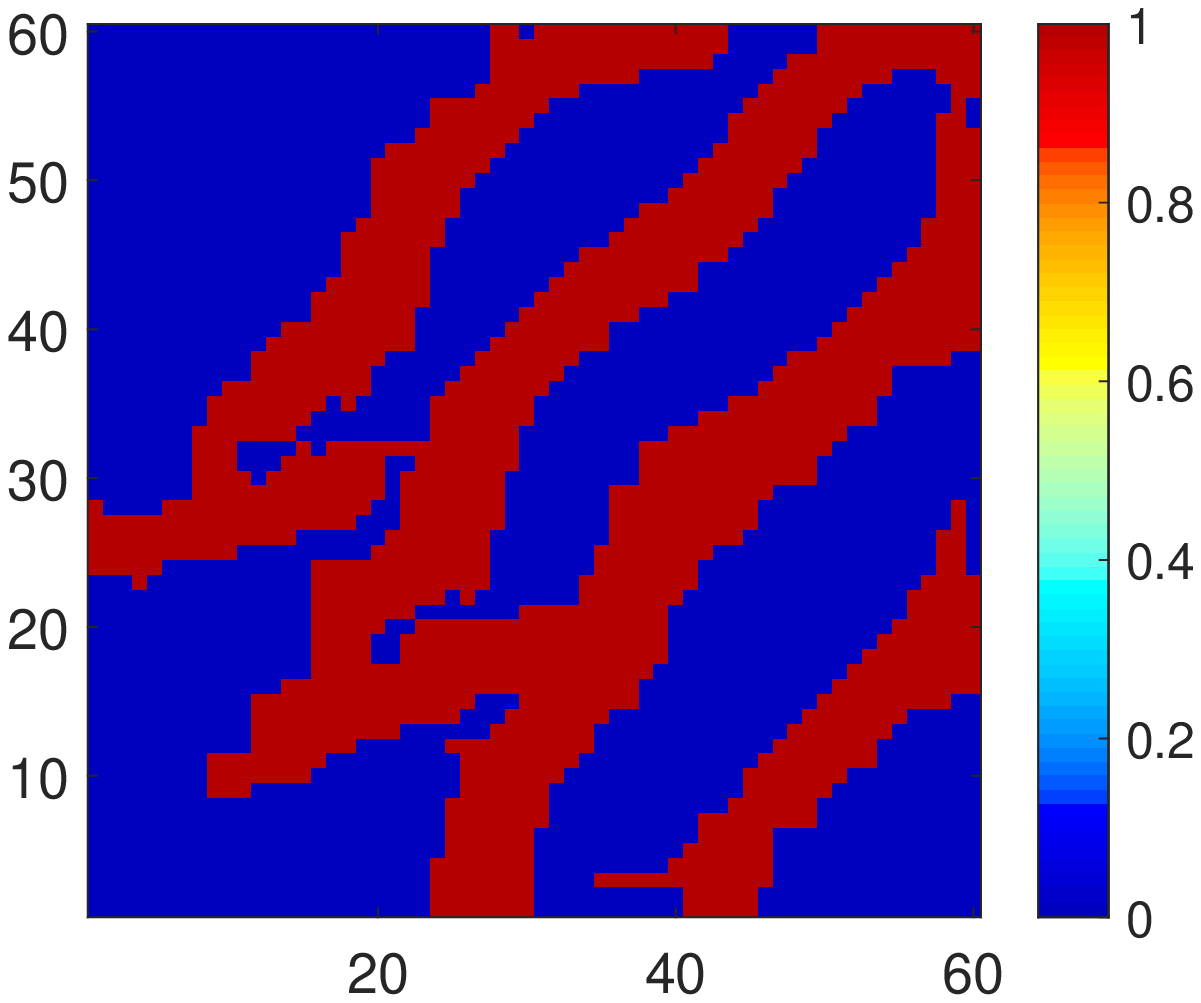}
        \caption{}
    \end{subfigure}%
        ~
    \begin{subfigure}[b]{0.24\textwidth}
        \includegraphics[width=1\textwidth]{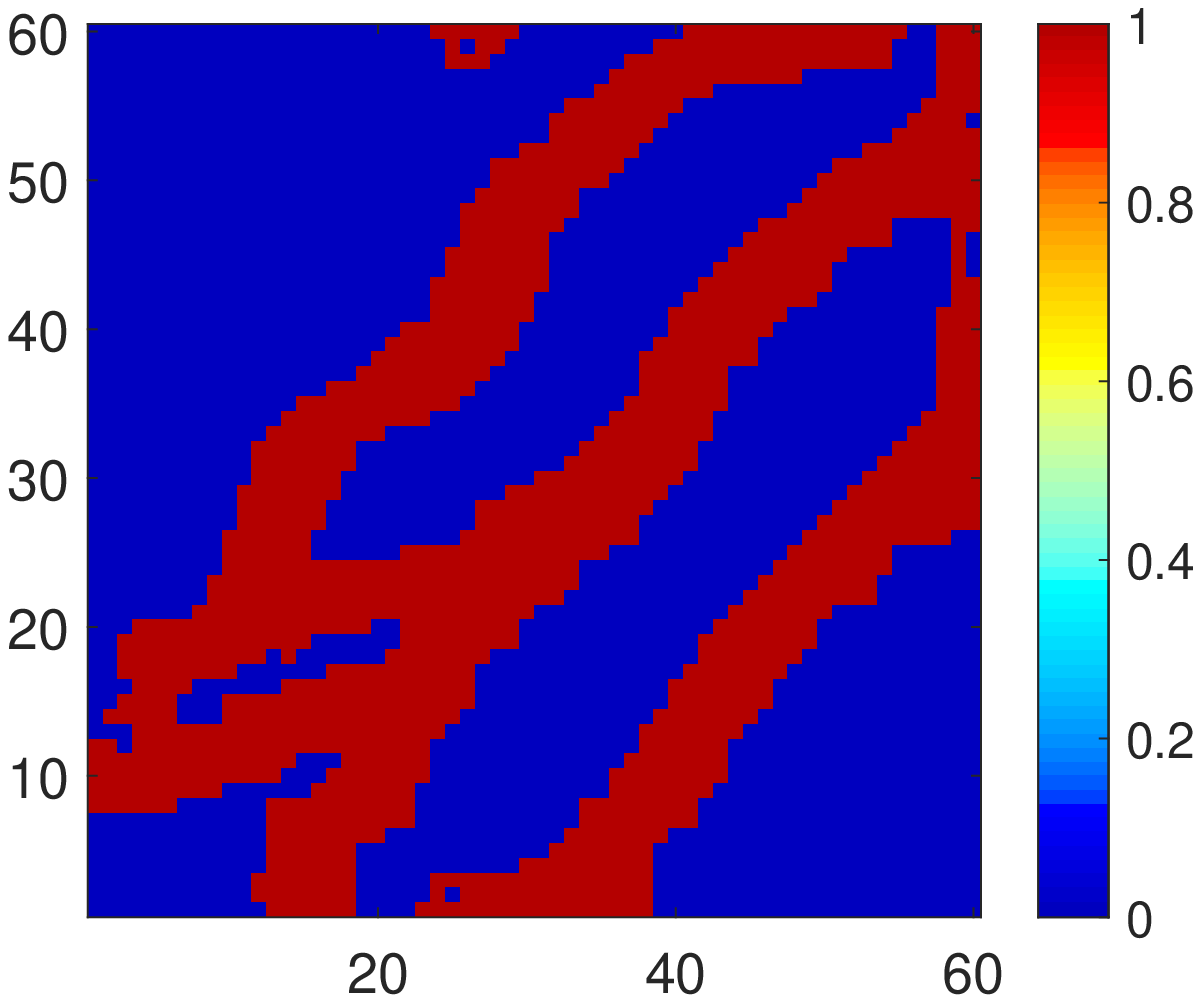}
        \caption{}
    \end{subfigure}
    \caption{CNN-PCA realizations with different style weight $\gamma_s$. \textbf{a, b} $\gamma_s = 0.05$ and \textbf{c, d} $\gamma_s = 3$.}
    \label{fig-weight}
\end{figure}

The impact of style weight $\gamma_s$ on CNN-PCA models is now illustrated. In general, small $\gamma_s$ leads to models that closely resemble the corresponding PCA models but display incorrect spatial correlation, while large $\gamma_s$ values have the opposite effect. In addition, the variability of CNN-PCA realizations collapses (i.e., all the CNN-PCA realizations become almost identical) when using overly large $\gamma_s$. These observations are apparent in Fig.~\ref{fig-weight}. Realizations constructed using a small style weight of $\gamma_s = 0.05$, shown in Fig.~\ref{fig-weight}a, b are basically block-wise mappings of the underlying PCA models (Fig.~\ref{fig-uc-res}a, d). They display relatively poor channel continuity and inconsistent channel width. With a large style weight of $\gamma_s=3$, the CNN-PCA realizations do not honor the general sand and mud locations indicated in the corresponding PCA models. They also show less variability in terms of channel geometry. A value of $\gamma_s=0.3$, used in this study, is seen to provide realizations (Fig.~\ref{fig-uc-res}c, f) that represent an appropriate balance between the two loss functions.

\subsection{Conditional Realizations}
\label{Subsect_cnn_con_real}
\begin{figure}[!htb]
    \centering
    \begin{subfigure}[b]{0.325\textwidth}
        \includegraphics[width=1\textwidth]{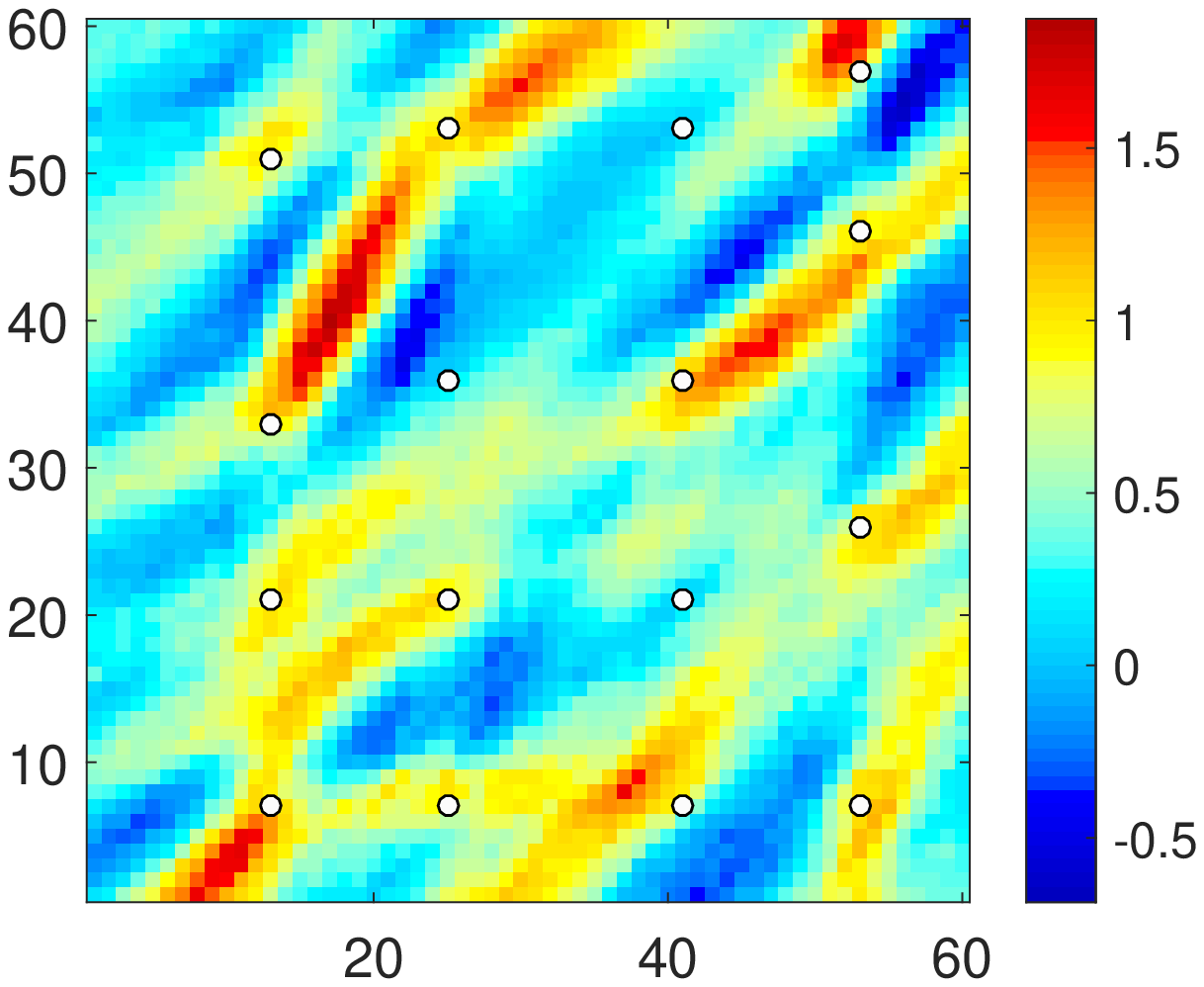}
        \caption{}
    \end{subfigure}%
    ~ 
    \begin{subfigure}[b]{0.32\textwidth}
        \includegraphics[width=1\textwidth]{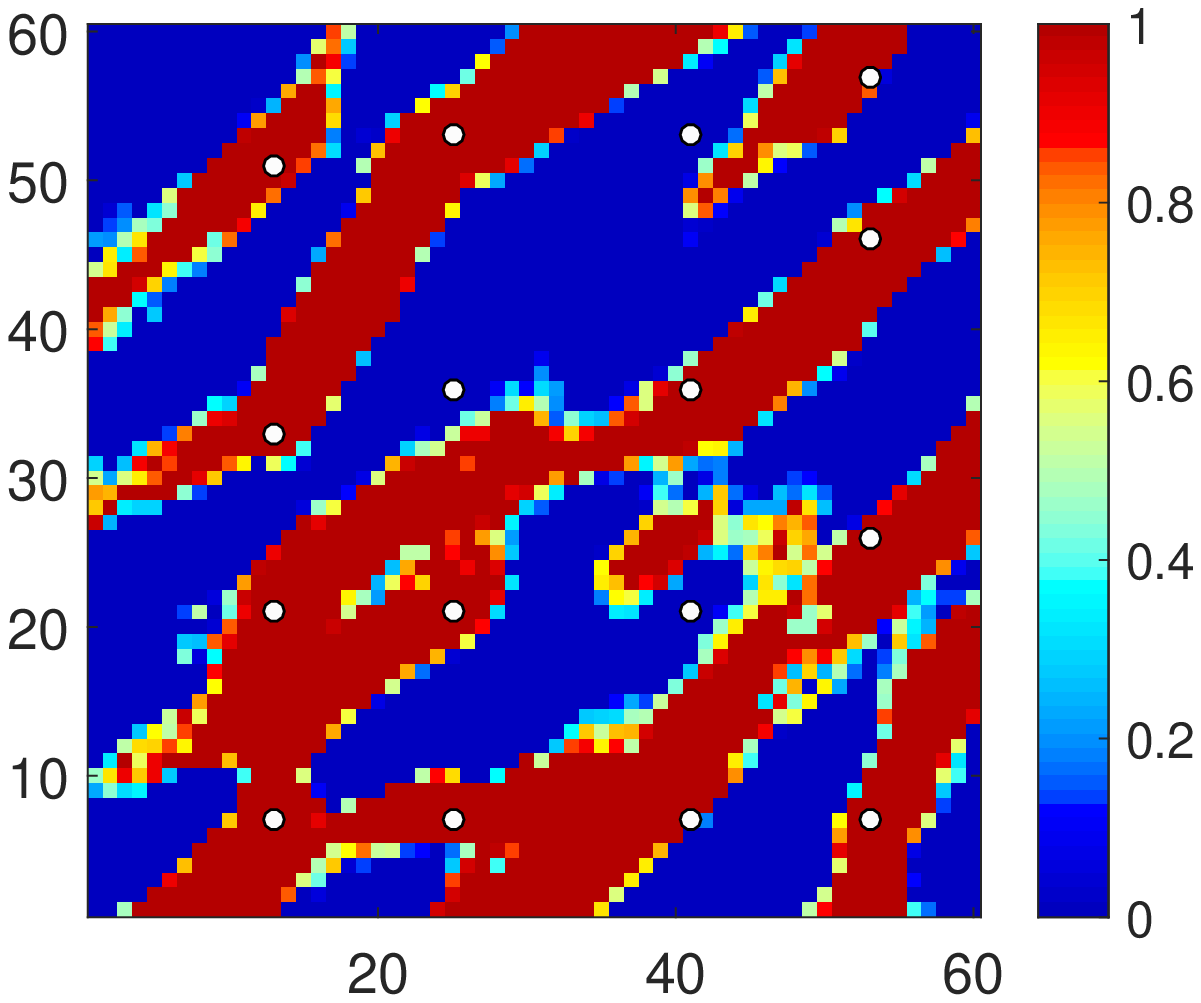}
        \caption{}
    \end{subfigure}%
    ~
    \begin{subfigure}[b]{0.32\textwidth}
        \includegraphics[width=1\textwidth]{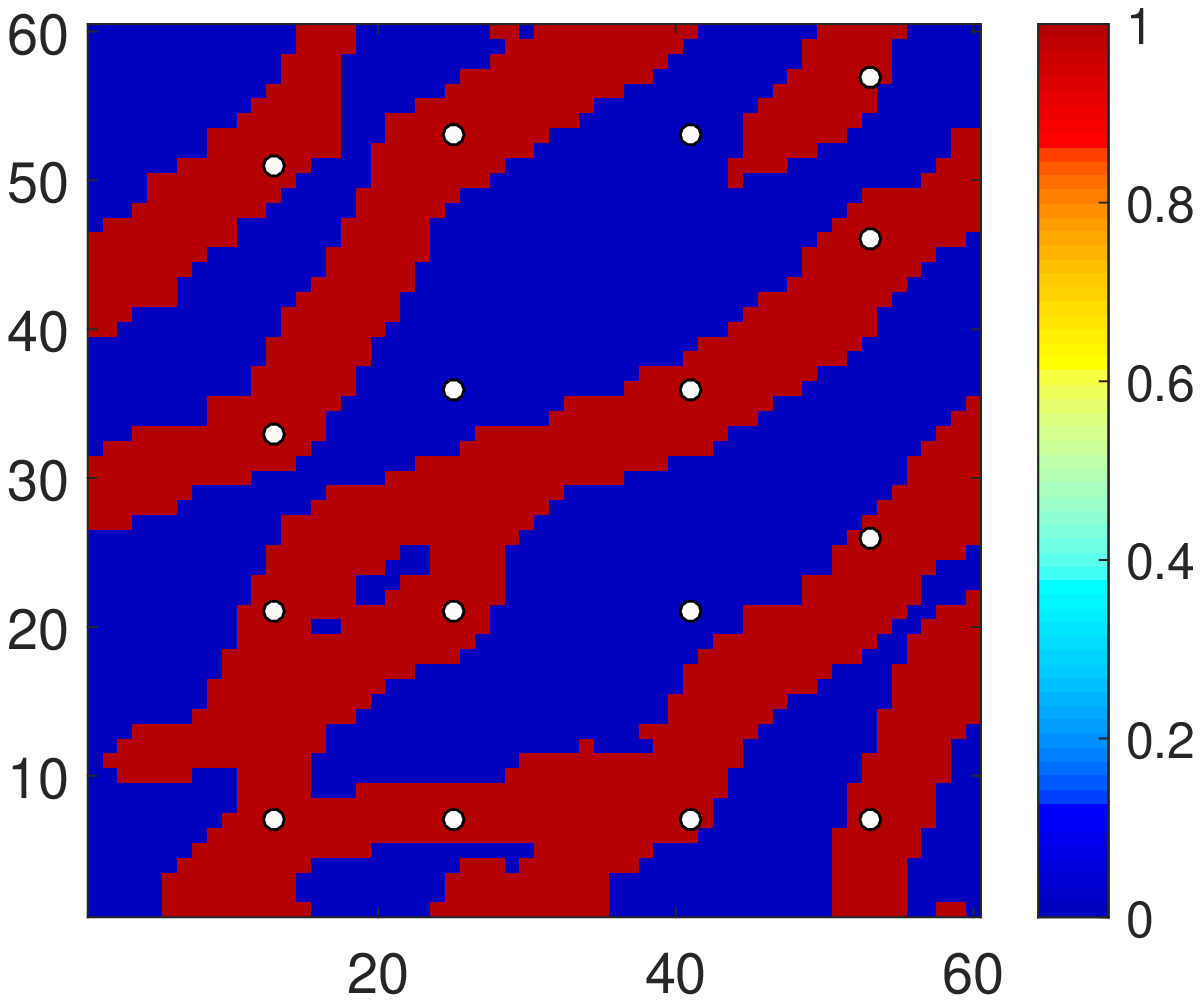}
        \caption{}
    \end{subfigure}%

    \begin{subfigure}[b]{0.325\textwidth}
        \includegraphics[width=1\textwidth]{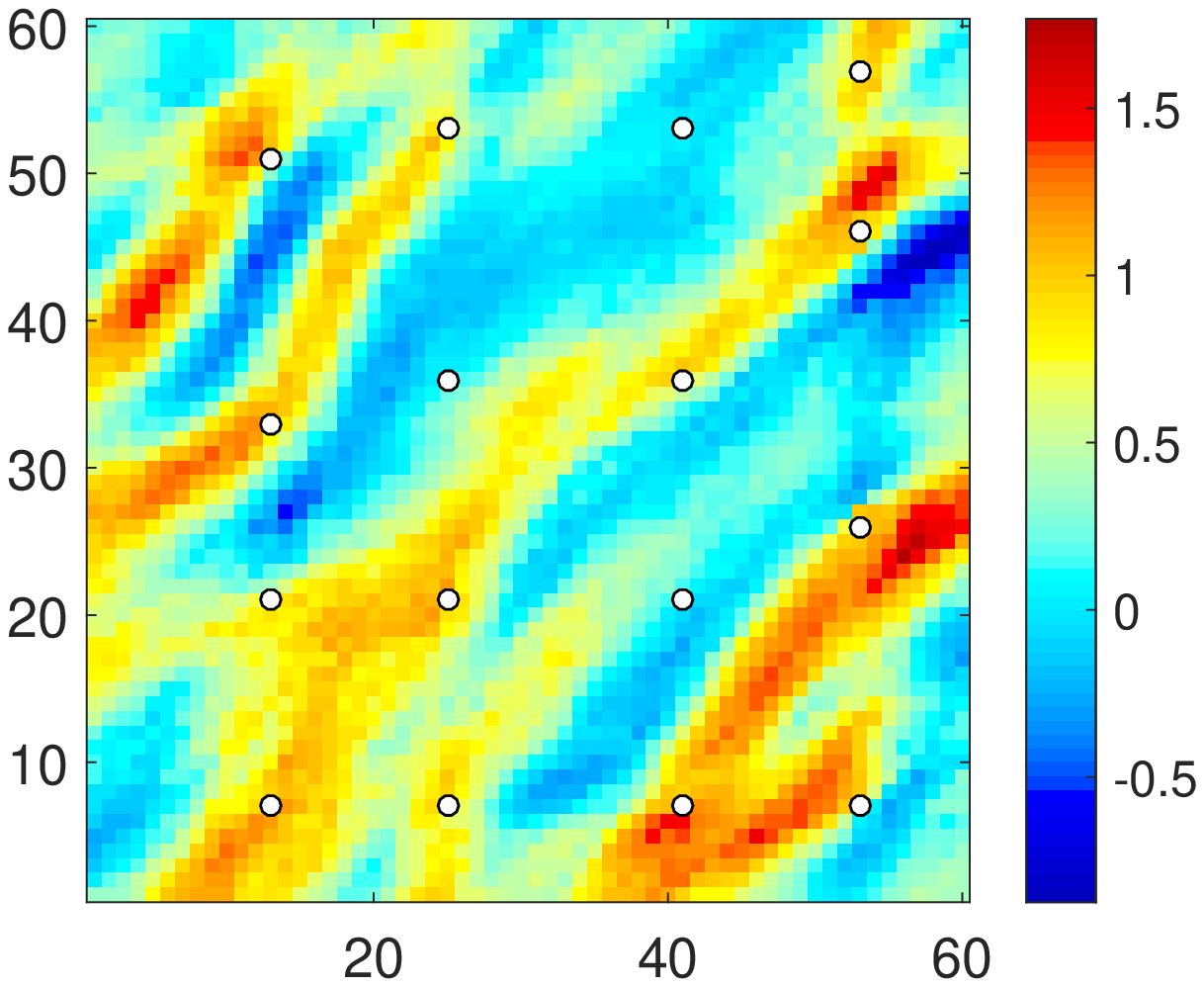}
        \caption{}
    \end{subfigure}%
    ~
    \begin{subfigure}[b]{0.32\textwidth}
        \includegraphics[width=1\textwidth]{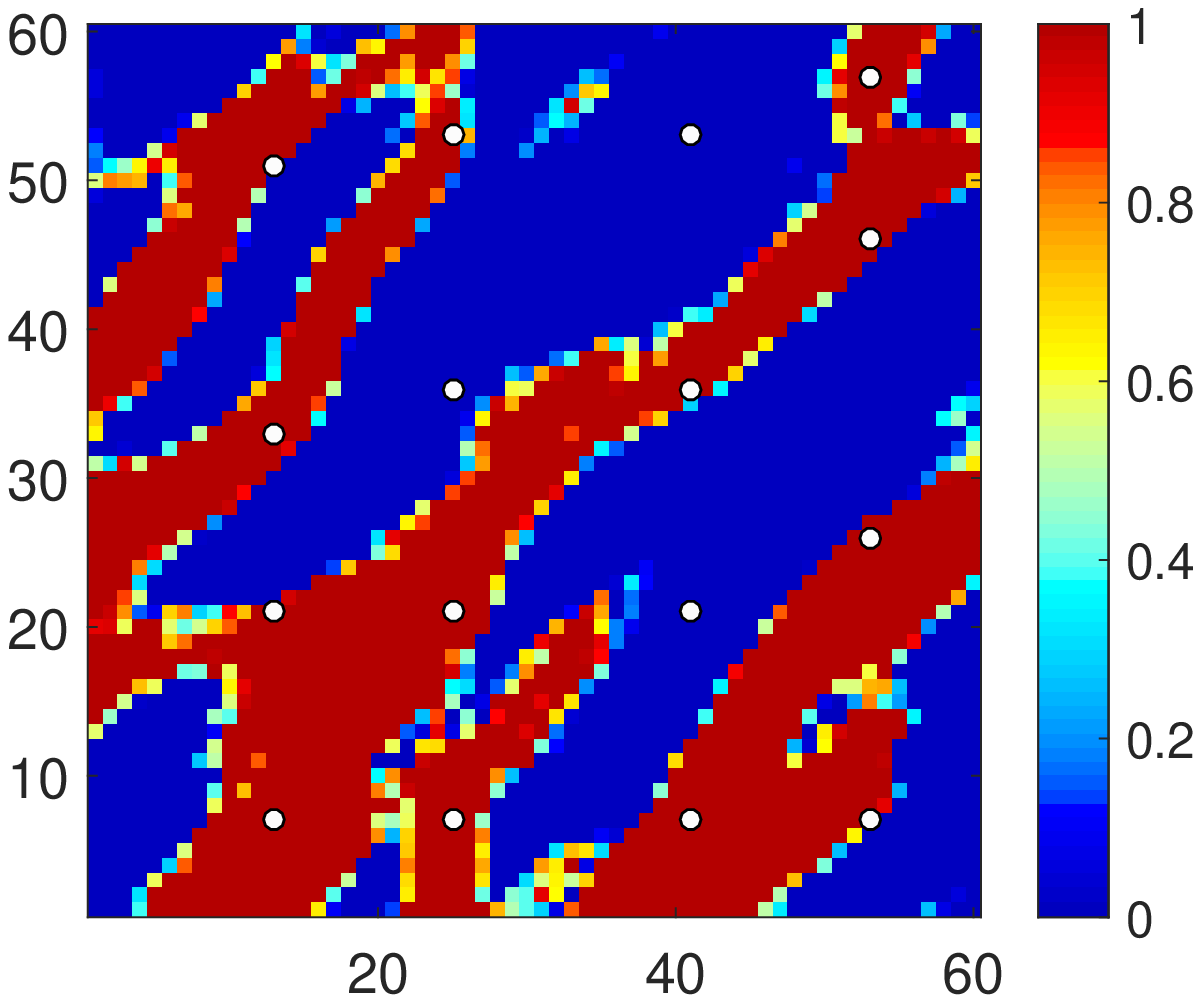}
        \caption{}
    \end{subfigure}%
    ~ 
    \begin{subfigure}[b]{0.32\textwidth}
        \includegraphics[width=1\textwidth]{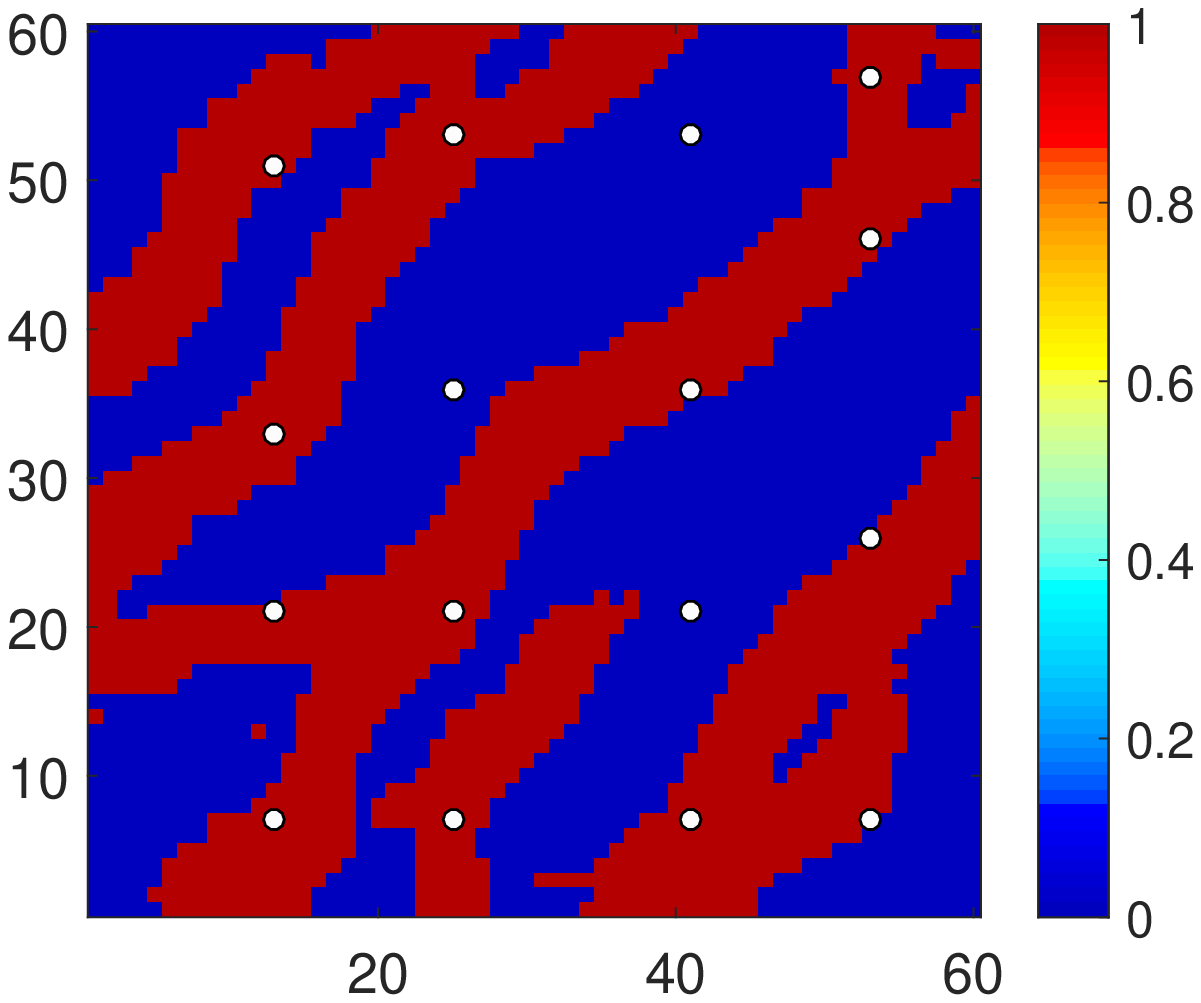}
        \caption{}
    \end{subfigure}%
    \caption{Conditional realizations for the 2D channelized facies model with hard data at 16 wells. \textbf{a, d} Two PCA realizations, \textbf{b, e} two corresponding O-PCA realizations, \textbf{c, f} two corresponding CNN-PCA realizations.}
    \label{fig-cond-res}
\end{figure}

The generation of CNN-PCA realizations conditioned to hard data at well locations (i.e., the facies type at well locations) is now considered. This setup was considered earlier in Sect.~\ref{subsec-opca}. The training image used in Sect.~\ref{subsec-opca} is again used here, and it is also assumed that hard data are available at 16 well locations, as in \cite{Vo2014}. The hard data locations are indicated by the white points in Fig.~\ref{fig-cond-res}. Three wells are located in mud and the other 13 wells are located in sand. The procedures for generating PCA, O-PCA and CNN-PCA models are as described in the previous section, except for two aspects. First, the $\Nr=1000$ SGeMS realizations are now conditional realizations that honor the hard data. Second, when training the model transform net, the hard data loss function (Eqs.~\ref{eq_cnnpca_hd} and \ref{eq_h_def}) is included. The weighting factor $\gamma_h$ is set to be 16 (based on numerical experimentation and the criterion given in Sect.~\ref{sec-hd}).

A total of 200 PCA realizations and corresponding O-PCA and CNN-PCA realizations are generated. As noted in Sect.~\ref{sec-hd}, PCA and O-PCA models always honor hard data. In this case, each of the 200 CNN-PCA models also honors all of the hard data. This demonstrates that the CNN-PCA procedure is able to generate conditional realizations for this binary system, and that the value of $\gamma_h$ is sufficiently large. Figure~\ref{fig-cond-res} displays two PCA realizations and the corresponding O-PCA and CNN-PCA realizations. It is evident that the CNN-PCA realizations do indeed honor facies type at all well locations. Additionally, the CNN-PCA realizations continue to display channel sinuosity and width consistent with the training image. Note finally that the O-PCA realizations in this case display better channel continuity than do those for the unconditioned case (compare Fig.~\ref{fig-cond-res}b,~e to Fig.~\ref{fig-uc-res}b,~e), though the channel geometry in the CNN-PCA realizations is still closer to that in the training image.

\section{Flow Simulation with Random CNN-PCA Realizations}
\label{sec-flow-stats}
In this section, the flow statistics (e.g., P10, P50, P90 responses) associated with an ensemble of CNN-PCA and O-PCA models are compared to those for reference SGeMS models. Comparisons of flow responses between O-PCA and SGeMs models were performed by \cite{Vo2014}. In that study, O-PCA was found to provide flow statistics in reasonable agreement with SGeMS for models with a relatively large amount of conditioning data. However, in the absence of conditioning data, or when such data are sparse, the flow responses for O-PCA models are less accurate. Here a conditional system with a small amount of conditioning data is considered. This is a particularly challenging case since production wells are far from injection wells, so maintaining channel continuity in the geomodel strongly impacts flow response.

\begin{figure}[!htb]
    \centering
    \begin{subfigure}[b]{0.32\textwidth}
        \includegraphics[width=1\textwidth]{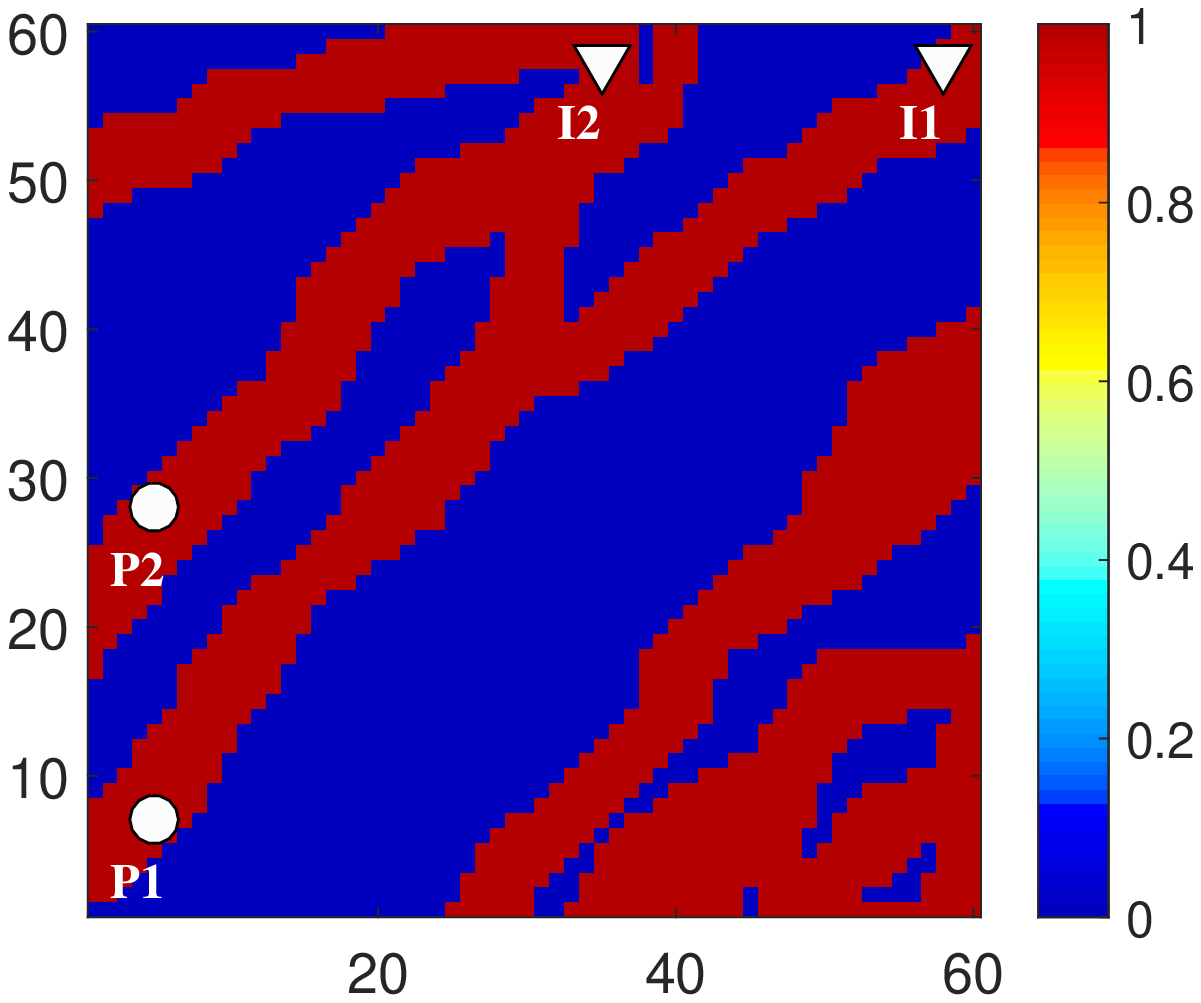}
        \caption{}
    \end{subfigure}%
    ~
    \begin{subfigure}[b]{0.32\textwidth}
        \includegraphics[width=1\textwidth]{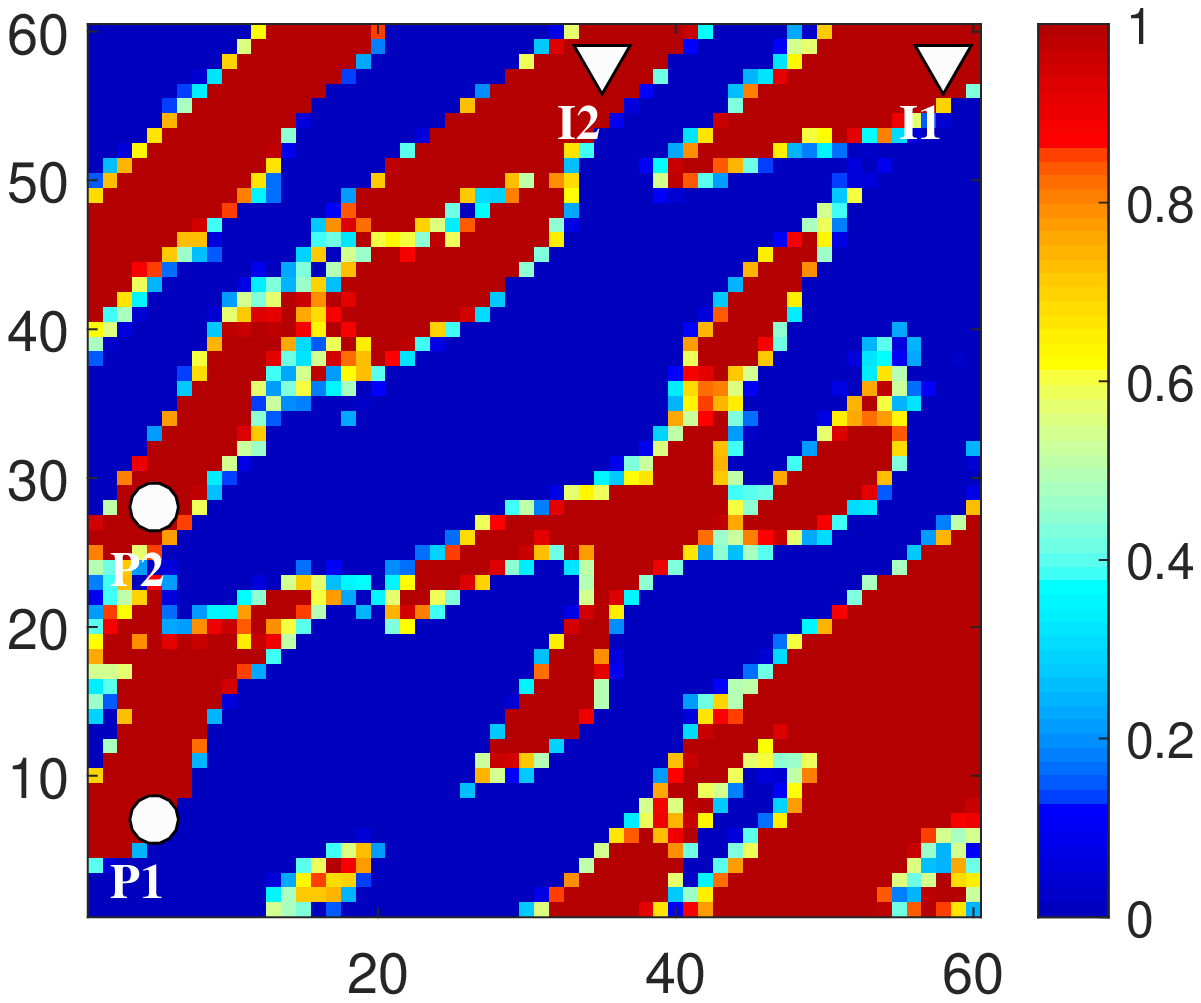}
        \caption{}
    \end{subfigure}%
    ~
    \begin{subfigure}[b]{0.32\textwidth}
        \includegraphics[width=1\textwidth]{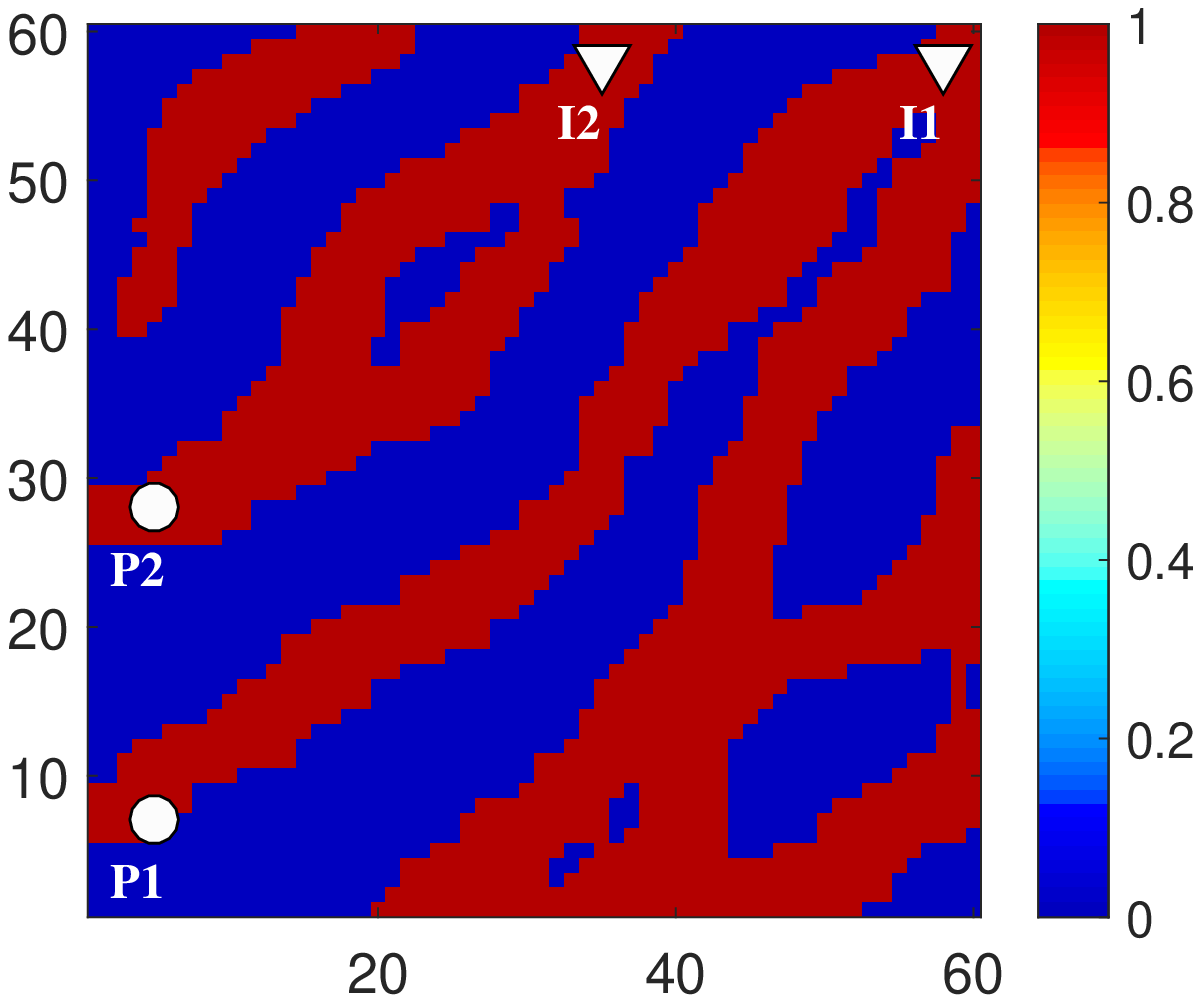}
        \caption{}
    \end{subfigure}%
    \caption{Random realizations conditioned to hard data at four well locations. \textbf{a}~SGeMS realization, \textbf{b}~O-PCA realization, and~\textbf{c}~corresponding CNN-PCA realization.}
    \label{fig-cond-4w-far}
\end{figure}

The reservoir model is defined on a $60\times60$ grid. Grid blocks are of size 50~m in the $x$ and $y$-directions, and 10~m in the $z$-direction. The binary facies model represents a conditional channelized system, with hard data at two injection wells (I1 and I2) and two production wells (P1 and P2), as shown in Fig.~\ref{fig-cond-4w-far}. All four wells are located in sand. The same setup as in Sect.~\ref{Subsect_cnn_con_real} is used to construct the PCA, O-PCA and CNN-PCA models, except the hard data weight $\gamma_h$ for CNN-PCA is 10 in this case. A smaller $\gamma_h$ may be required here than in the earlier example because there are fewer hard data to honor in this case. A total of 200 random PCA realizations and corresponding O-PCA and CNN-PCA realizations are generated. For the SGeMS models, 200 of the 1000 realizations used to construct the PCA representation are selected randomly.

The binary facies model $\Bm$ denotes the grid-block facies type, with $m_i=1$ indicating sand and $m_i=0$ indicating mud in block $i$. The grid-block porosity is specified as $\phi _i = \phi_s m_i+\phi_m(1- m_i)$, where $\phi_s=0.25$ is the sand porosity, and $\phi_m=0.15$ is the mud porosity. Grid-block permeability (taken to be isotropic) is specified as $k_i=a\exp(b m_i)$, where the two constants are $a=2$~md and $b=\ln 1000$. This results in permeability values of 2000~md for sand and 2~md for mud, which corresponds to a strong contrast between facies. Note that this contrast is 10 times more than that in \cite{Vo2014}, where permeabilities of 2000~md for sand and 20~md for mud were used.

Figure~\ref{fig-cond-4w-far} shows one random realization from each method. The SGeMS and CNN-PCA realizations display continuous channels connecting the producers and injectors. In the O-PCA realization (Fig.~\ref{fig-cond-4w-far}b), however, the channels are not as continuous, and I1 is not connected to P1. In the overall set of 200 realizations, the probability of producers being connected to injectors is higher for CNN-PCA and SGeMS models than for O-PCA models. This will be seen to have a strong impact on the flow statistics.

\begin{figure}[!htb]
    \centering
        
        \floatbox[{\capbeside\thisfloatsetup{capbesideposition={left,top},capbesidewidth=6.2cm}}]{figure}[\FBwidth]
{\caption{Oil-water relative permeability curves used for all simulations.}\label{fig-kr}{\hspace{5.5\baselineskip}}}
    {\includegraphics[width=0.4\textwidth]{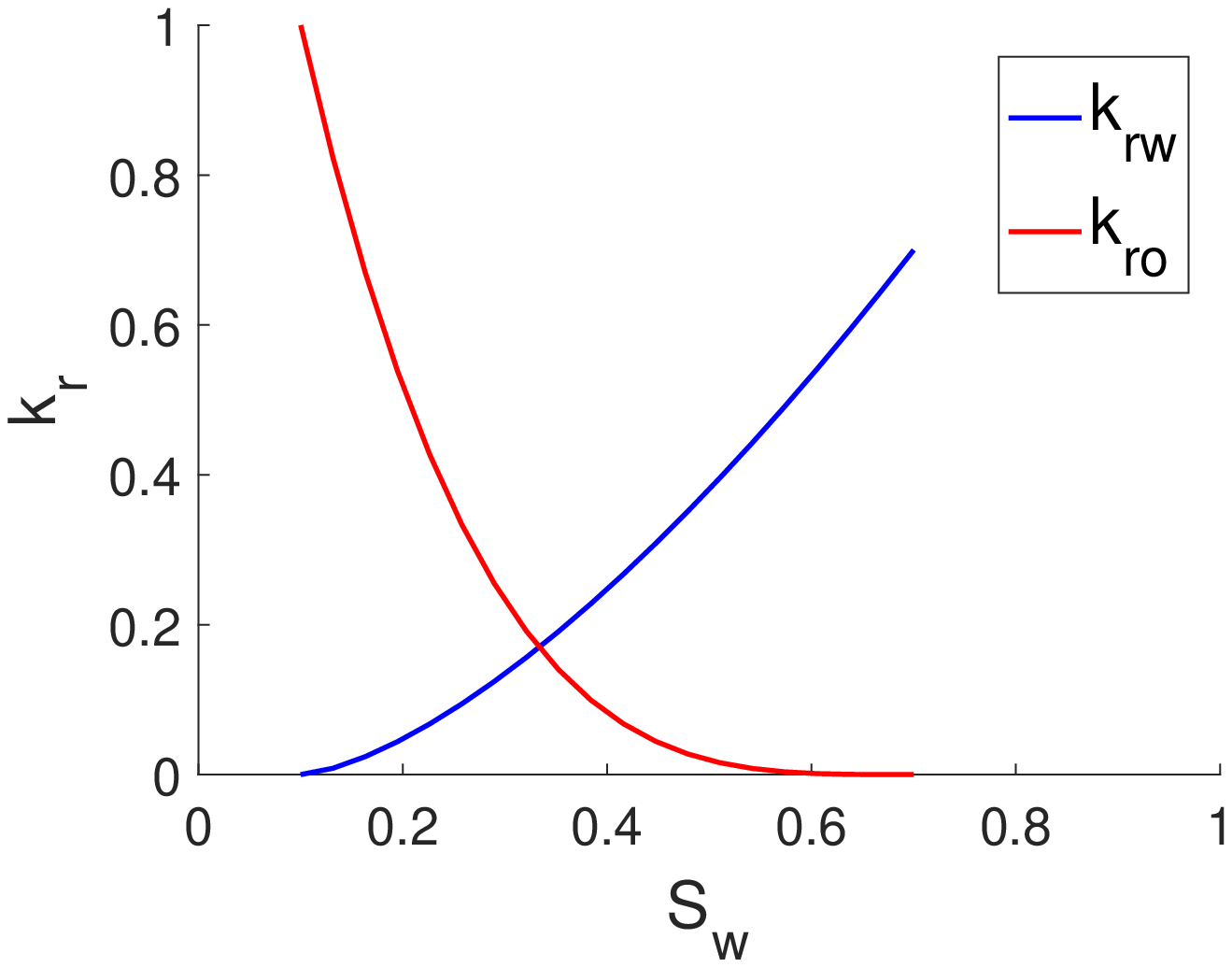}}
\end{figure}

Two-phase oil-water flow is considered in all of the simulations presented in this study. The relative permeability curves are shown in Fig.~\ref{fig-kr}. Initial oil and water saturations are 0.9 and 0.1, respectively. Water viscosity is constant at 0.31~cp. Oil viscosity is 0.29~cp at the initial reservoir pressure of 325~bar. The two water injectors and the two producers operate at constant bottom-hole pressures (BHPs) of 335~bar and 315~bar. The simulation time frame is taken to be 5000~days. This period is sufficiently long such that the producers that are connected to injectors through sand will experience water breakthrough. Note that the use of BHP control for all wells, rather than the specification of injection rates as in \cite{Vo2014}, poses a more stringent test of the models. This is because the prediction quantities of interest here are phase production rates, and these rates can show large errors if channel connectivity is not maintained.


\begin{figure}[!htb]
    \centering
    \begin{subfigure}[b]{0.43\textwidth}
        \includegraphics[width=1\textwidth]{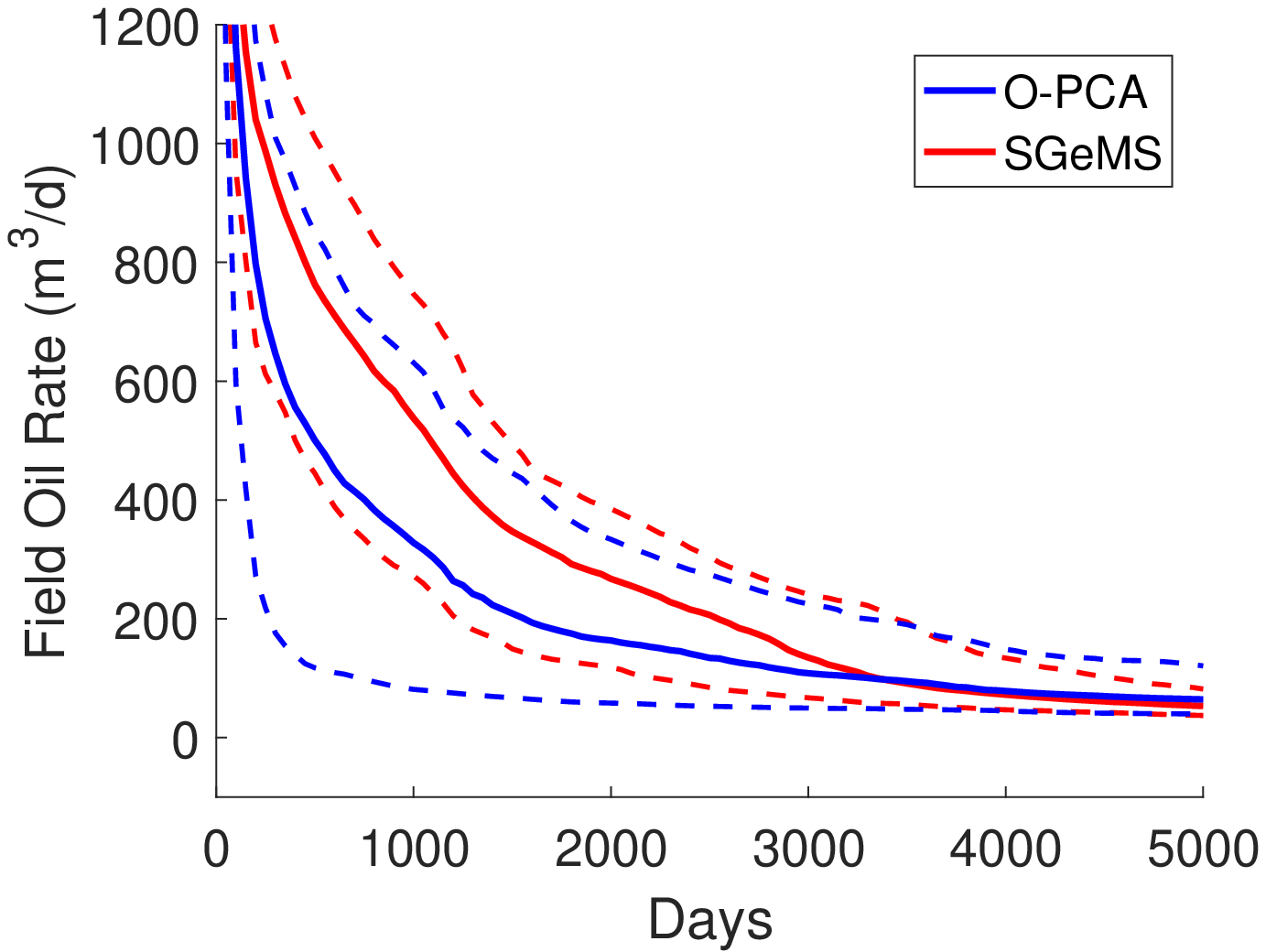}
        \caption{}
    \end{subfigure}%
    \hspace{2\baselineskip}
    \begin{subfigure}[b]{0.43\textwidth}
        \includegraphics[width=1\textwidth]{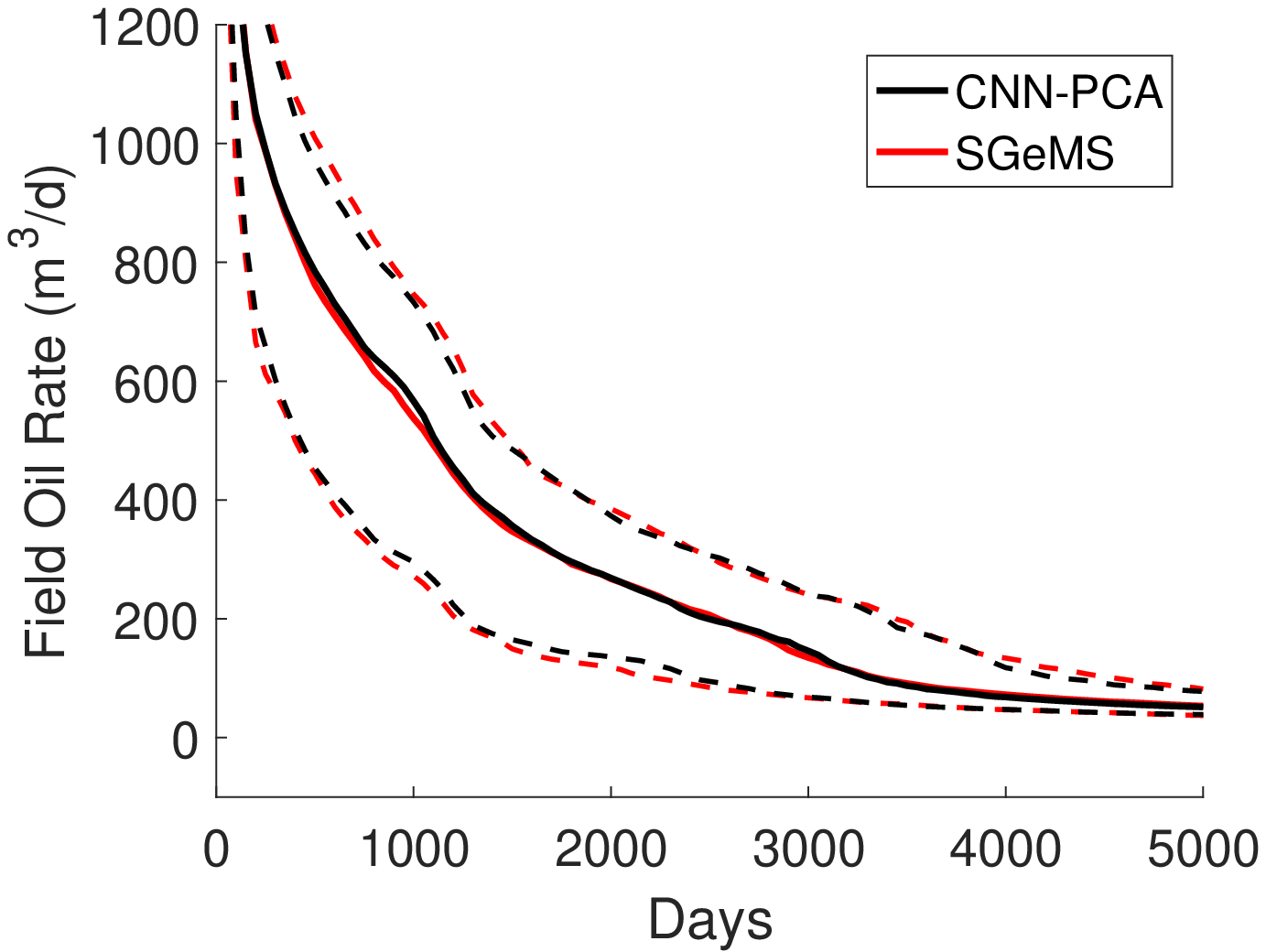}
        \caption{}
    \end{subfigure}%
    
    \begin{subfigure}[b]{0.43\textwidth}
        \includegraphics[width=1\textwidth]{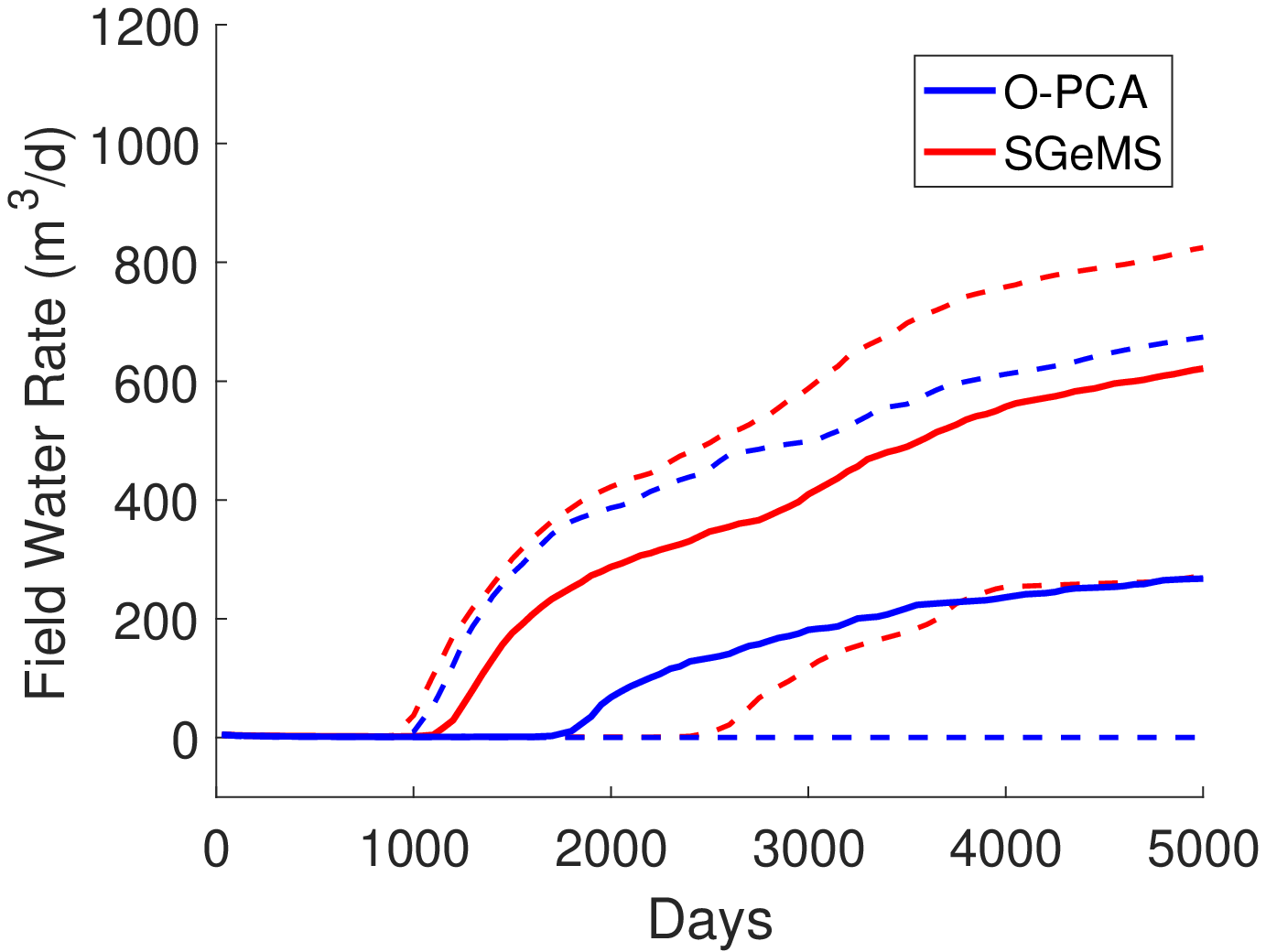}
        \caption{}
    \end{subfigure}%
    \centering
    \hspace{2\baselineskip}
    \begin{subfigure}[b]{0.43\textwidth}
        \includegraphics[width=1\textwidth]{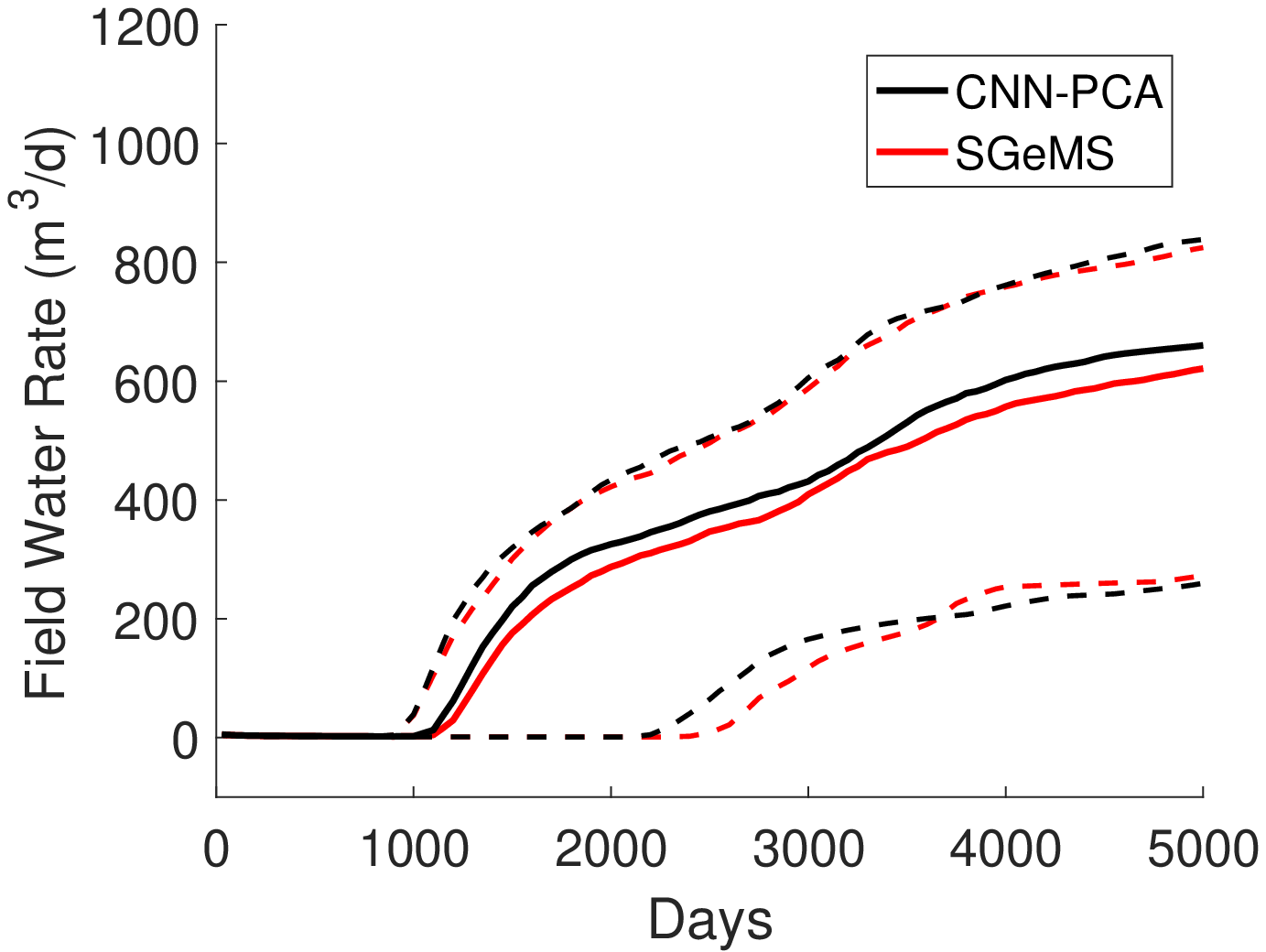}
        \caption{}
    \end{subfigure}%
    
    \caption{P10 (lower set of dashed curves), P50 (solid curves) and P90 (upper set of dashed curves) for oil and water production rates. \textbf{a,~b}~field oil rate and \textbf{c,~d}~field water rate. O-PCA and SGeMS results shown in \textbf{a,~c}, and CNN-PCA and SGeMs results shown in \textbf{b,~d}.}
    \label{fig-flow-stats}
\end{figure}

Figure~\ref{fig-flow-stats} displays the P10, P50 and P90 responses for oil and water production rates obtained from the SGeMS, O-PCA, and CNN-PCA models. The P90 result for a particular quantity (and similarly for P10 and P50 results) at a particular time corresponds to the 90th percentile in the response at that time. At different times, the P90 response is associated with different models. The P50 results are shown in Fig.~\ref{fig-flow-stats} as solid curves, while the P10 and P90 results are shown as dashed curves. 

Figure~\ref{fig-flow-stats}a,~c compares O-PCA results (blue curves) to reference SGeMS results (red curves) for field oil and water rates. Significant differences are evident for this challenging case, and O-PCA results are seen to consistently under-predict oil and water rates. These discrepancies are due to the lack of channel connectivity in many of the O-PCA realizations, and the fact that the wells in these simulations are under BHP control.

CNN-PCA flow responses, shown in Fig.~\ref{fig-flow-stats}b,~d (black curves), are much closer to the reference SGeMS results. Although a slight over-prediction in water rate is evident in Fig.~\ref{fig-flow-stats}d, these results are overall very accurate. With reference to the P50 water rate curves, it is interesting to note that the SGeMS results display two periods of sharper increase in rate. The first such period is at around 1000~days, when the initial breakthrough occurs, while the second is at around 3000~days, corresponding to breakthrough at the second well. This behavior is captured accurately by the CNN-PCA models, indicating that the impact of channel connectivity is correctly captured. The fact that the CNN-PCA P10 and P90 oil and water rate results also agree with the SGeMS results suggests that the variability in the ensemble of 200 models does indeed correspond to that in the SGeMS models. This is an important observation, since it is possible to generate models that are channelized but do not display the appropriate degree of variability (i.e., by specifying a large value for $\gamma_s$, as in Fig.~\ref{fig-weight}c, d).

\begin{figure}[!htb]
    \centering
    \begin{subfigure}[b]{0.43\textwidth}
        \includegraphics[width=1\textwidth]{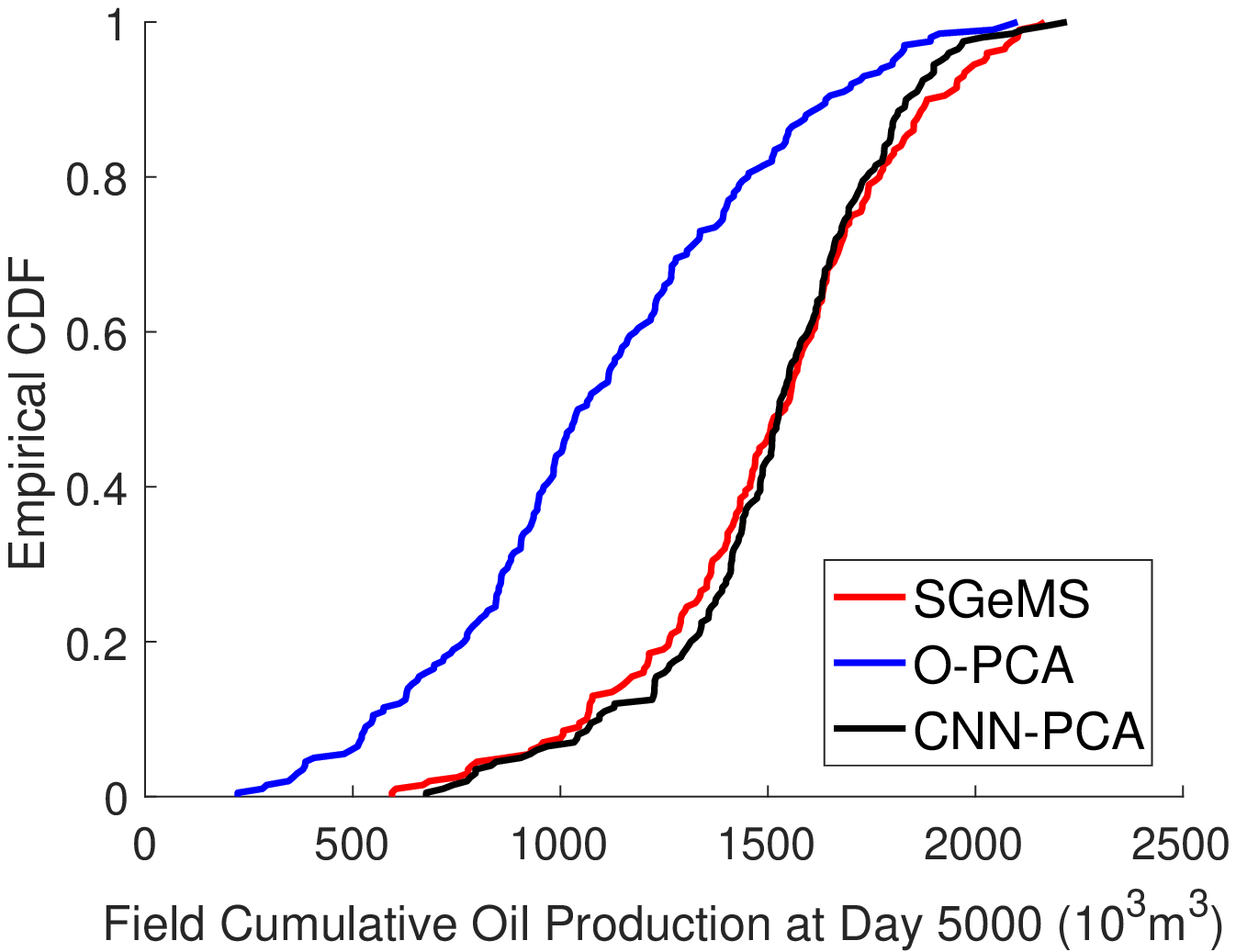}
        \caption{}
    \end{subfigure}%
    \hspace{2\baselineskip}
    \begin{subfigure}[b]{0.43\textwidth}
        \includegraphics[width=1\textwidth]{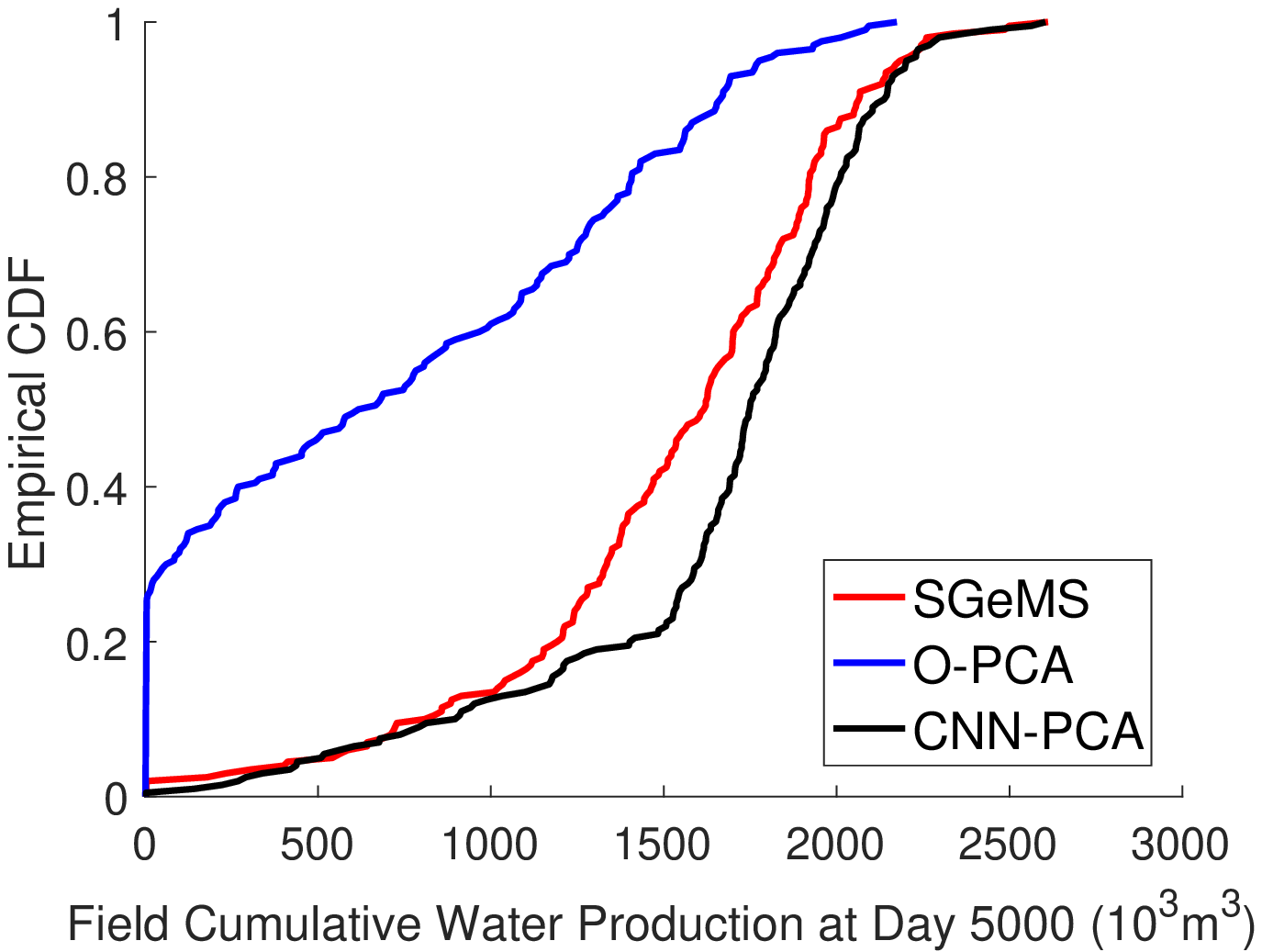}
        \caption{}
    \end{subfigure}%
    \caption{Statistics of cumulative field oil and water production at the end of the simulation from SGeMS models, O-PCA models, and CNN-PCA models. \textbf{a} cumulative oil production CDFs and \textbf{b} cumulative water production CDFs.} 
    \label{fig-flow-stats-cdf}
\end{figure}

Figure~\ref{fig-flow-stats-cdf} presents the cumulative distribution functions (CDFs) for cumulative oil and water produced over the entire simulation time frame (5000~days) for SGeMS, O-PCA and CNN-PCA models. In general CNN-PCA and SGeMS results match closely, though some discrepancy is evident in the water CDFs. The O-PCA models significantly underestimate cumulative production, and about 25\% of the models do not predict any water production at the end of the simulation time frame (Fig.~\ref{fig-flow-stats-cdf}b). This is again due to the lack of channel connectivity in the O-PCA realizations.

We reiterate that this problem setup represents a much more difficult case than those considered in \cite{Vo2014} for several reasons. Specifically, (1) very little conditioning data are used here, (2) the permeability contrast is an order of magnitude larger in the current setup, and (3) all wells are under BHP control in this example. As noted earlier, O-PCA performs well for less extreme cases. For example, with more hard data, channels are better resolved by the underlying PCA representation, leading to more accurate connectivity in O-PCA models. In addition, under rate control, errors in phase flow rates are typically smaller (though BHP errors may then be significant). It is noteworthy, however, that CNN-PCA performs well even for this more extreme case.

\section{History Matching Using CNN-PCA}
\label{sec-hm}
Results in the previous section showed that random CNN-PCA models, generated by sampling the lower-dimensional variable $\Bxi$ from the standard normal distribution, provide flow statistics in close agreement with those from SGeMS models. While SGeMS models are generated from a non-parametric algorithm, the CNN-PCA models are associated with uncorrelated lower-dimensional variables that can be conveniently varied to generate models that match observed production data. The use of CNN-PCA for history matching and posterior uncertainty quantification is now described.

\subsection{History Matching Procedure}

After applying a geological parameterization that maps model parameters $\Bm$ to low-dimensional variables $\Bxi$, the history matching process involves finding the values of $\Bxi$ that minimize the mismatch between simulation results and observed data. The randomized maximum likelihood (RML) method has been shown to be effective at providing reasonably accurate estimates of posterior uncertainty. In this approach, multiple posterior models are obtained by performing multiple minimizations with perturbed objective functions \citep{Kitanidis1986,Oliver1996}.

\cite{Vo2015} provided a detailed description of RML-based subspace history matching for generating multiple posterior samples of $\Bxi$ (and thus $\Bm$). The minimization problem is expressed as
\begin{equation}
\label{Eq_rml}
\Bxirml^* = \argmin{\Bxi} \Big\{ \dfrac{1}{2}\big(\Bd(\Bm(\Bxi))-\Bdobs^*\big)^T\Cdinv\big(\Bd(\Bm(\Bxi))-\Bdobs^*\big) + \dfrac{1}{2}\big(\Bxi-\Bxiuc\big)^T\big(\Bxi-\Bxiuc\big) \Big\}.
\end{equation}
This objective function consists of two terms. The first term, referred to as the data mismatch, is a weighted square mismatch that quantifies the difference between simulated data $\Bd(\Bm(\Bxi))$ and (perturbed) observation data $\Bdobs^*$. Evaluating $\Bd(\Bm(\Bxi))$ involves mapping $\Bxi$ to the model parameter $\Bm$, which is then used to calculate reservoir properties. Simulated flow responses (e.g., well rates and BHPs) at selected time points are then extracted to form $\Bd(\Bm(\Bxi))$. The measurement errors for the observed data are assumed to be independent and to follow a multivariate Gaussian distribution with zero mean and diagonal covariance matrix $\Cd$. In RML, the observed data are perturbed with random noise sampled from $N({\bf 0},\Cd)$. 

The second term, referred to as the model mismatch, is a regularization term that acts to constrain $\Bxi$ to be close to a random prior realization $\Bxiuc$ sampled from the standard normal distribution. Multiple minimizations are performed, each with different random realizations of $\Bdobs^*$ and $\Bxiuc$, to obtain multiple posterior models. Simulation results obtained from these posterior models are used to estimate the posterior uncertainty of the production forecast.

\subsection{Problem Setup}
\begin{figure}[htb!]
  \centering
  \begin{tabular}{ccc}
    \multirow{2}{*}[1pt]{
    \begin{subfigure}[c]{0.31\textwidth}
      \includegraphics[width=\textwidth]{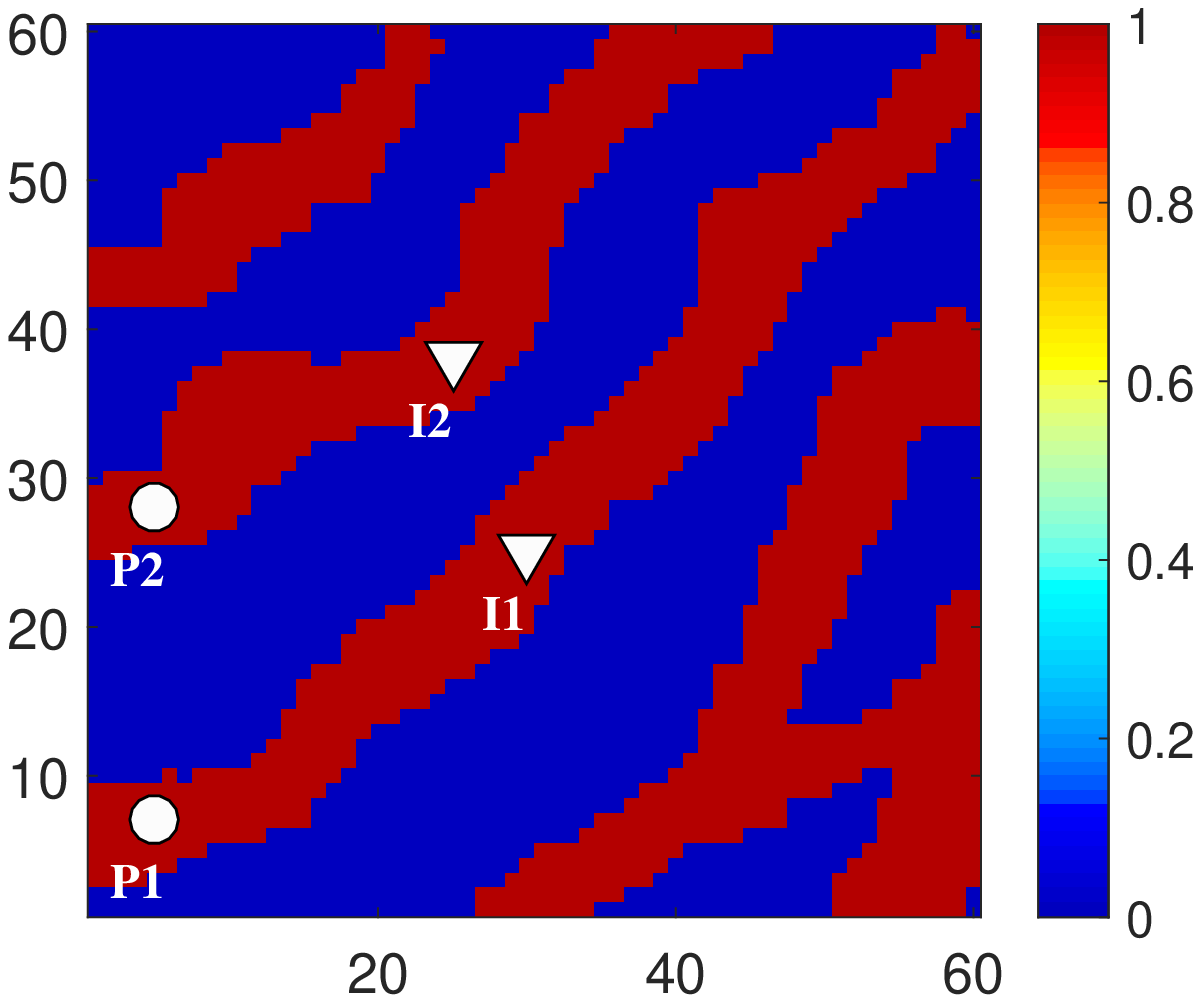}
      \caption{}
    \end{subfigure}
}&
   \begin{subfigure}[c]{0.31\textwidth}
      \includegraphics[width=\textwidth]{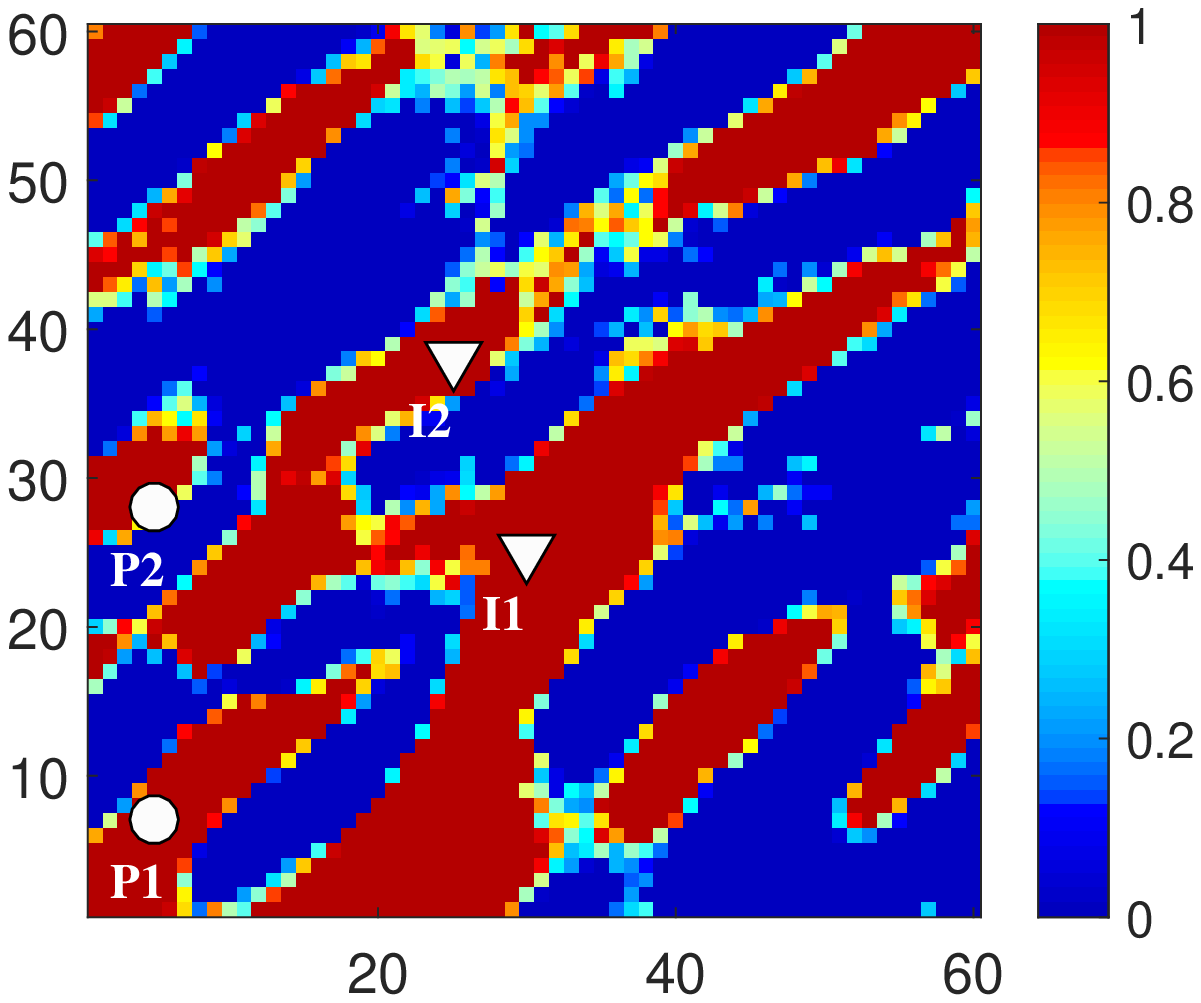} 
      \caption{}
    \end{subfigure}&
    \begin{subfigure}[c]{0.31\textwidth}
      \includegraphics[width=\textwidth]{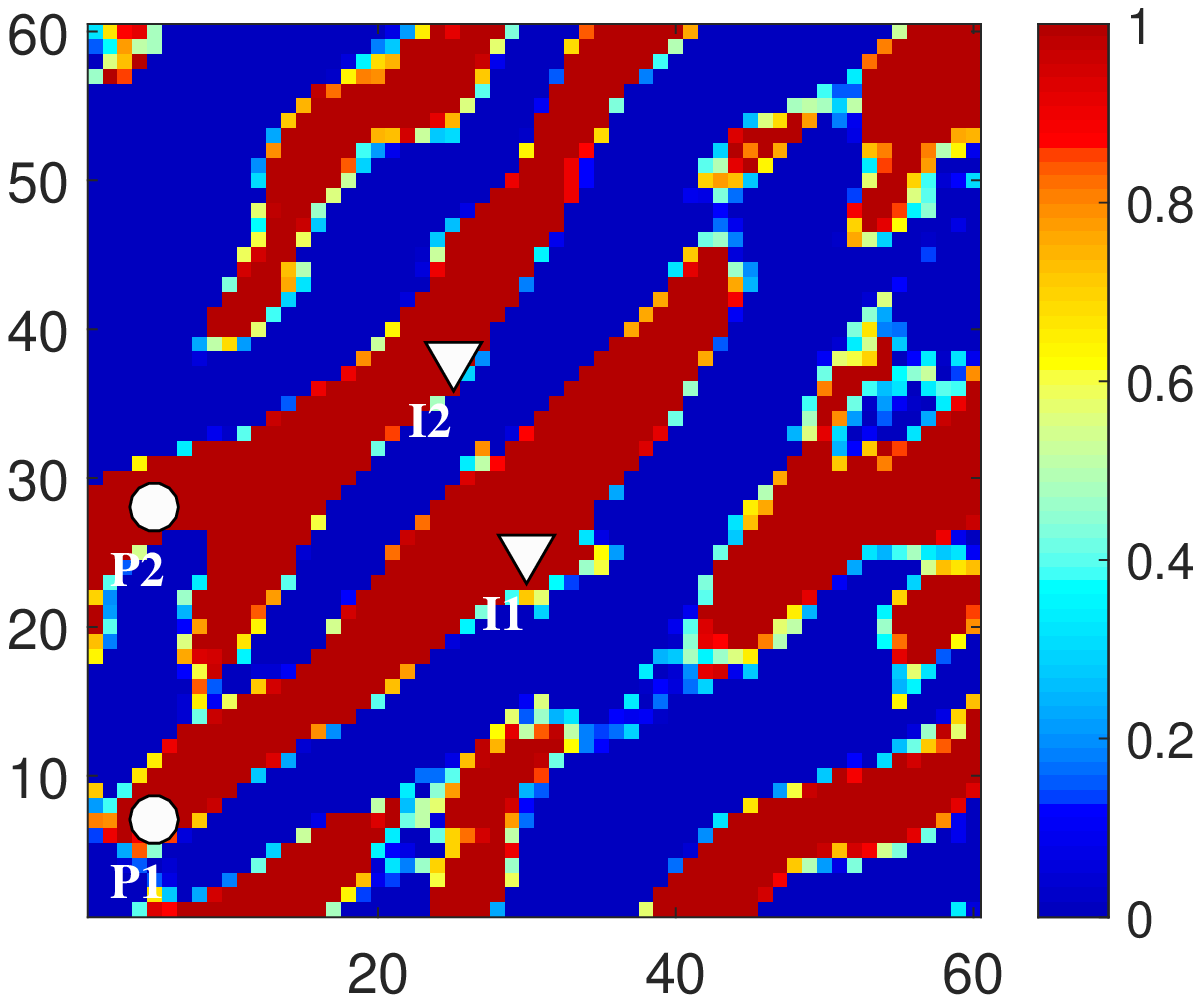}
      \caption{}
    \end{subfigure}
    \\
    &
     \begin{subfigure}[c]{0.31\textwidth}
      \includegraphics[width=\textwidth]{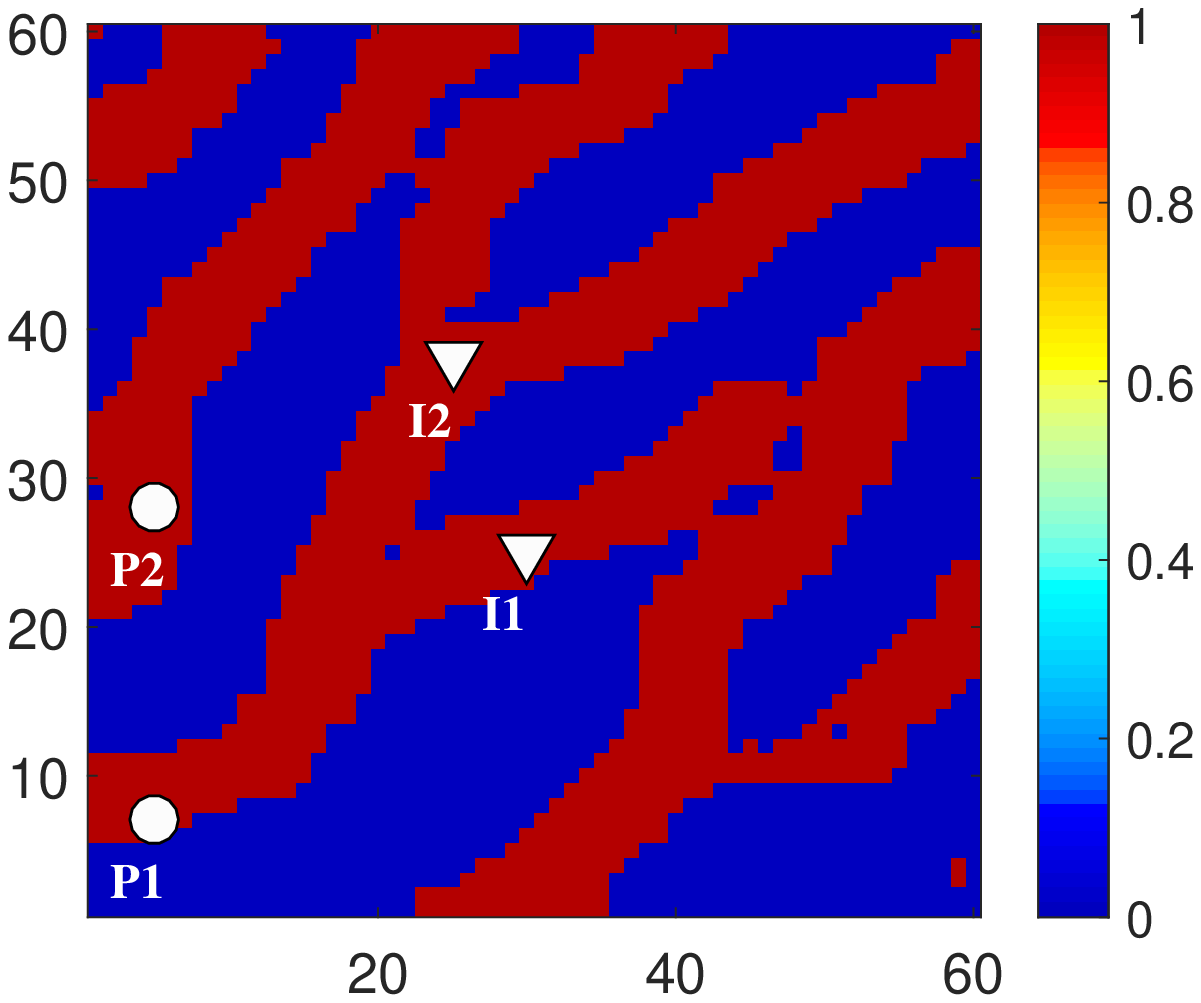}
      \caption{}
    \end{subfigure}&
    \begin{subfigure}[c]{0.31\textwidth}
      \includegraphics[width=\textwidth]{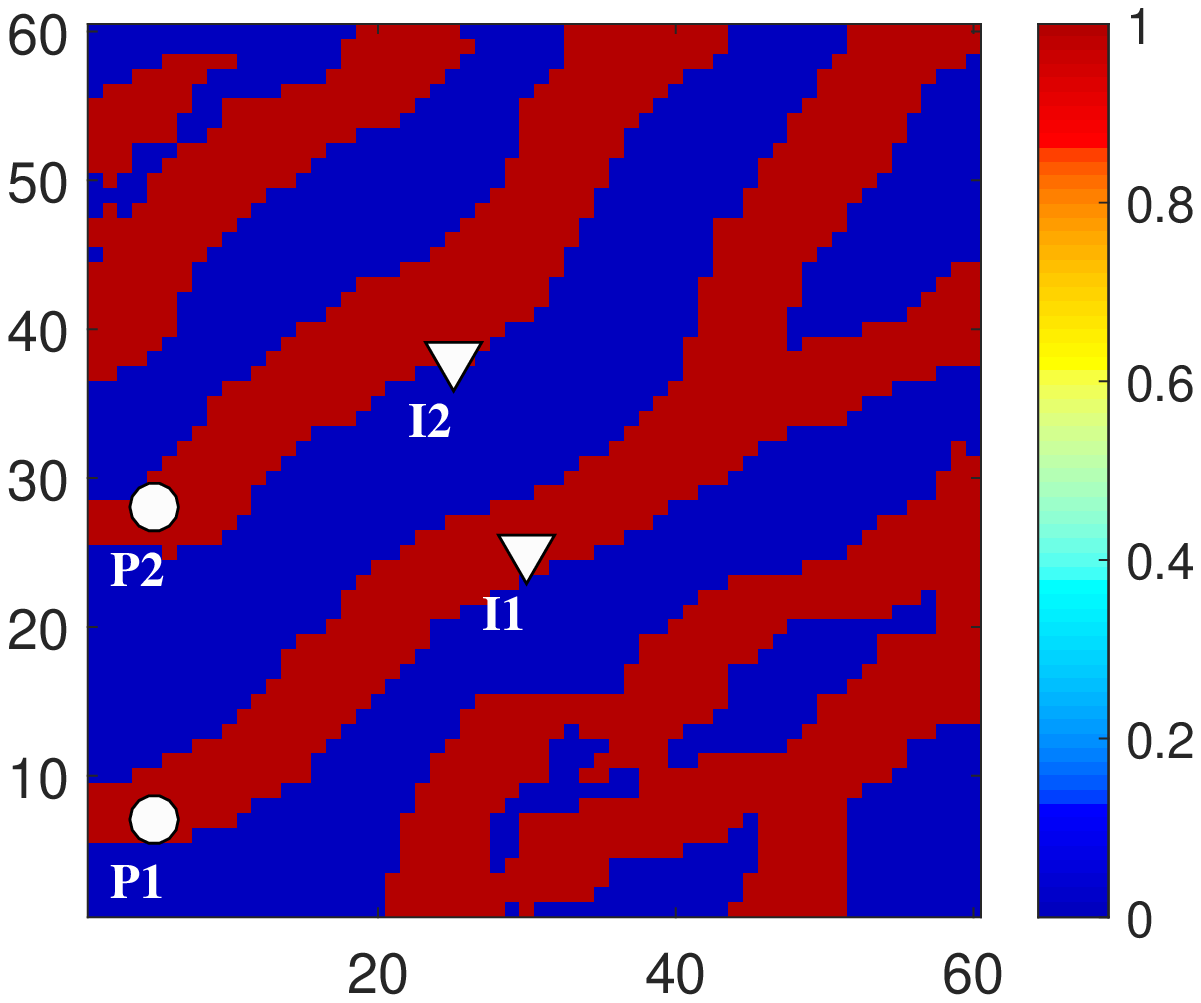}
      \caption{}
    \end{subfigure} 
  \end{tabular}    
  \caption{Facies models for the conditional 2D channelized reservoir with hard data at four wells. \textbf{a} True SGeMS model, \textbf{b} one O-PCA prior model, \textbf{c} corresponding O-PCA posterior model, \textbf{d} one CNN-PCA prior model, \textbf{e} corresponding CNN-PCA posterior model.}
  \label{fig-hm-logk}
\end{figure}

The setup used for history matching is very similar to that considered for flow assessment in Sect.~\ref{sec-flow-stats}. The reservoir model is again a 2D channelized system defined on a $60 \times 60$ grid. Here, however, the two injection wells are located around the center of the model, as shown in Fig.~\ref{fig-hm-logk}, and the simulation time frame is only 2000~days. Injection and production well BHPs are set to 335~bar and 315~bar, respectively. In a later stage, new production wells will be introduced in the upper-right corner of the model. Predictions using both CNN-PCA posterior models and O-PCA posterior models, for both existing wells and new wells, will be presented.

The setup for constructing conditional O-PCA and CNN-PCA representations is the same as described previously. A total of $\Nr=1000$ conditional SGeMS realizations are generated to construct the PCA representation, with reduced dimension of $l=70$. The SGeMS realizations are conditioned to hard data at the four wells, all of which are in sand. One SGeMS realization is randomly selected as the `true' facies model (Fig.~\ref{fig-hm-logk}a). This realization is not included in the models used to construct the PCA basis. In constructing CNN-PCA, $\Nt=3000$ random PCA realizations are used to train the model transform net. The weighting factors in Eq.~\ref{eq_cnnpca_hd} are $\gamma=0.3$ and $\gamma_h=4$. A lower $\gamma_h$ value may be required here than in Sect.~\ref{sec-flow-stats} since the wells are closer (and thus the hard data tend to be honored more easily).

Parameters for the flow simulations are as described in Sect.~\ref{sec-flow-stats}. The history-match parameters (components of $\Bxi$) provide $m_i$, $i=1, \ldots, \Nc$, from which grid-block properties are computed as $\phi _i = \phi_s m_i+\phi_m(1- m_i)$ and $k_i=a\exp(b m_i)$, with constants as given in Sect.~\ref{sec-flow-stats}. Observed data include water injection rate and oil and water production rates over the first 1000~days, recorded every 100~days. The total number of data points is $\Nd=60$. Observed data are generated using simulation results for the true model (Fig.~\ref{fig-hm-logk}a), perturbed with independent random noise following a Gaussian distributions with zero mean. The standard deviation for the random noise is set to be 10\% of the corresponding true data, with a minimum value of 2~m$^3/$day. A total of 30 posterior (history matched) models are generated to quantify uncertainty using the subspace RML procedure described in Eq.~\ref{Eq_rml}.

A derivative-free optimization method is applied here to solve the minimization problem defined in Eq.~\ref{Eq_rml}. This type of approach is non-intrusive with respect to the simulator, so it can be readily used with any subsurface flow simulator. As discussed in Sect.~\ref{Subsect_cnn_pca}, the CNN-PCA representation is differentiable when the final hard-thresholding step is replaced with a differentiable transformation. The algorithm can then be applied in conjunction with efficient adjoint-gradient-based methods, as was done with O-PCA in \cite{Vo2015}.

The derivative-free optimizer used here is the hybrid PSO--MADS algorithm developed by \citep{Isebor2014}. This approach combines particle swarm optimization (PSO), which is a global stochastic search algorithm, with mesh adaptive direct search (MADS), a local pattern search method. By alternating between the two methods, PSO--MADS provides some amount of global exploration along with convergence to a local minimum. This was found to be useful in field development optimization problems. Standalone PSO and standalone MADS have been shown to be effective in history matching with PCA \citep{Echeverria2009,Rwechungura2011} and O-PCA \citep{Liu2017} parameterizations. The application of PSO--MADS for history matching, which was found to be effective for the cases considered in this study, does not appear to have been reported previously.

In the example considered here, the dimension of the search space ($l$) is 70. A PSO swarm size of 50 is used, which means that 50 flow simulations are required at each PSO iteration. Each MADS iteration involves evaluation of a set of stencil-based trial points located around the current-best solution. This entails $2l=140$ flow simulations for the MADS version applied here. Both PSO and MADS parallelize naturally, so if 140 compute nodes are available, the algorithm can be run fully parallelized. The initial guess for the minimization in Eq.~\ref{Eq_rml} is set to be a random prior realization $\Bxiuc$. PSO--MADS is run in all cases for 24 iterations, which was found to provide an appropriate level of reduction in the mismatch defined in Eq.~\ref{Eq_rml}. Detailed optimizer settings and treatments were not explored here since the intent is to evaluate the general performance of CNN-PCA for history matching. Future work should assess both alternative PSO--MADS specifications as well as the use of adjoint-gradient-based minimization.

\subsection{History Matching Results} 
Examples of prior and posterior (history-matched) O-PCA and CNN-PCA models are presented in Fig.~\ref{fig-hm-logk}, along with the true model. The posterior models correspond to the prior models, meaning the $\Bxiuc$ vector associated with the prior model is used as the initial guess in the optimization. It is evident that, in the true model (Fig.~\ref{fig-hm-logk}a), high-permeability channels connect injector-producer pairs I1--P1 and I2--P2. These connectivities are not fully captured in the prior models (Fig.~\ref{fig-hm-logk}b,~d), but both the O-PCA and CNN-PCA posterior models (Fig.~\ref{fig-hm-logk}c,~e) resolve these aspects of the system.

\begin{figure}[!htb]
    \centering
    \begin{subfigure}[b]{0.43\textwidth}
        \includegraphics[width=1\textwidth]{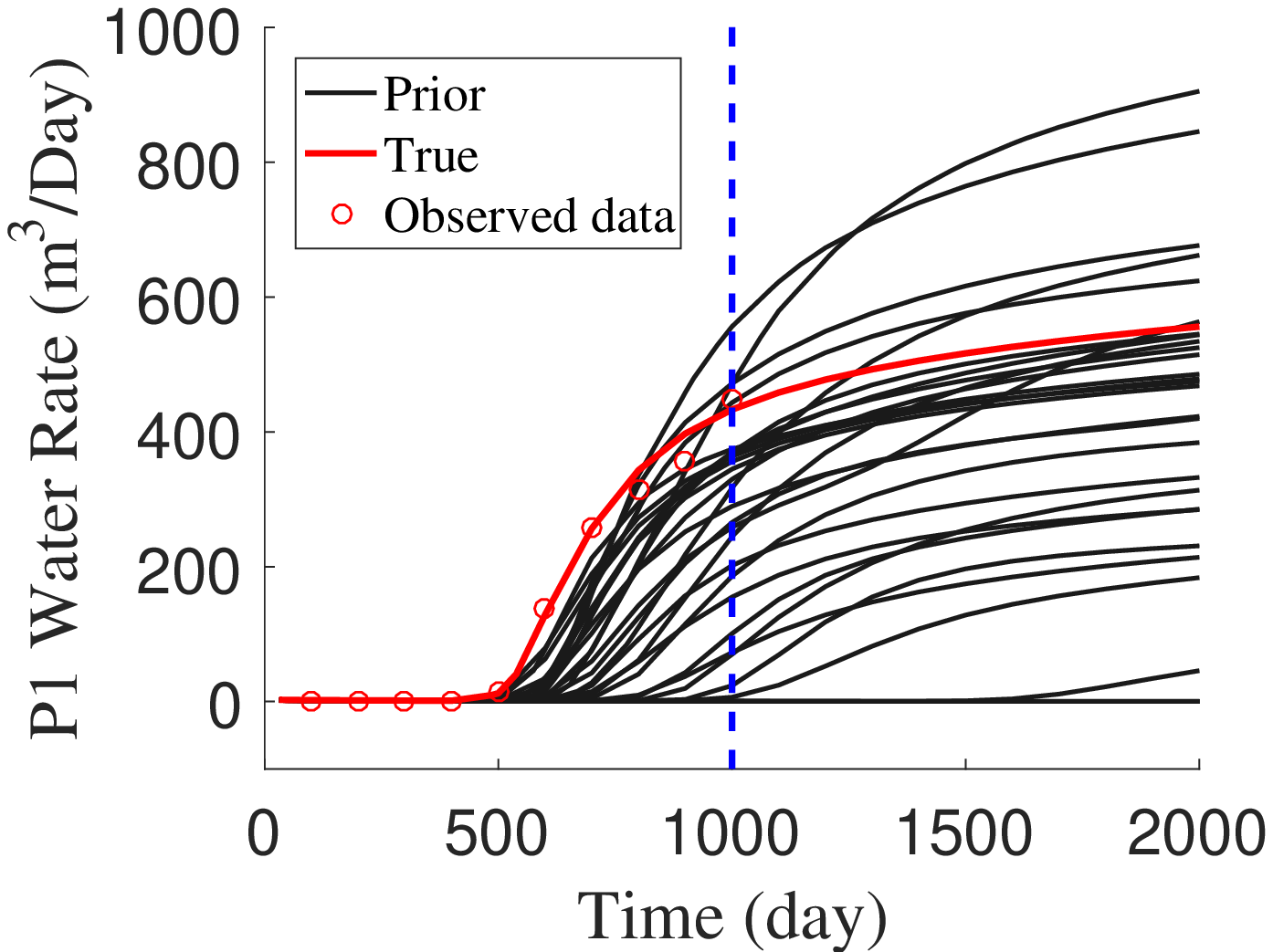}
        \caption{}
    \end{subfigure}%
    \hspace{2\baselineskip}
    \begin{subfigure}[b]{0.43\textwidth}
        \includegraphics[width=1\textwidth]{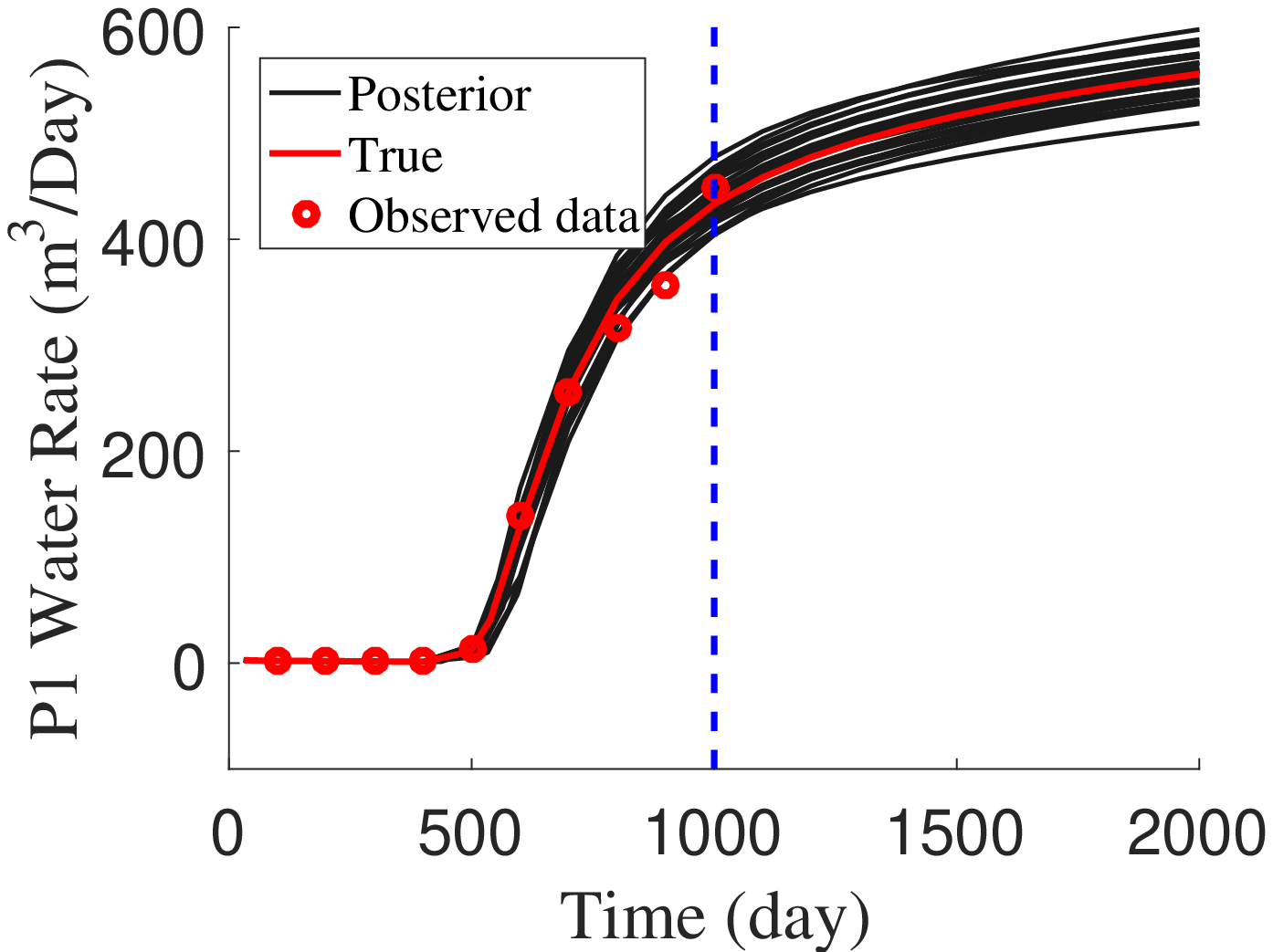}
        \caption{}
    \end{subfigure}%
    
    \begin{subfigure}[b]{0.43\textwidth}
        \includegraphics[width=1\textwidth]{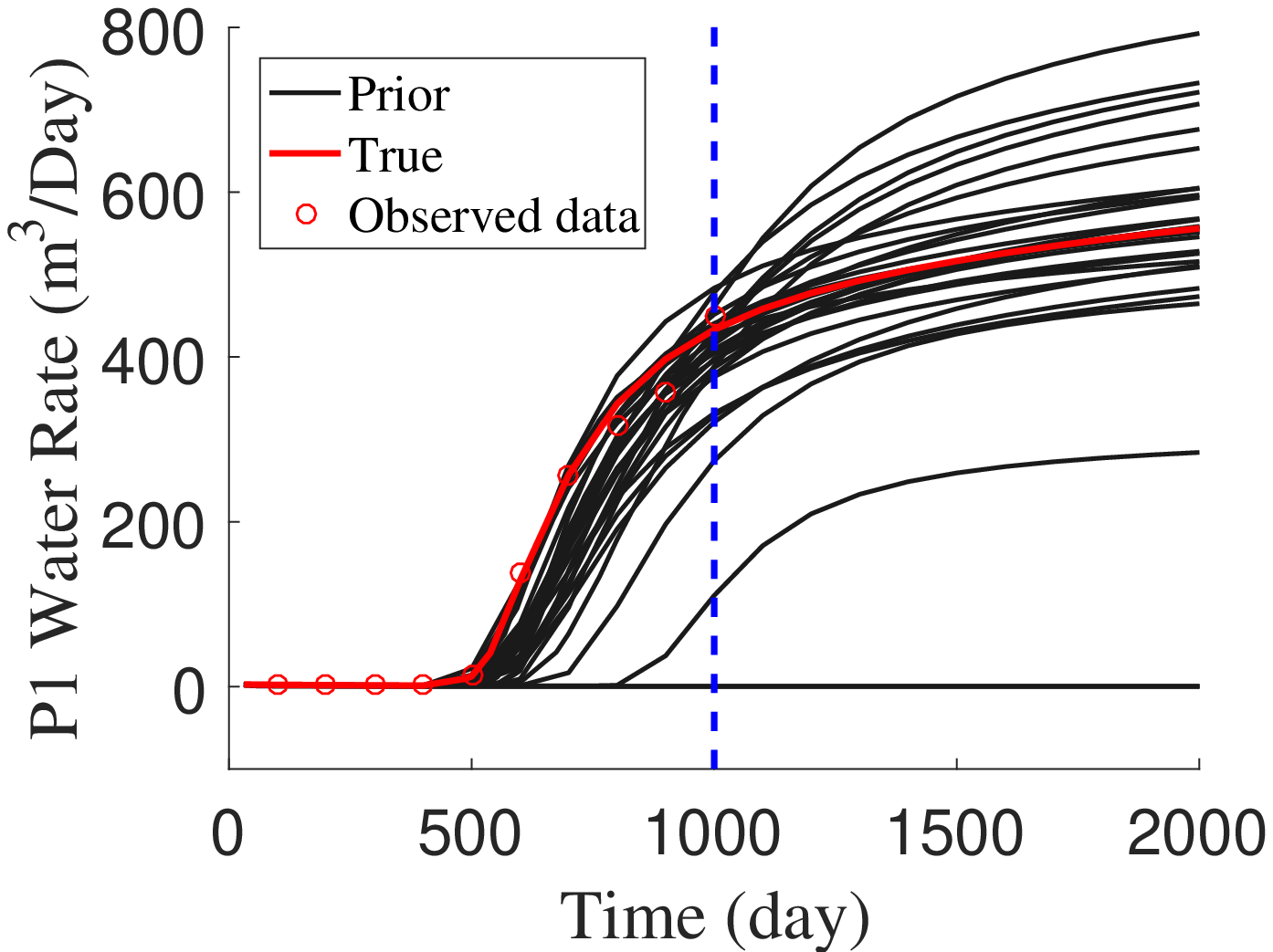}
        \caption{}
    \end{subfigure}%
    ~ 
    \hspace{2\baselineskip}
    \begin{subfigure}[b]{0.43\textwidth}
        \includegraphics[width=1\textwidth]{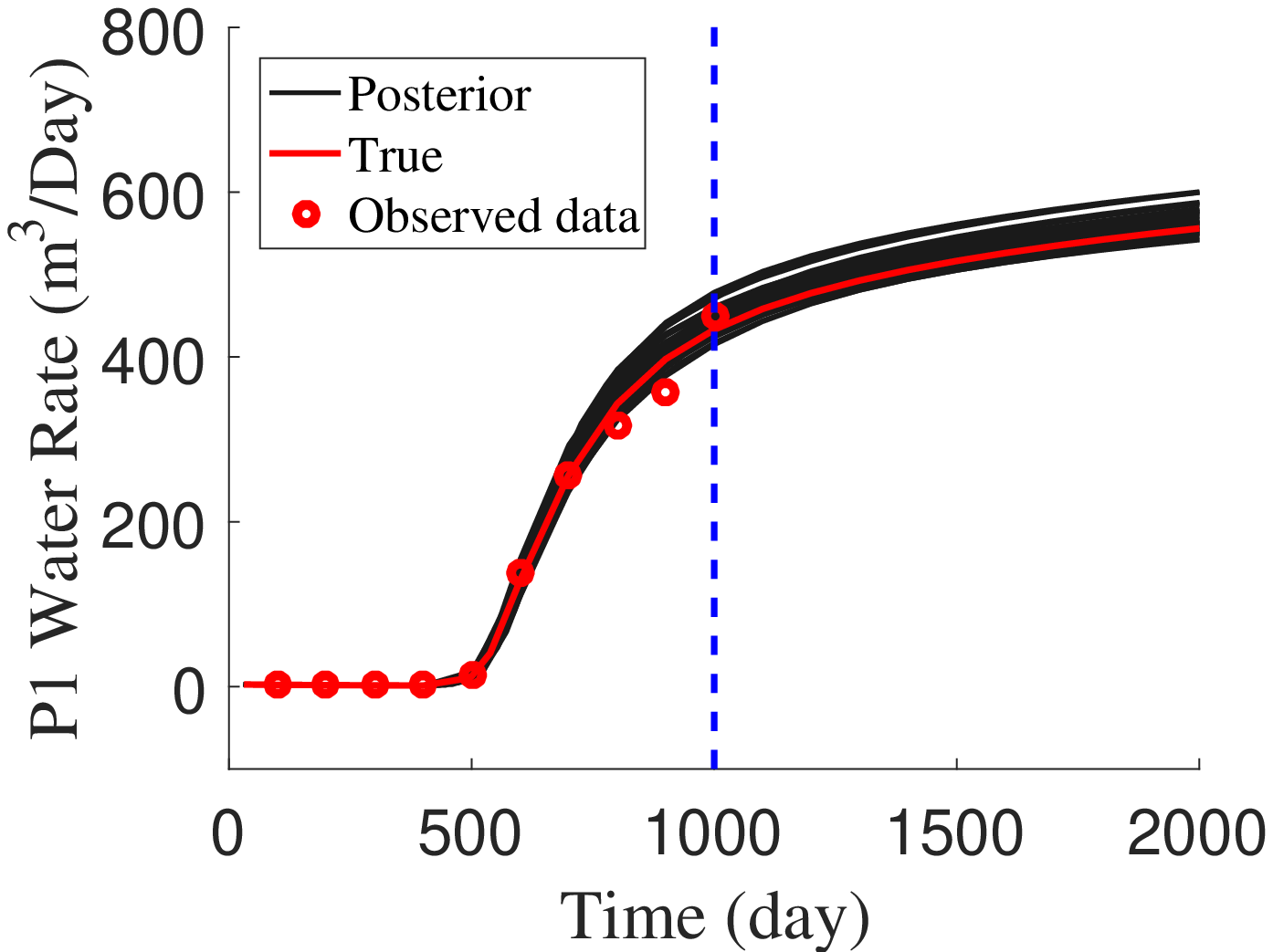}
        \caption{}
    \end{subfigure}%
    
    \caption{Prior and posterior predictions for P1 water production rates. \textbf{a} O-PCA prior predictions, \textbf{b} O-PCA posterior predictions, \textbf{c} CNN-PCA prior predictions, \textbf{d} CNN-PCA posterior predictions. Vertical dashed line at 1000~days indicates end of history matching period.}
    \label{fig-hm-prod}
\end{figure}

Figure~\ref{fig-hm-prod} displays water production results for producer P1. The vertical dashed line at 1000~days indicates the end of the history matching period. Prior O-PCA and CNN-PCA results are shown in Fig.~\ref{fig-hm-prod}a,~c, and posterior results in Fig.~\ref{fig-hm-prod}b,~d. It is evident that, even though the O-PCA prior results suggest that channel connectivity is insufficient in many of the realizations (this inference derives from the fact that water rate is quite low in some cases), posterior predictions from O-PCA are close to the true data. With CNN-PCA, the prior predictions suggest better channel connectivity in the prior models. Posterior CNN-PCA predictions, like those of O-PCA, are close to the true data. Both methods result in much smaller spreads in posterior predictions compared to the prior results. Similar behaviors are observed for well P2. For this assessment it is evident that, although CNN-PCA provides higher-quality posterior models than O-PCA in terms of geological realism, there is not much difference in posterior flow predictions.

\begin{figure}[!htb]
    \centering
    \floatbox[{\capbeside\thisfloatsetup{capbesideposition={left,top},capbesidewidth=8.5cm}}]{figure}[\FBwidth]
{\caption{True facies model and locations of new wells P3 and P4.}\label{fig-true-model-iw}{\hspace{2.2\baselineskip}}}
    {\includegraphics[width=0.35\textwidth]{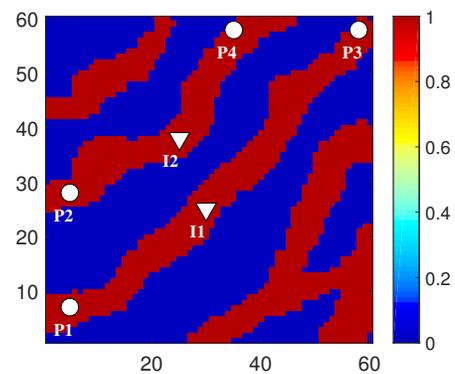}}
    
\end{figure}

A more challenging type of posterior prediction is now considered. In this assessment, the performance of posterior models in predicting production for new wells located away from existing wells is evaluated. Figure~\ref{fig-true-model-iw} displays the problem setup. The two new wells, P3 and P4, are in the upper-right region of the model. Based on the true facies model, wells P3 and P4 are connected to I1 and I2 through sand channels. Both new wells begin production at the end of the history matching period (1000~days), with BHPs specified to be 315~bar. Again, the posterior models are simulated to 2000~days. 

\begin{figure}[!htb]
    \centering
    \begin{subfigure}[b]{0.32\textwidth}
        \includegraphics[width=1\textwidth]{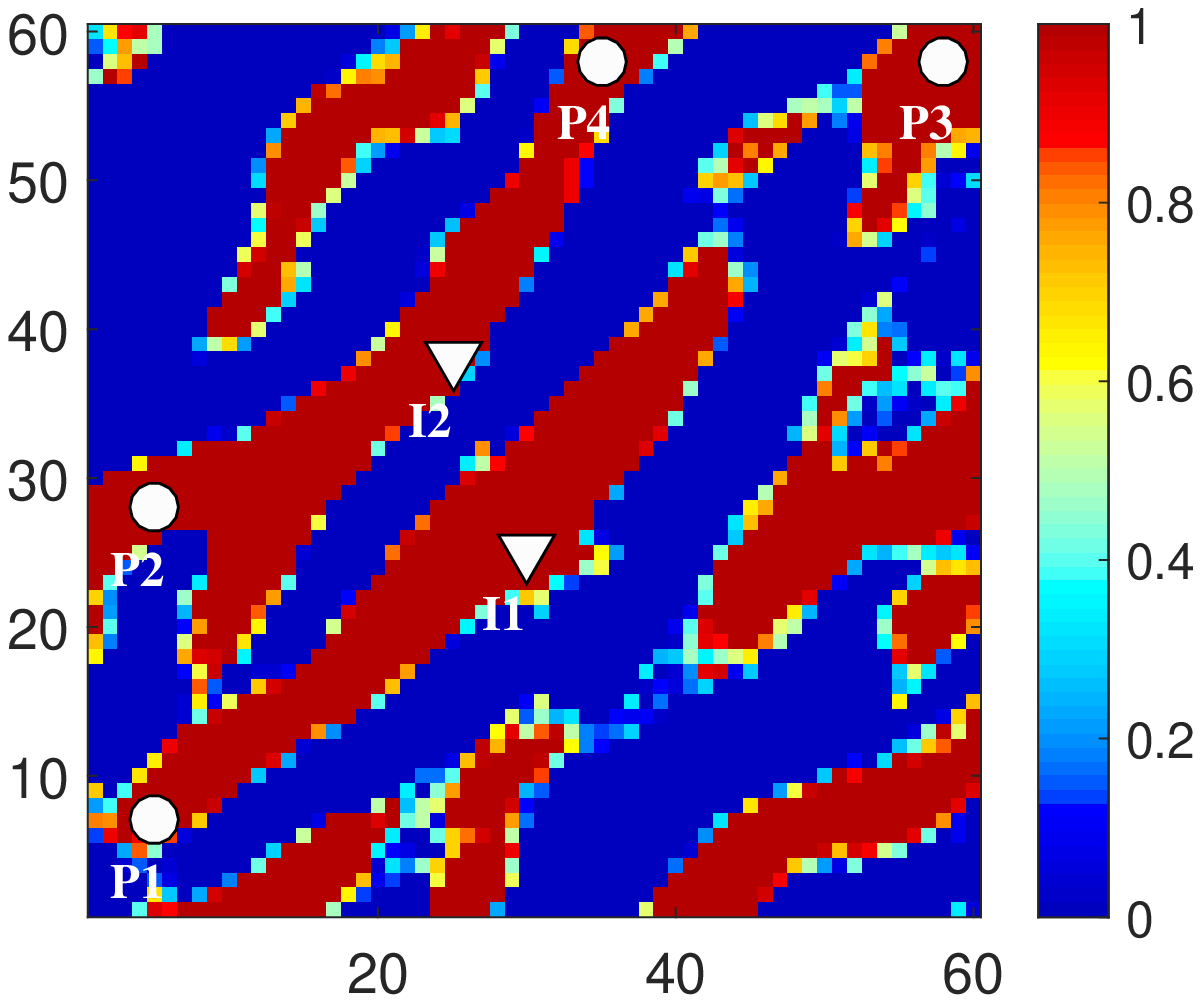}
        \caption{}
    \end{subfigure}%
    \hspace{0.1\baselineskip}
    \begin{subfigure}[b]{0.32\textwidth}
        \includegraphics[width=1\textwidth]{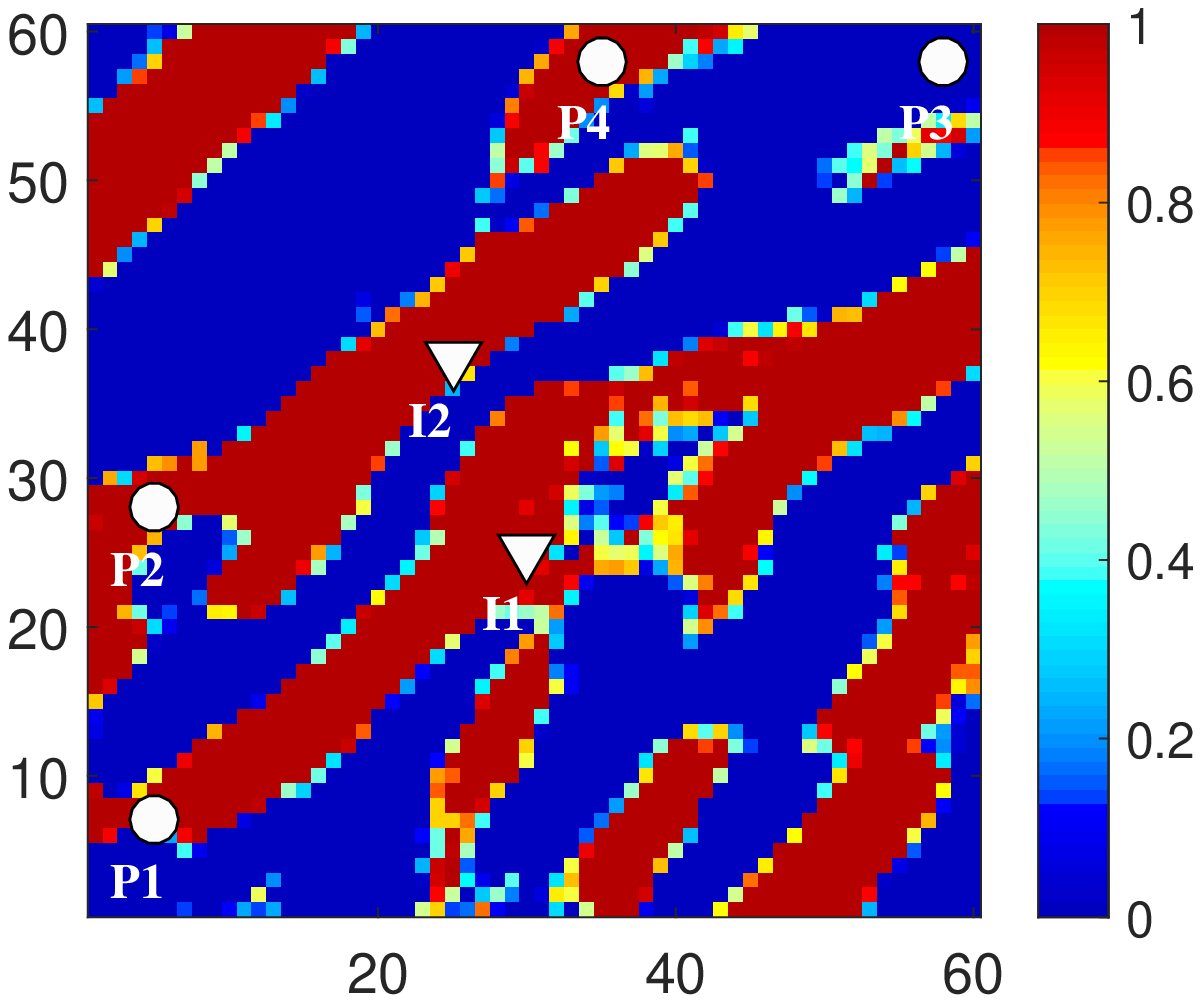}
        \caption{}
    \end{subfigure}%
    \hspace{0.1\baselineskip}
    \begin{subfigure}[b]{0.32\textwidth}
        \includegraphics[width=1\textwidth]{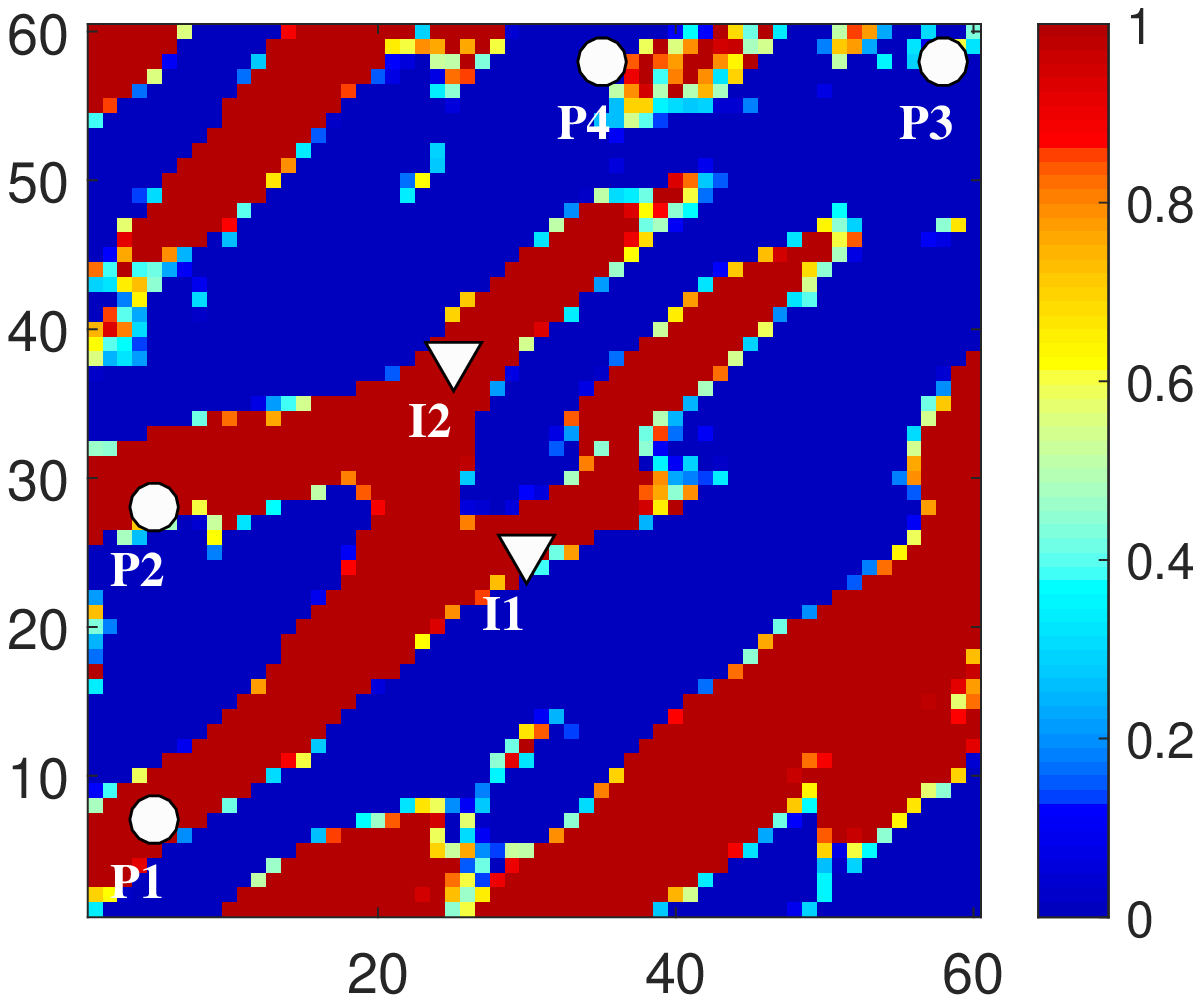}
        \caption{}
    \end{subfigure}%
    
    \begin{subfigure}[b]{0.32\textwidth}
        \includegraphics[width=1\textwidth]{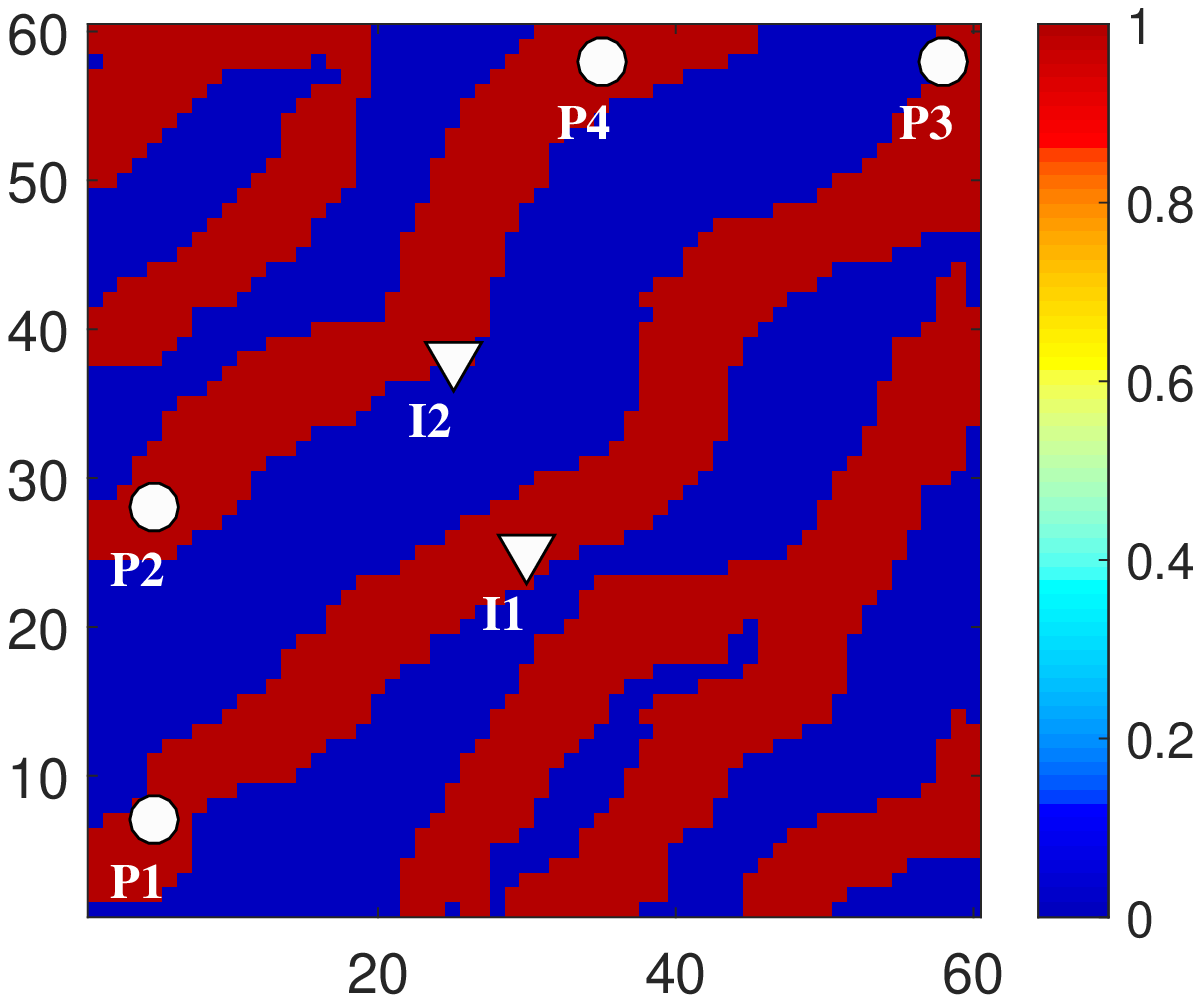}
        \caption{}
    \end{subfigure}%
    \hspace{0.1\baselineskip}
    \begin{subfigure}[b]{0.32\textwidth}
        \includegraphics[width=1\textwidth]{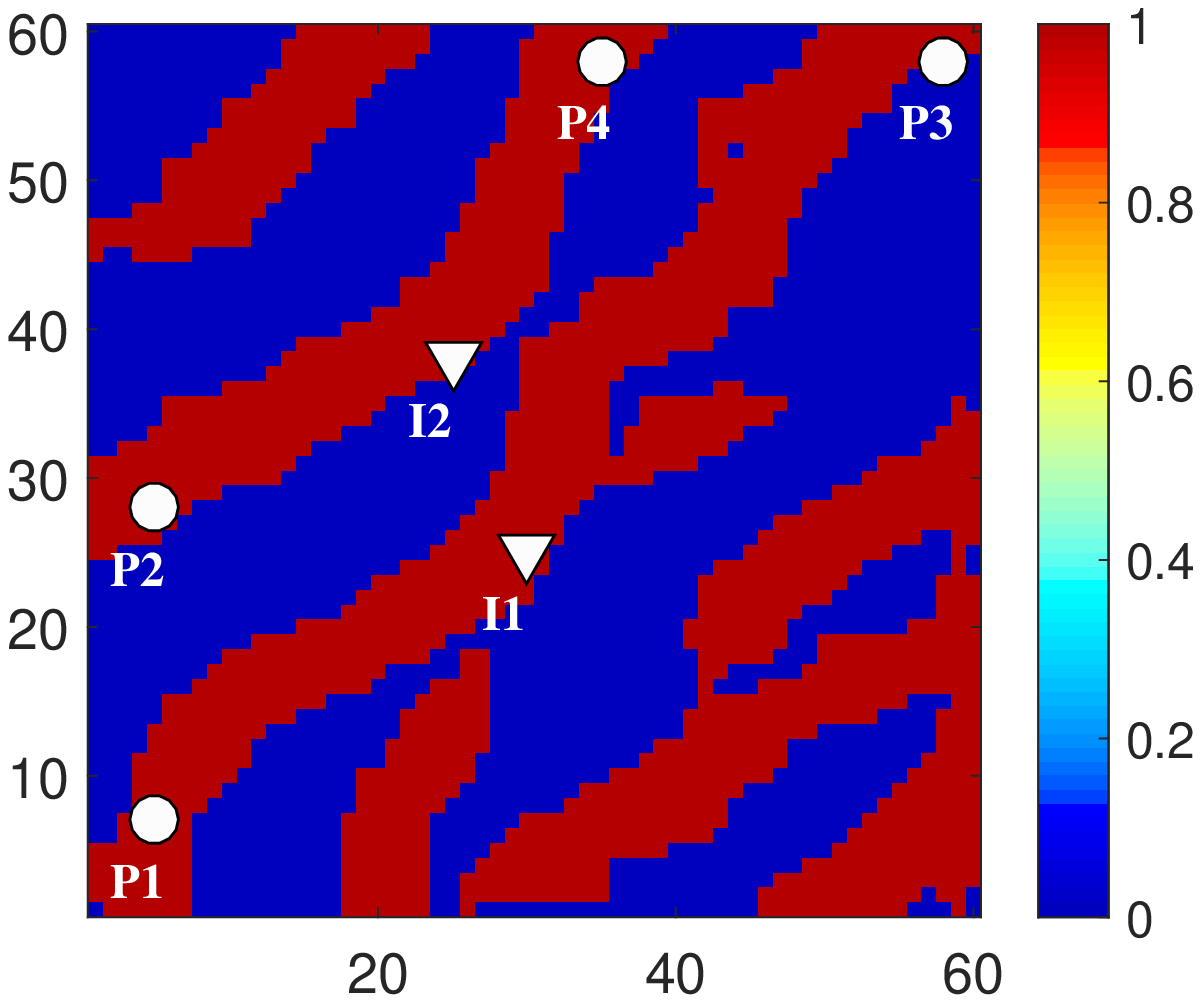}
        \caption{}
    \end{subfigure}%
    \hspace{0.1\baselineskip}
    \begin{subfigure}[b]{0.32\textwidth}
        \includegraphics[width=1\textwidth]{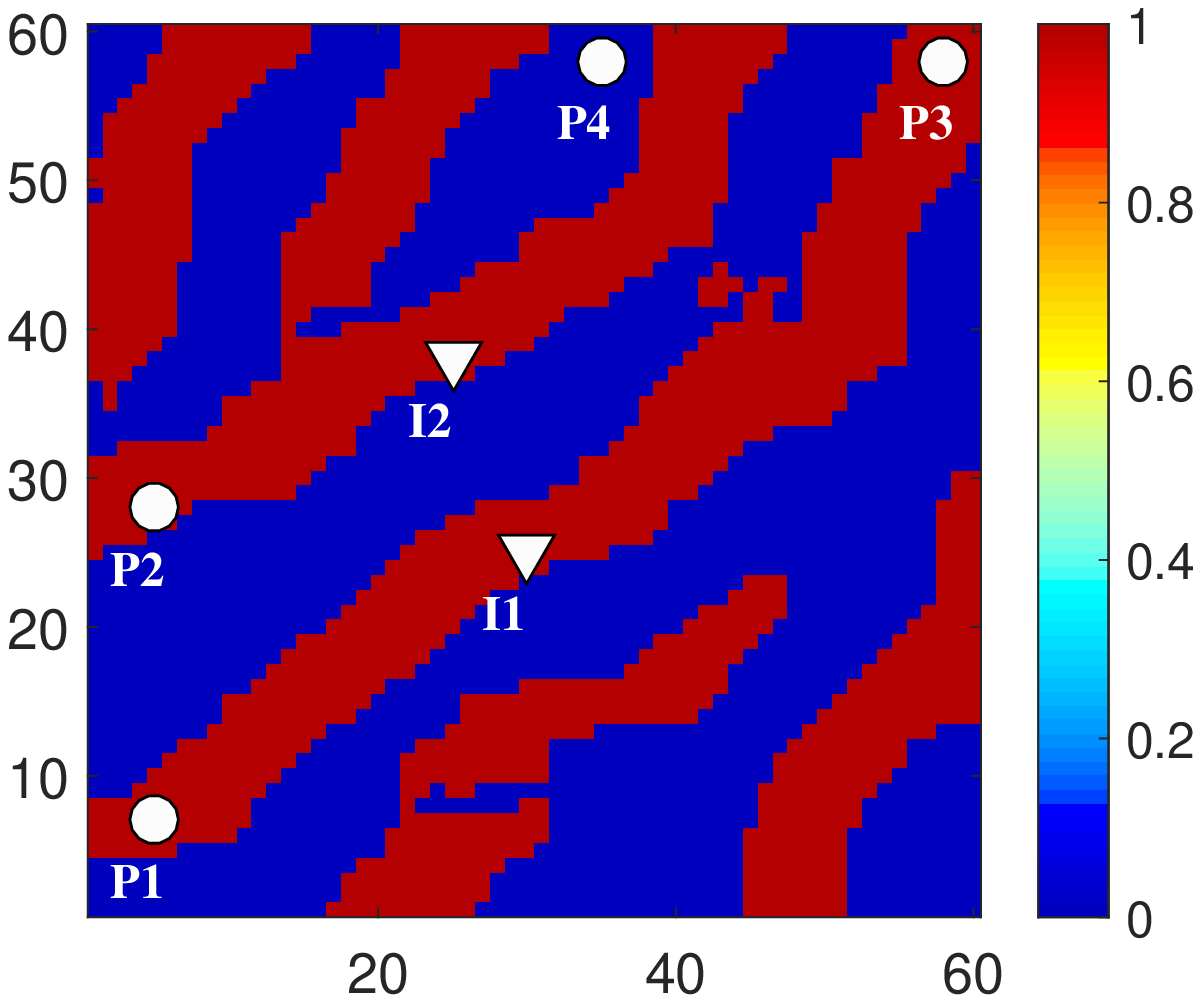}
        \caption{}
    \end{subfigure}%
    
    \caption{Posterior facies models with new wells P3 and P4. \textbf{a,~b,~c} Three O-PCA posterior models, \textbf{d,~e,~f} three CNN-PCA posterior models.}
    \label{fig-hm-facies-iw}
\end{figure}

Production forecasting for new wells is highly uncertain as the prior models are not conditioned to hard data at the new well locations, and the production data observed in the first 1000~days are not sensitive to model properties in the vicinity of the new wells. Figure~\ref{fig-hm-facies-iw} shows O-PCA and CNN-PCA posterior models together with new wells P3 and P4. For both sets of posterior models, relatively large uncertainty is observed for facies type at the new well locations, and in the connectivity between the new wells and the existing injectors. The CNN-PCA posterior models, however, display better channel connectivity than the O-PCA models, so they would appear to be better suited to describe the uncertain range of performance associated with wells P3 and P4.

\begin{figure}[!htb]
    \centering
    \begin{subfigure}[b]{0.43\textwidth}
        \includegraphics[width=1\textwidth]{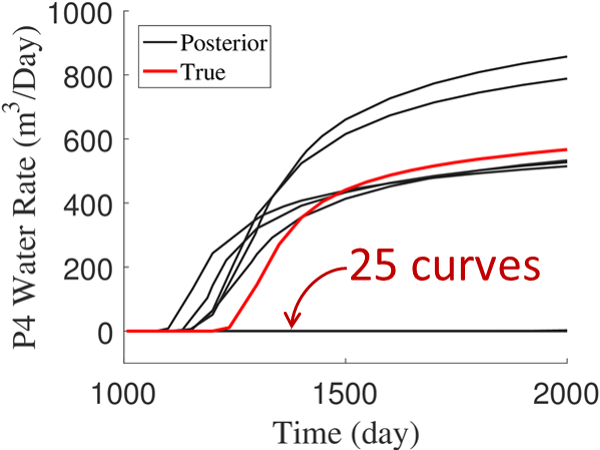}
        \caption{}
    \end{subfigure}%
    \hspace{2\baselineskip}
    \begin{subfigure}[b]{0.43\textwidth}
        \includegraphics[width=1\textwidth]{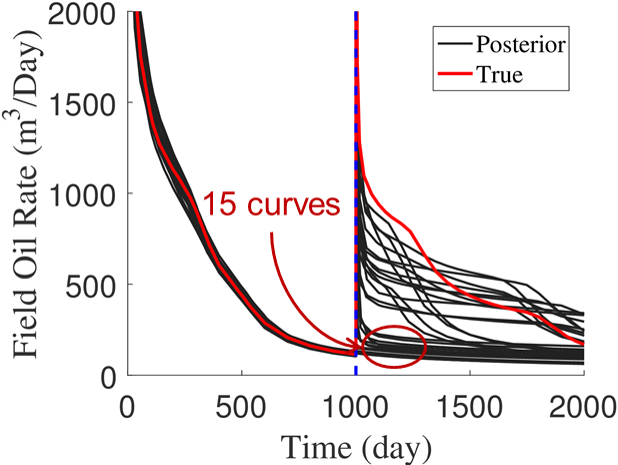}
        \caption{}
    \end{subfigure}%
    
    \begin{subfigure}[b]{0.43\textwidth}
        \includegraphics[width=1\textwidth]{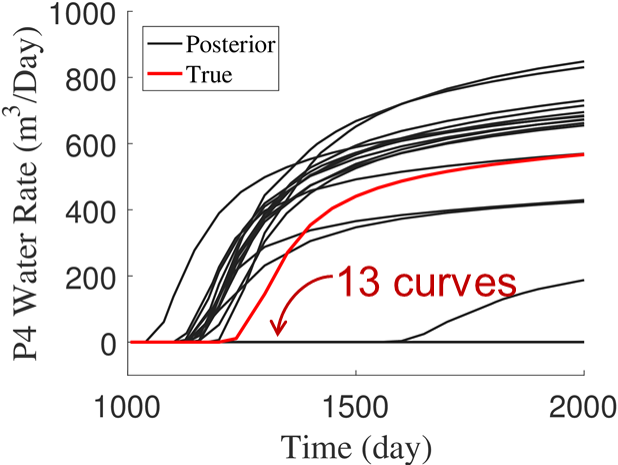}
        \caption{}
    \end{subfigure}%
    ~ 
    \hspace{2\baselineskip}
    \begin{subfigure}[b]{0.43\textwidth}
        \includegraphics[width=1\textwidth]{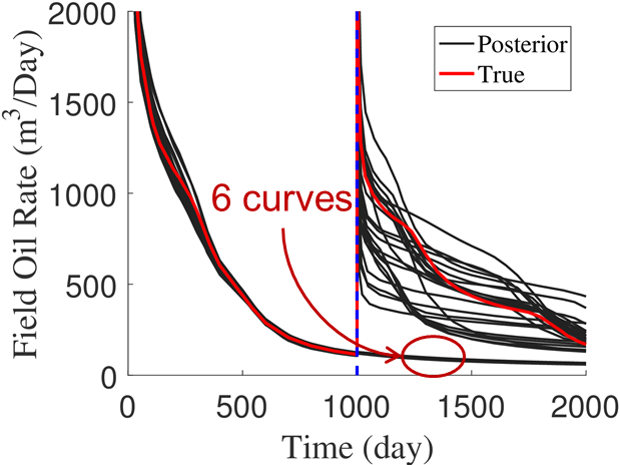}
        \caption{}
    \end{subfigure}%
    
    \caption{Production forecasts with new wells from the true model (red curves) and posterior models (black curves, 30 posterior models generated). \textbf{a} P4 water rate from O-PCA posterior models, 
    \textbf{b} field oil rate from O-PCA posterior models, \textbf{c} P4 water rate from CNN-PCA posterior models, and \textbf{d} field oil rate from CNN-PCA posterior models.}
    \label{fig-hm-prod-iw}
\end{figure}

Water production rate predictions for new well P4 are shown in Fig.~\ref{fig-hm-prod-iw}a,~c. It is evident that only a relatively small fraction of O-PCA posterior models (5 out of 30) predict water breakthrough during the simulation period. By contrast, more than half of the CNN-PCA posterior models (17 out of 30), predict the relatively early water breakthrough that occurs for the true model. Analogous results are observed for P3 water production. Also of interest are predictions for field oil rate (Fig.~\ref{fig-hm-prod-iw}b,~d). In the true model, oil contained in the channels connecting I1 and P3, and I2 and P4, is produced when the new wells begin operating. Thus the true predictions entail a large increase in oil production rate at 1000~days. For O-PCA, half of the posterior models predict little or no increase in oil rate at this time (Fig.~\ref{fig-hm-prod-iw}b). In addition, portions of the true response lie outside the O-PCA posterior predictions. In comparison, most of the CNN-PCA posterior models (24 out of 30) predict a large increase in oil production at 1000~days, and the true results fall within the CNN-PCA posterior predictions.

Although the results in Figs.~\ref{fig-hm-facies-iw} and \ref{fig-hm-prod-iw} suggest qualitatively that CNN-PCA is performing well for this problem, such a conclusion cannot be definitively made until CNN-PCA results are compared to those from a rigorous posterior sampling strategy such as rejection sampling. In rejection sampling, a large number of prior SGeMS realizations are generated and simulated. Realizations are rejected or accepted based on the degree of mismatch between simulation results and observed data. Predictions from the accepted realizations provide an accurate assessment of posterior uncertainty. Rejection sampling is, however, extremely expensive and is thus not performed in this study.

\section{CNN-PCA for Bimodal Deltaic Fan Systems}
\label{sec-bimodal-df}
In this section, the CNN-PCA procedure is extended for application to a more complex bimodal deltaic fan system. The development is limited to the generation of realizations -- no flow results will be presented. The model here corresponds to a non-stationary random field with global trends in channel direction and channel width. Thus, the spatial correlation structure of the deltaic fan model is more difficult to characterize compared to the channelized case. In addition, this system is bimodal rather than binary, meaning the properties within each of the facies are not constant but instead follow prescribed (Gaussian) distributions. 

\begin{figure}[!htb]
    \centering
    \begin{subfigure}[b]{0.34\textwidth}
        \includegraphics[width=1\textwidth]{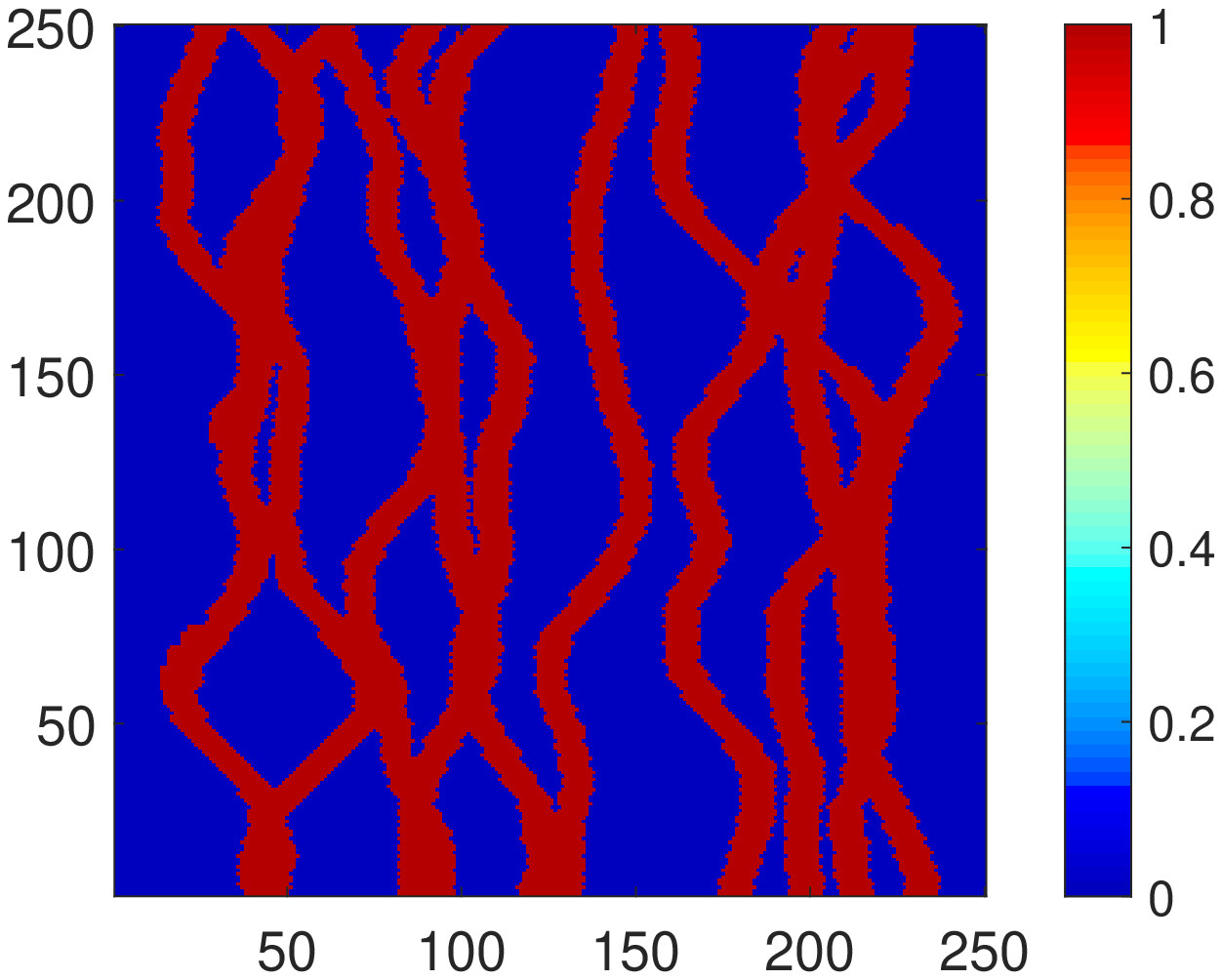}
        \caption{}
    \end{subfigure}%
    ~
    \begin{subfigure}[b]{0.32\textwidth}
        \includegraphics[width=1\textwidth]{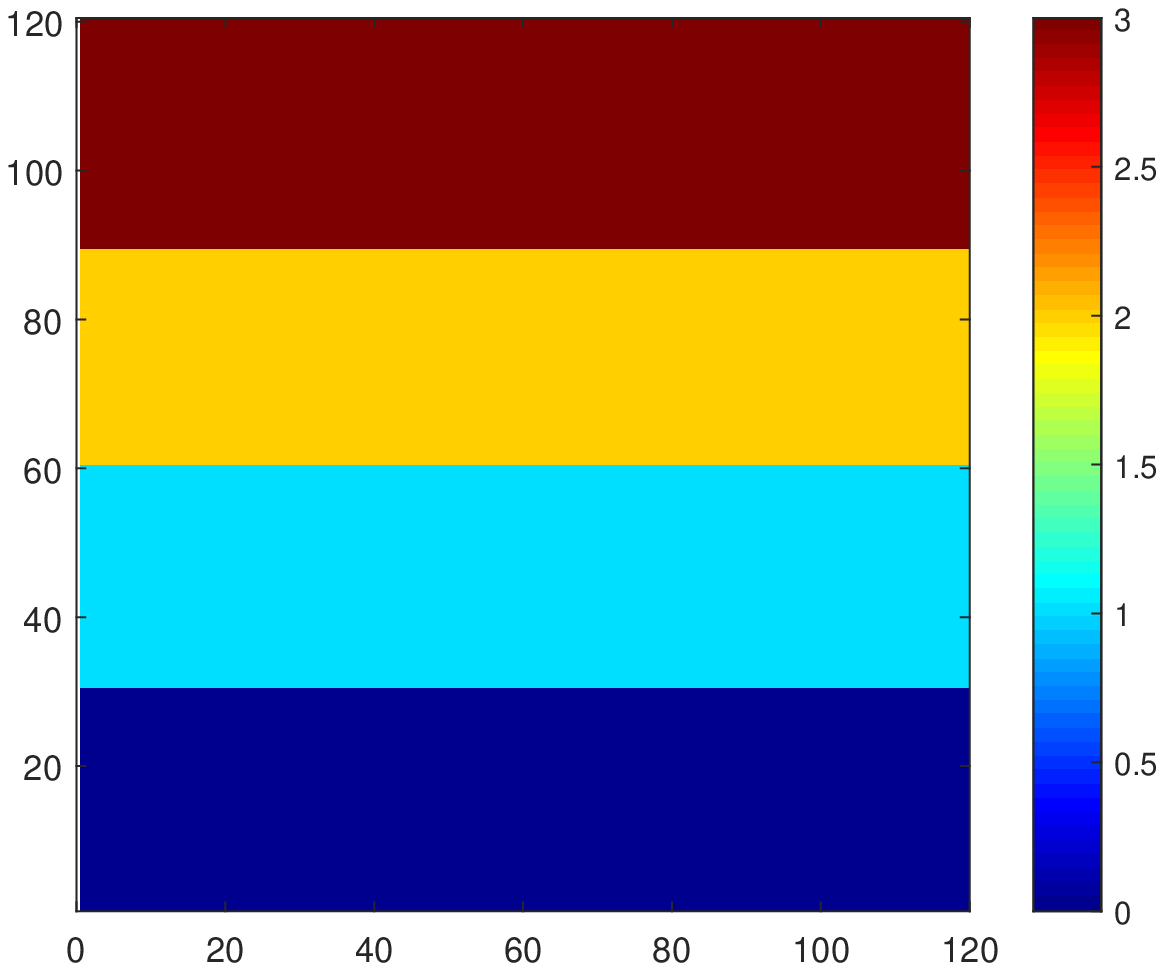}
        \caption{}
    \end{subfigure}%
    ~
    \begin{subfigure}[b]{0.32\textwidth}
        \includegraphics[width=1\textwidth]{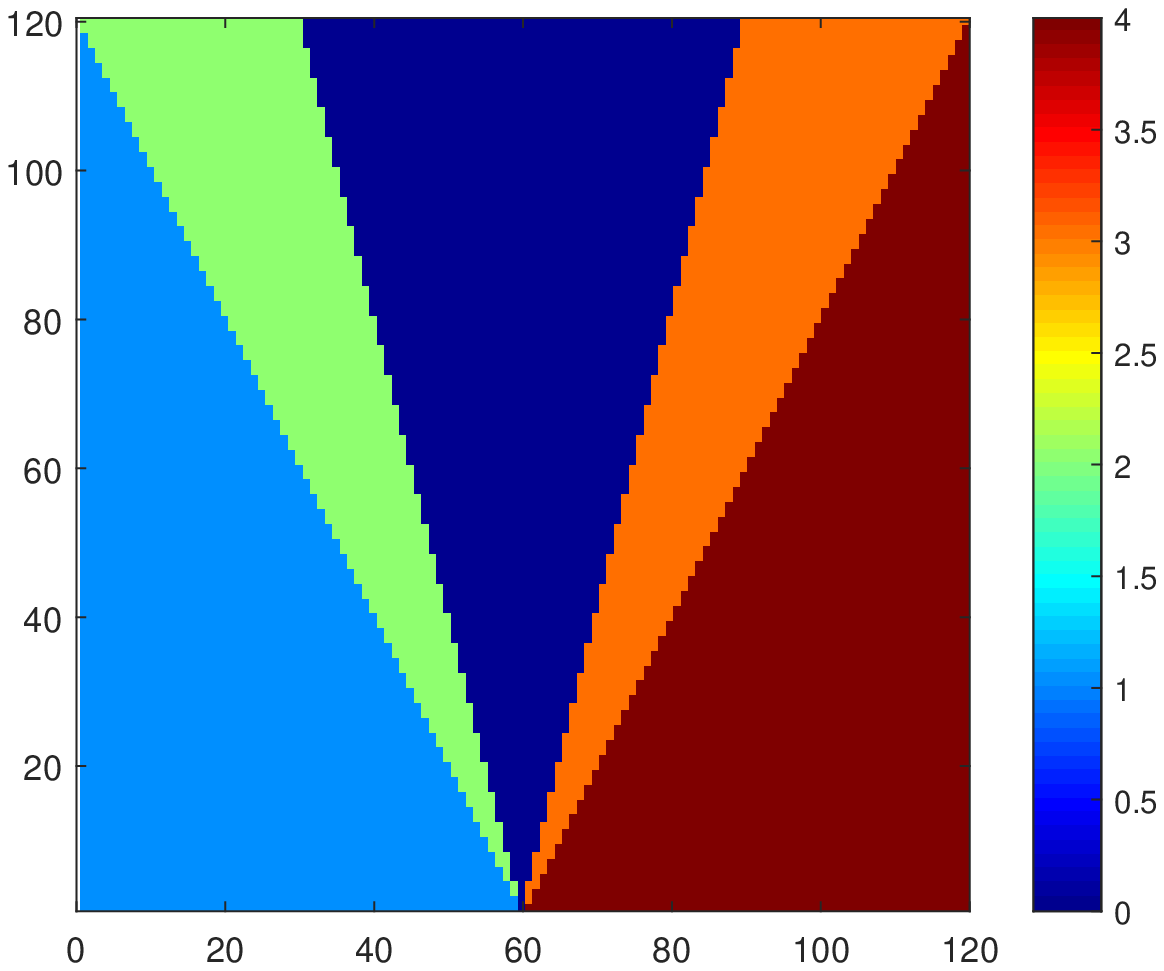}
        \caption{}
    \end{subfigure}%
    \caption{Input for the `snesim' algorithm to generate a binary facies model for the deltaic fan system. \textbf{a} training image, \textbf{b} local affinity transformation region, \textbf{c} local rotation region.}
    \label{fig-snesim-df}
\end{figure}

In this case, the model is defined on a $120 \times 120$ grid and the components of $\Bm$ represent log-permeability ($\log{k}$) in each grid block. The procedure for generating realizations for the deltaic fan system using SGeMS is described in \cite{Vo2016}. The first step is to generate a binary facies model using the `snesim' algorithm with local affinity transformation and rotation. Figure~\ref{fig-snesim-df} shows the training image and local regions for affinity transformation and rotation. The training image, defined on a $250 \times 250$ grid, represents a binary channelized system with the main channel orientation in the $y$-direction. The scaling factors for the four affinity transformation regions are 1.5, 1, 0.5 and 0.3. The rotation angles for the five rotation regions are -45, -30, 0, 30 and 45 degrees. In the next step, Gaussian realizations for $\log{k}$ within each facies (with mean and variance of 8 and 0.8 in sand, and 3 and 0.4 in mud) are simulated independently using the `sgsim' algorithm in SGeMS. The log-permeability for each block is then assigned according to the binary facies model using the cookie-cutter approach.

\begin{figure}[!htb]
    \centering
    \begin{subfigure}[b]{0.24\textwidth}
        \includegraphics[width=1\textwidth]{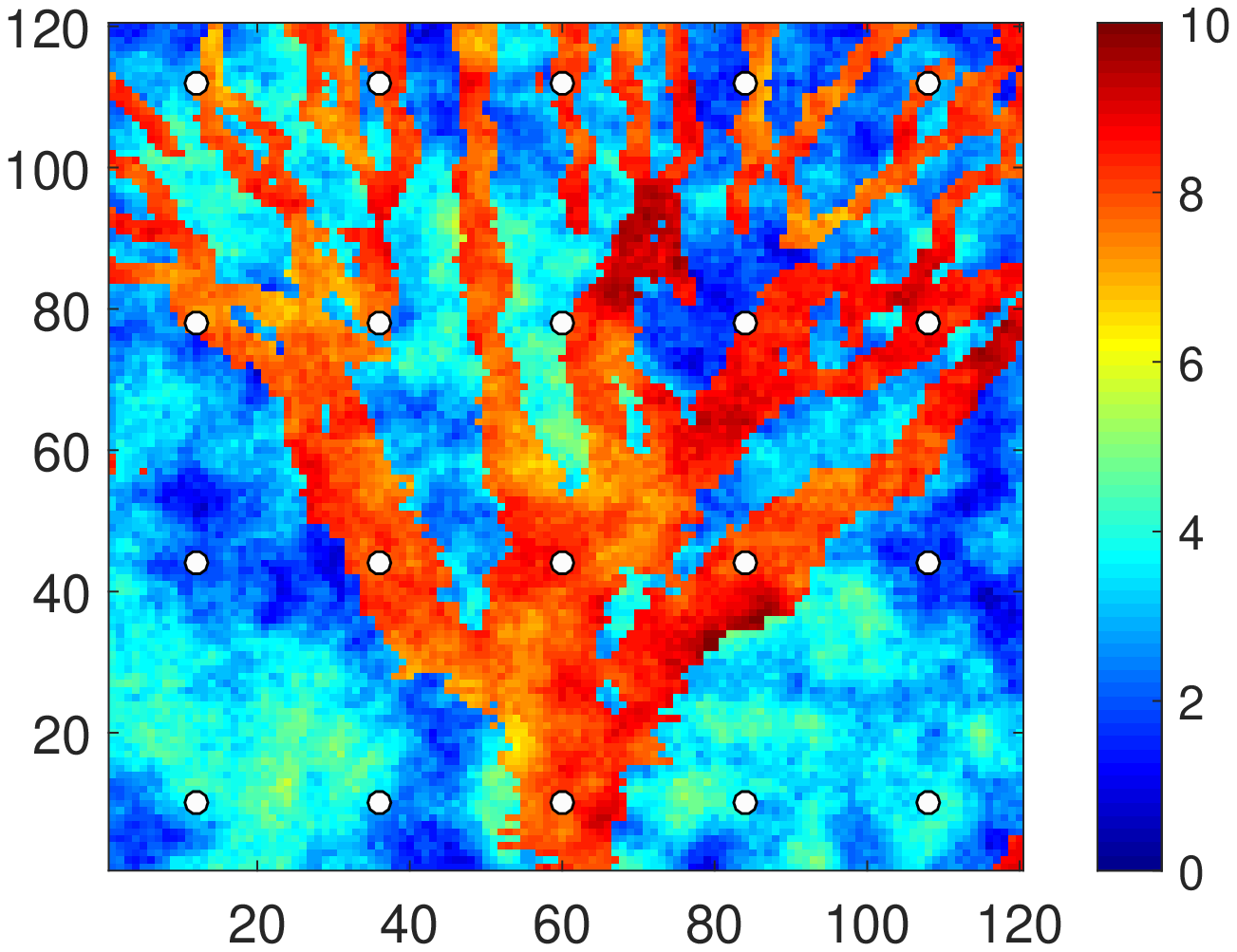}
        \caption{}
    \end{subfigure}%
    \begin{subfigure}[b]{0.24\textwidth}
        \includegraphics[width=1\textwidth]{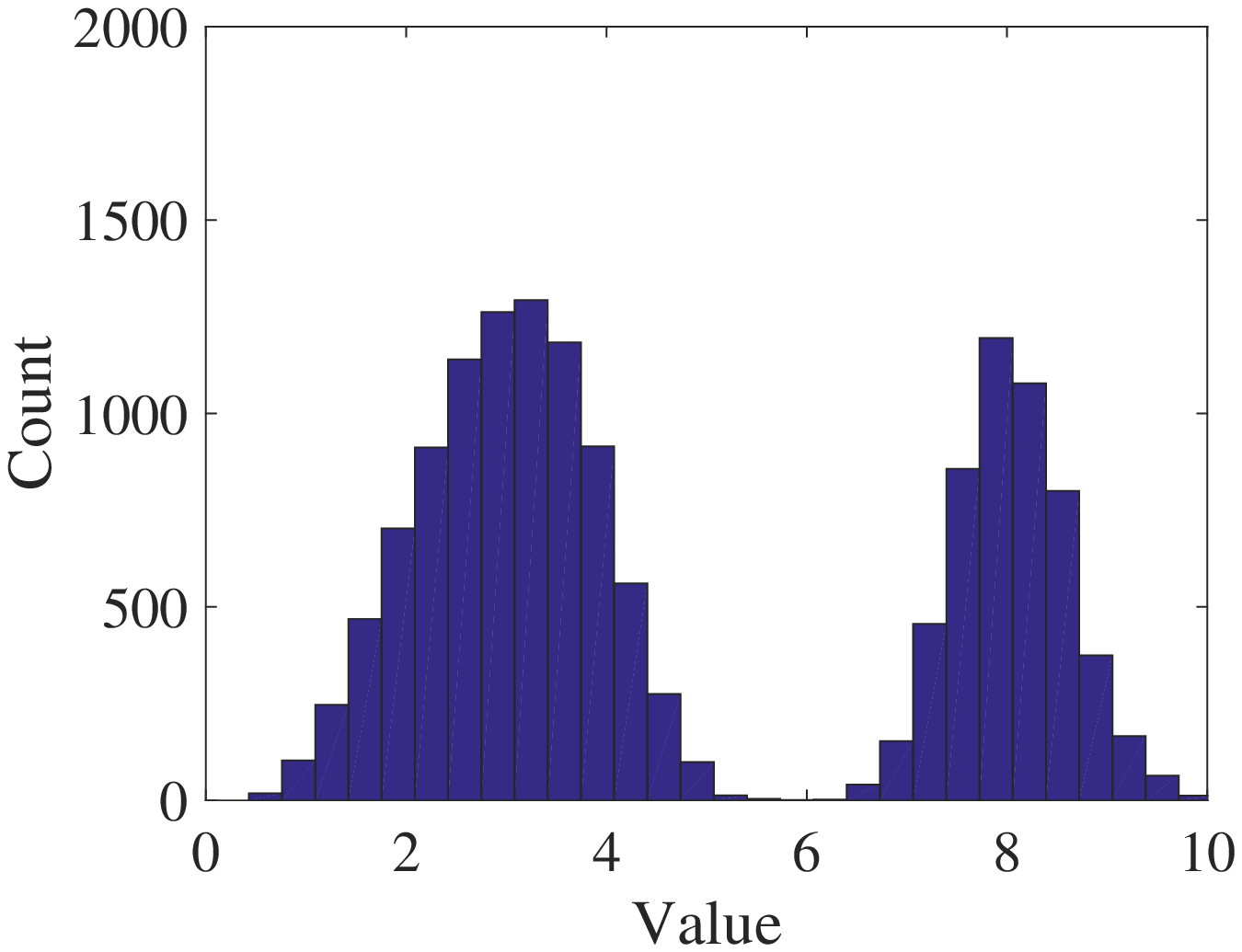}
        \caption{}
    \end{subfigure}%
    \begin{subfigure}[b]{0.24\textwidth}
        \includegraphics[width=1\textwidth]{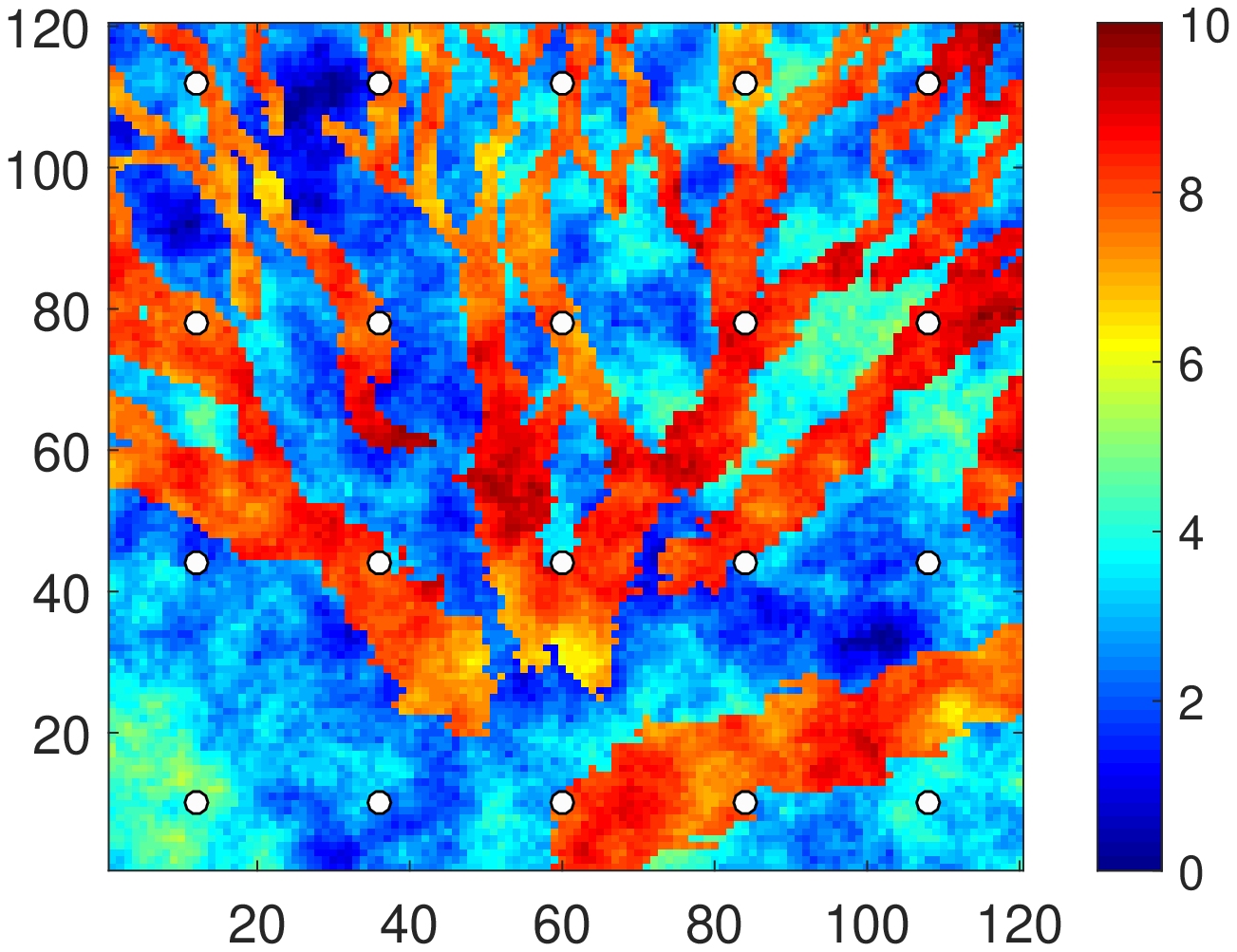}
        \caption{}
    \end{subfigure}%
    \begin{subfigure}[b]{0.24\textwidth}
        \includegraphics[width=1\textwidth]{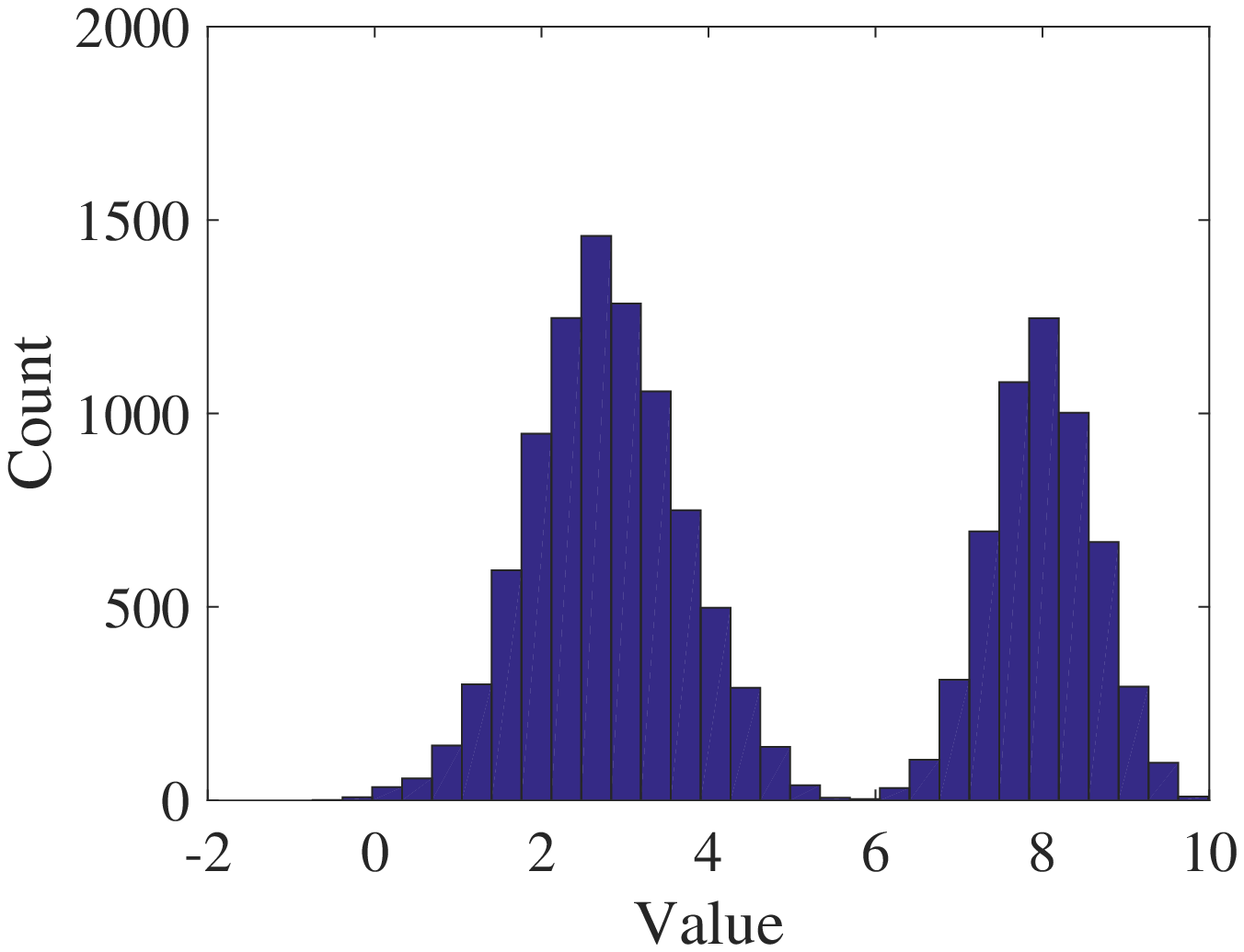}
        \caption{}
    \end{subfigure}%
    \caption{Two SGeMS realizations for $\log{k}$ (\textbf{a,~c}) and corresponding histograms (\textbf{b,~d}) for the bimodal deltaic fan system.}
    \label{fig-sgems-df}
\end{figure}

A total of $\Nr=1500$ realizations, conditioned to hard data at 20 well locations, are generated. Figure~\ref{fig-sgems-df} shows two SGeMS realizations and the corresponding histograms. The six wells toward the lower-left and lower-right portions of the model are in mud, while the other 14 wells are in sand. A PCA representation, with $l=150$, is constructed from the SGeMS realizations. The procedure for constructing the O-PCA representation for bimodal systems is described in detail in \cite{Vo2015}. The basic idea is to post-process PCA models with the minimization 
\begin{equation}
\label{eq-opca-bimodal}
\argmin{\Bx}\big\{||\Bmpca-\Bx||_2^2 + \gamma R(\Bx)\big\}, \  \: x_i \in [\log(k_{\text{min}}),\: \log(k_{\text{max}})],
\end{equation}
where $\gamma$ is a weighting factor and the regularization term $R(\Bx)$ is defined as
\begin{equation}
\label{eq-opca-bimodal-r}
R(\Bx) = \sum_{i=1}^{\Nc} { \Bigg \{ 1-\dfrac{1}{\sqrt[]{2\pi\sigma_1^2}}\exp\bigg(-\dfrac{(x_i-\mu_1)^2}{2\sigma_1^2}\bigg)-\dfrac{1}{\sqrt[]{2\pi\sigma_2^2}}\exp\bigg(-\dfrac{(x_i-\mu_2)^2}{2\sigma_2^2}\bigg) \Bigg \}}.
\end{equation}
This regularization leads to a histogram transformation to match a bimodal distribution, with the two modes at $\mu_1$ and $\mu_2$ and standard deviations $\sigma_1$ and $\sigma_2$. The values for the parameters in the regularization term, together with the weighting factor $\gamma$, are determined through a preliminary optimization such that the average histograms evaluated over a set of O-PCA realizations match closely with the target average histogram computed from the $\Nr$ SGeMS realizations. In this case, the optimal values for the parameters thus obtained are $\gamma=15.3$, $\mu_1=3$, $\mu_2=8.3$, $\sigma_1^2=4.1$, and $\sigma_2^2=1.6$. 

The first step to construct the CNN-PCA models is to select a reference model $\mref$. In this case, there is no deltaic fan training image that depicts the non-stationary spatial correlation structure. In addition, not all SGeMS realizations provide the expected geological connectivity. This is the case with the realization shown in Fig.~\ref{fig-sgems-df}c, where the main channels are `broken' at the bottom. Therefore, one particular SGeMS realization (Fig.~\ref{fig-sgems-df}a), which displays the expected continuity, is taken as the reference model $\mref$. Again, a total of $\Nt=3000$ random PCA realizations are used to train the model transform net. In this case, the weighting factors in Eq.~\ref{eq_cnnpca_hd} are $\gamma_s=7.5$ and $\gamma_h=10$, both obtained through numerical experimentation. 

Training for 3~epochs (2250~iterations) requires approximately 10~minutes for this case. This is significantly faster than the deep-learning-based geological parameterizations discussed in \cite{Laloy2017b,Laloy2017}, where training takes around 3 to 5~hours for 2D binary facies models of similar size on one GPU (NVIDIA Tesla K40). The improved performance here is mainly due to the smaller number of epochs (3 epochs compared to over 50 epochs in \cite{Laloy2017b,Laloy2017}) needed for the model transform net to converge. Faster convergence for the training here may be due to the fact that the CNNs here use the PCA model as input rather than random noise, as in \cite{Laloy2017b,Laloy2017}. The PCA model preserves the target spatial structure up to two-point correlations, so transforming PCA models to the final post-processed models is a simpler process than transforming from random noise. In addition, in the previous studies either an encoder network or a discriminator network needed to be trained to capture the multiple-point correlations of the models, while in this study an off-the-shelf network (the VGG network) is used directly to characterize the multipoint statistics.

After appropriate training, it was found that there was still discrepancy between the histogram of the models $f_W(\Bmpca)$ and the target histogram from the SGeMS models. Therefore, a final histogram transformation step is performed by applying O-PCA to post-process $f_W(\Bmpca)$. The final CNN-PCA model for the bimodal system is (analogous to Eq.~\ref{eq-opca-bimodal})
\begin{equation}
\label{eq-cnnopca-bimodal}
\Bmcnnpca = \argmin{\Bx}\big\{||f_W(\Bmpca)-\Bx||_2^2 + \gamma R(x)\big\},  \  \: x_i \in [\log(k_{\text{min}}), \: \log(k_{\text{max}})],
\end{equation}
where the regularization term $R(\Bx)$ is defined in Eq.~\ref{eq-opca-bimodal-r}. A separate preliminary optimization was performed to obtain the values for $\gamma$ and the parameters in $R(\Bx)$. These values were found to be $\gamma=14.6$, $\mu_1=2.8$, $\mu_2=8.1$, $\sigma_1^2=3.1$, and $\sigma_2^2=1.3$. Note that, using this O-PCA post-processing as the final step, the CNN-PCA procedure is differentiable and can thus be used with gradient-based history matching.

\begin{figure}[!htb]
    \centering
    \begin{subfigure}[b]{0.33\textwidth}
        \includegraphics[width=1\textwidth]{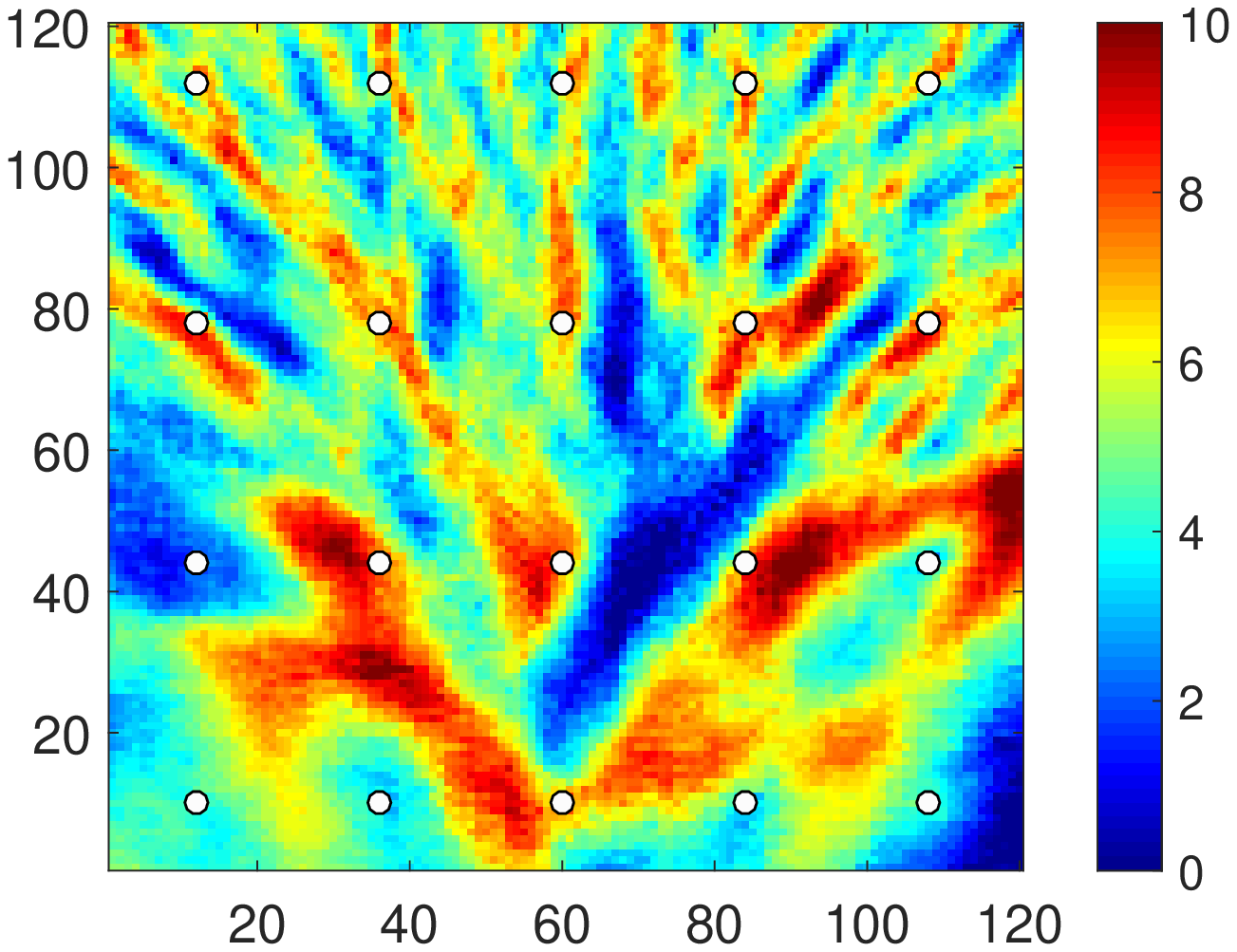}
        \caption{}
    \end{subfigure}%
    ~
    \begin{subfigure}[b]{0.33\textwidth}
        \includegraphics[width=1\textwidth]{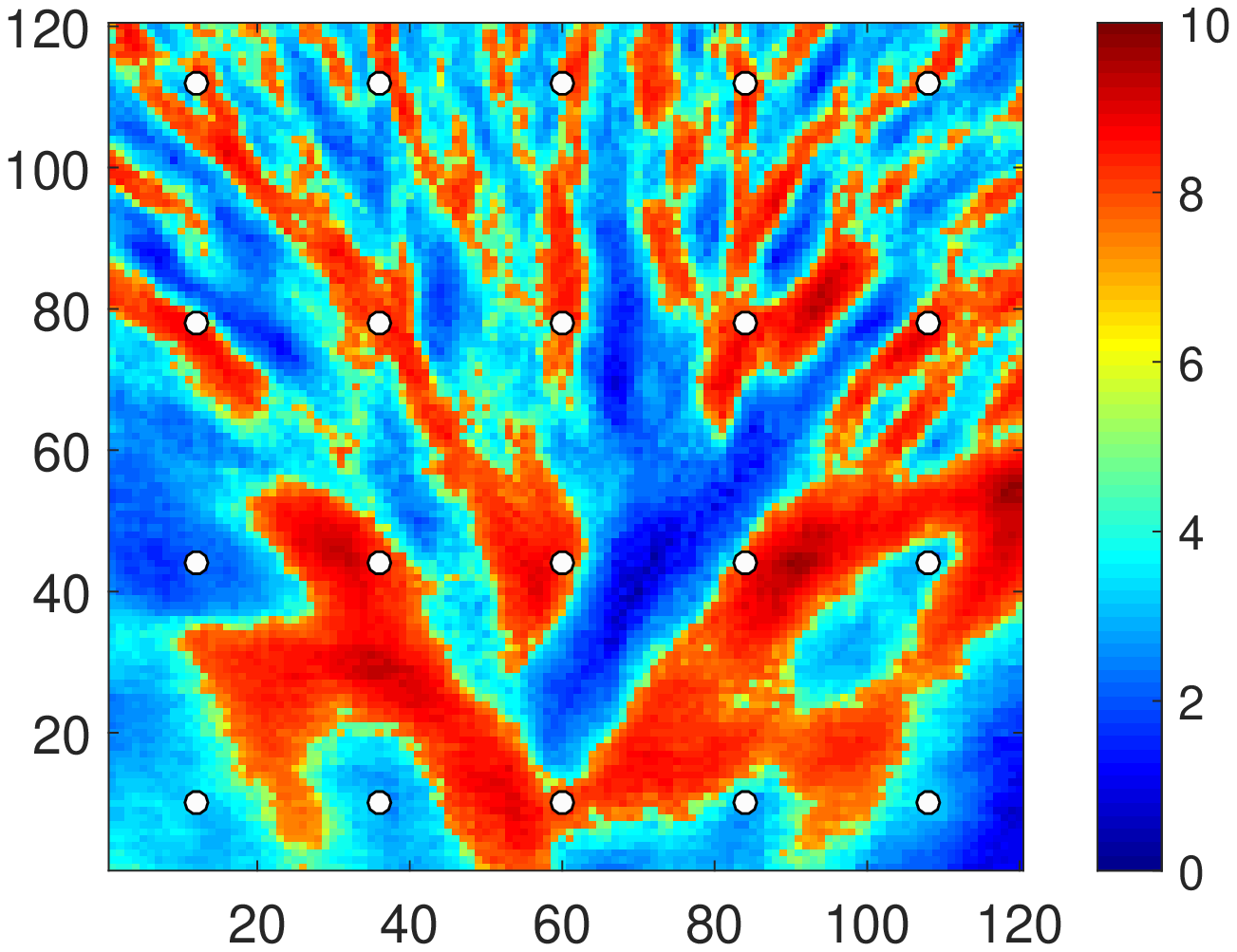}
        \caption{}
    \end{subfigure}%
    ~
    \begin{subfigure}[b]{0.33\textwidth}
        \includegraphics[width=1\textwidth]{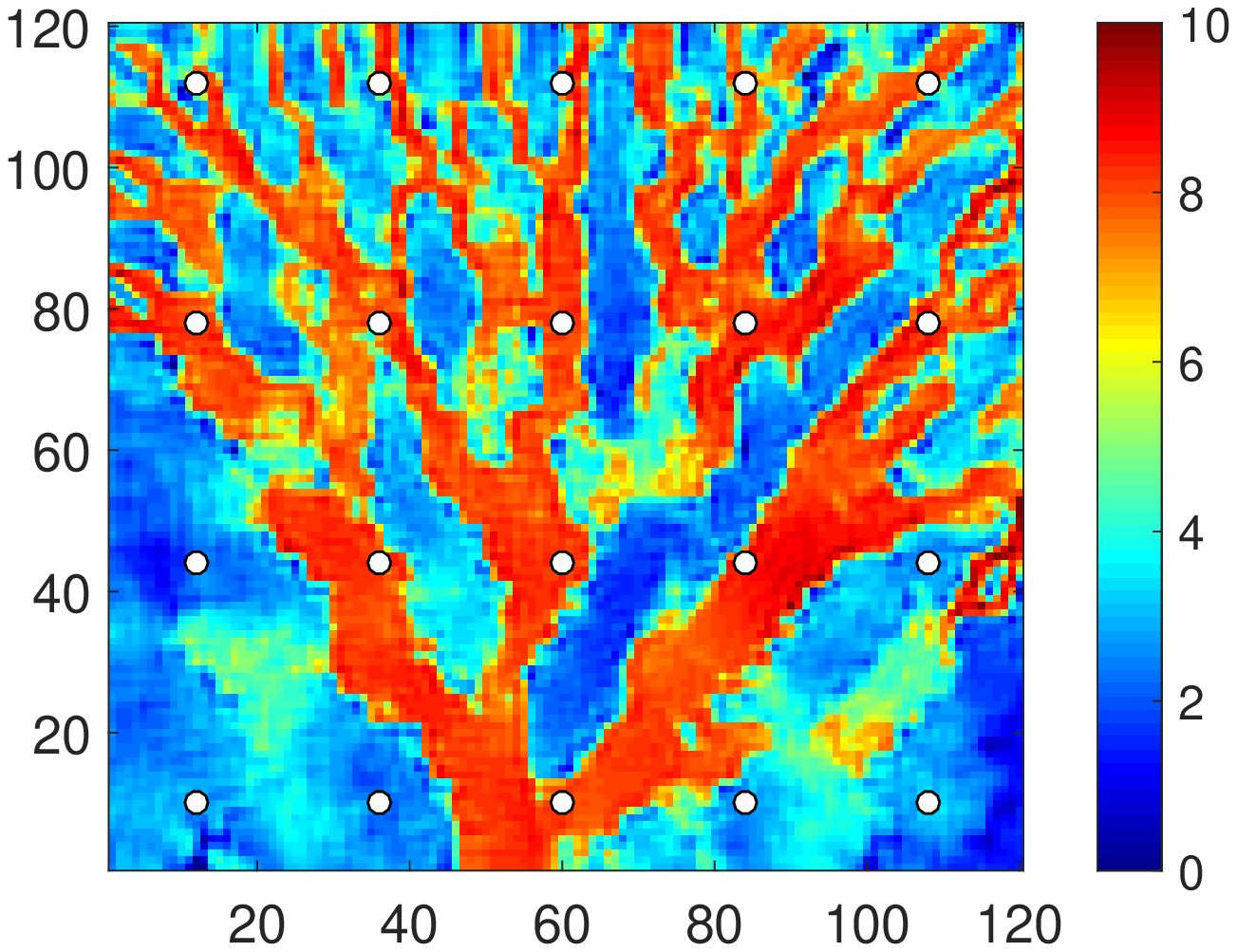}
        \caption{}
    \end{subfigure}%
    
    \hspace{-1\baselineskip}
    \begin{subfigure}[b]{0.33\textwidth}
        \includegraphics[width=1\textwidth]{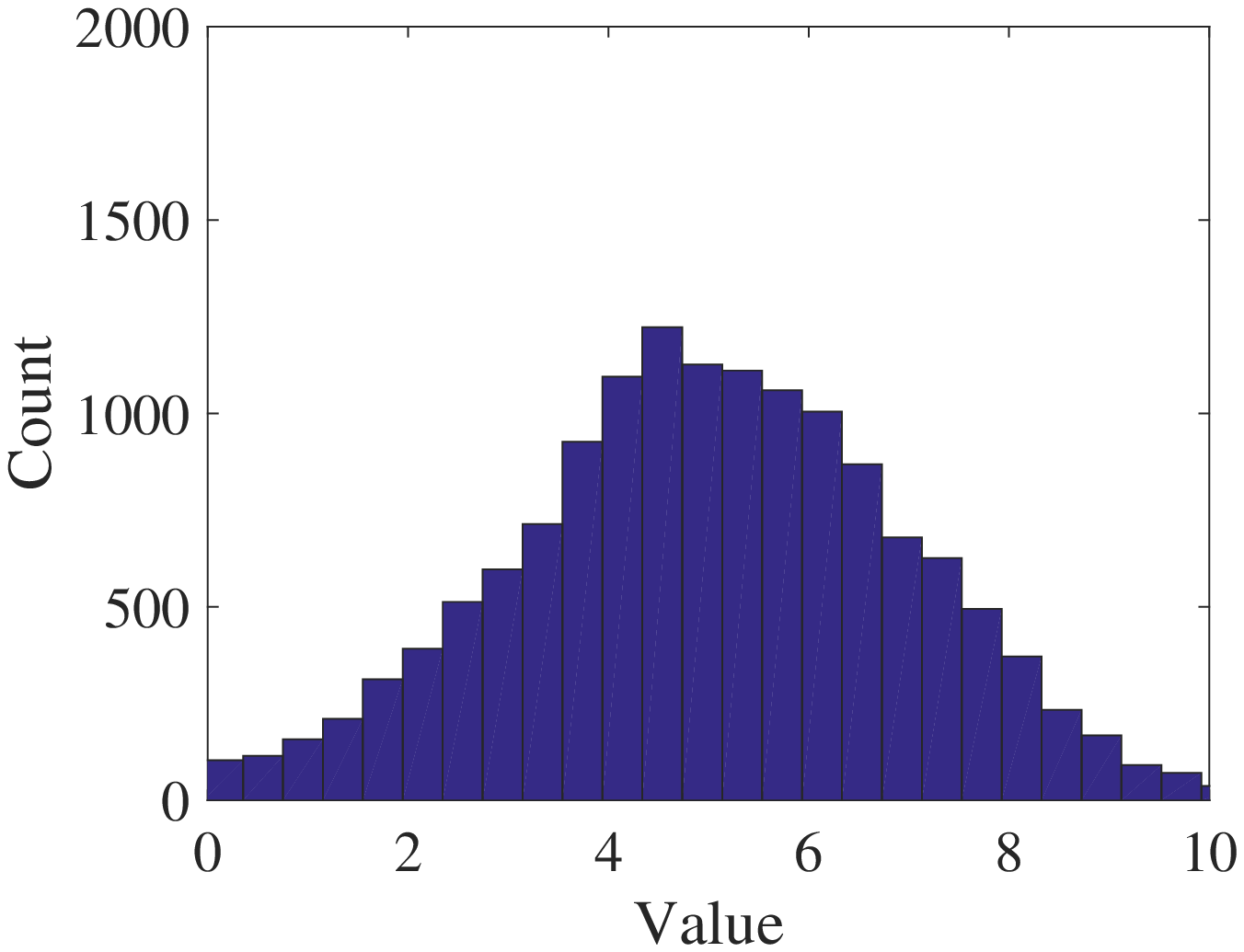}
        \caption{}
    \end{subfigure}%
    ~
    \begin{subfigure}[b]{0.33\textwidth}
        \includegraphics[width=1\textwidth]{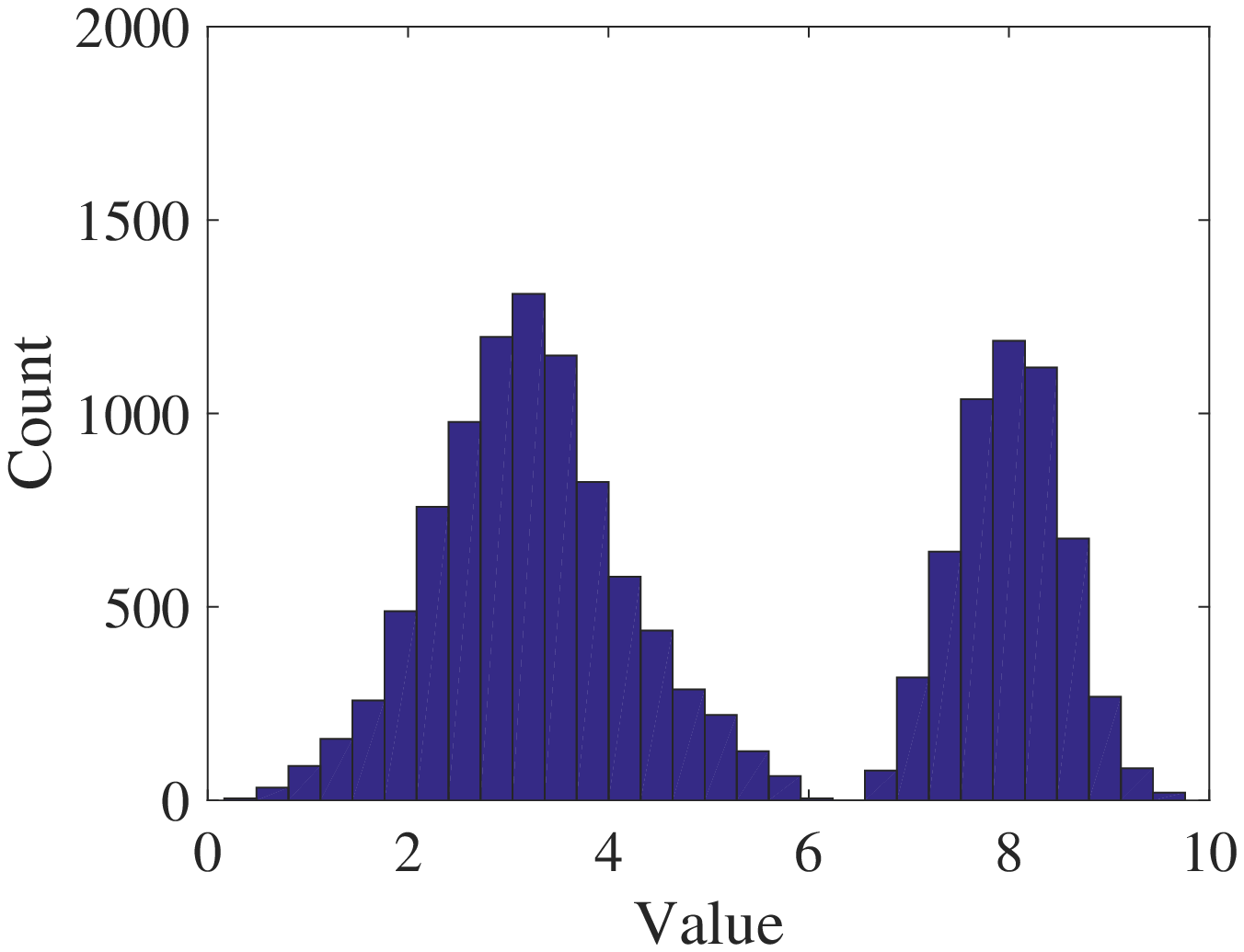}
        \caption{}
    \end{subfigure}%
    ~
    \begin{subfigure}[b]{0.33\textwidth}
        \includegraphics[width=1\textwidth]{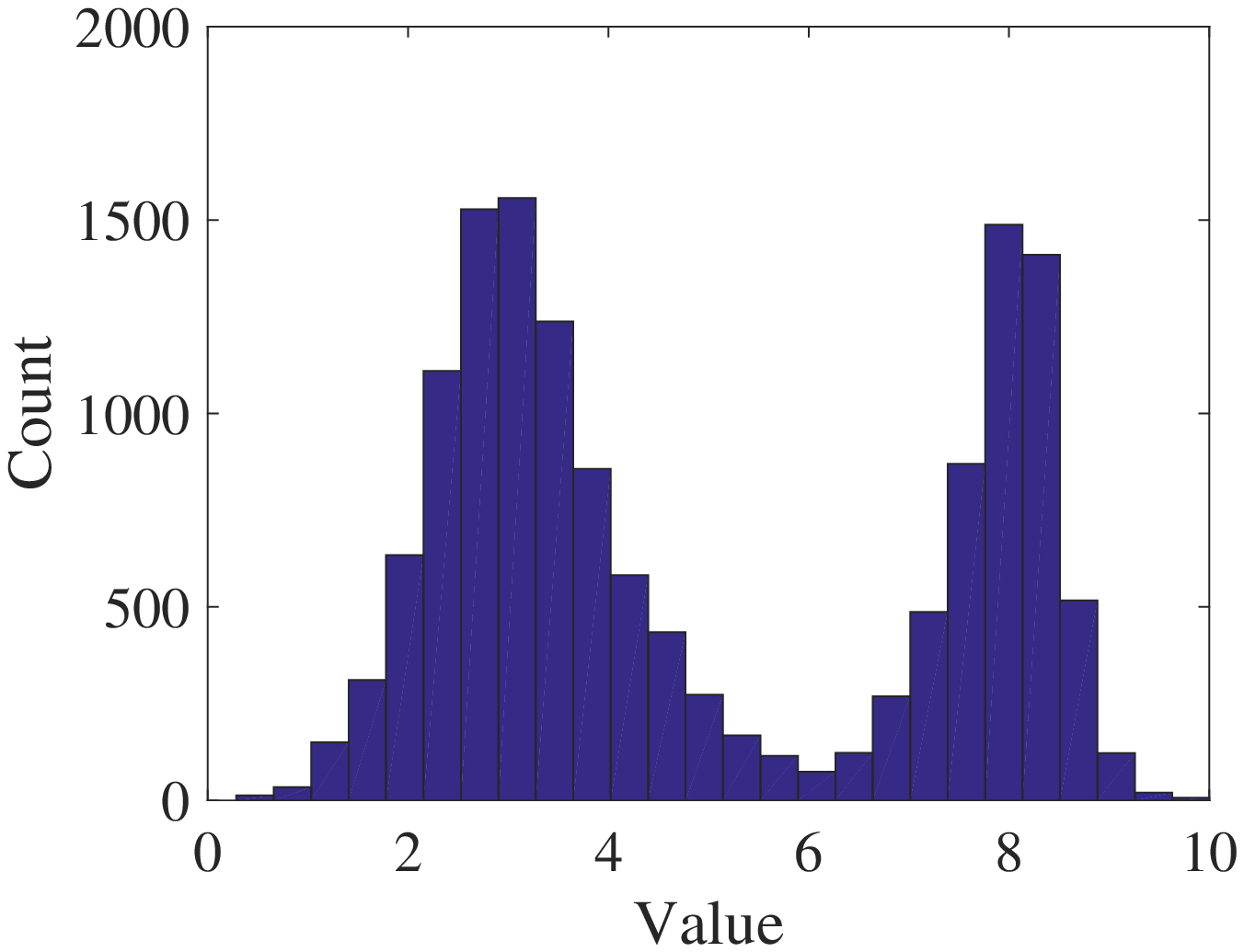}
        \caption{}
    \end{subfigure}%
    \caption{Random conditional realizations for bimodal deltaic fan system with hard data at 20 wells. \textbf{a,~d} PCA realization and histogram, \textbf{b,~e} corresponding O-PCA realization and histogram, \textbf{c,~f} corresponding CNN-PCA realization and histogram.}
    \label{fig-cnnpca-df}
\end{figure}

Figure~\ref{fig-cnnpca-df} displays a PCA realization and the corresponding O-PCA and CNN-PCA realizations, along with the histograms for the three models. The non-stationary features, including decreasing channel width with increasing $y$ and the fan-shaped channel structure, are honored to varying extents in the three realizations. Both the O-PCA and CNN-PCA models have bimodal histograms, while the PCA model has an essentially Gaussian histogram. In terms of channel continuity, the CNN-PCA realization appears more consistent with the reference SGeMS model (in Fig.~\ref{fig-sgems-df}) than the PCA and O-PCA realizations, for both large channels near the bottom of the domain and smaller channels towards the top. In addition, the spurious channels present in the lower-left and lower-right portions of the PCA and O-PCA models do not appear in the CNN-PCA model. This demonstrates that CNN-PCA can be applied for the bimodal non-stationary deltaic fan system. Flow results and history matching for this and other complicated geological systems will be considered in future work.

\FloatBarrier

\section{Concluding Remarks}
\label{sec-concl}

In this paper, a deep-learning-based approach, referred to as CNN-PCA, was developed for the low-dimensional parameterization of geological models characterized by multipoint spatial statistics. CNN-PCA represents a generalization of the O-PCA method, as both procedures are based on the post-processing of PCA models to match the style of a reference model. CNN-PCA utilizes a set of metrics based on a pre-trained convolutional neural network to characterize multiple point correlations, which enables the method to preserve the complex geological patterns present in the reference model (e.g., training image). A hard data loss term was introduced to generate conditional CNN-PCA realizations. For the 2D systems considered in this study, CNN-PCA is essentially a direct application of the fast neural style transfer algorithm, originally developed in computer vision for dealing with images. In CNN-PCA, the PCA models are post-processed quickly, by feeding them through a convolutional neural network called the model transform net. The computational cost of CNN-PCA is mainly associated with the offline training of this model transform net. The training process is significantly faster with CNN-PCA than with existing deep-learning-based geological parameterization techniques.

Model realizations generated using CNN-PCA were presented for a 2D binary channelized system. Even in the absence of hard data, unconditional CNN-PCA realizations provided a high-level of channel continuity and uniform channel width, consistent with the reference training image. Unconditional O-PCA realizations, by contrast, do not provide the same degree of geological realism. For conditional systems, CNN-PCA models were shown to honor all hard data for the cases considered. Application of CNN-PCA for a non-stationary bimodal deltaic fan system was also presented. For this system, O-PCA was used as a final post-processing step to ensure that the histogram of the final model matches the reference histogram. The CNN-PCA realizations for the bimodal deltaic fan system were shown to preserve the global trends in channel direction and channel width. 

The consistency between CNN-PCA models and reference SGeMS models was demonstrated by evaluating flow-response statistics for an oil-water two-phase flow problem. P10, P50, P90 results and CDFs for injection and production quantities, obtained from random SGeMS, O-PCA, and CNN-PCA realizations, were compared. CNN-PCA models were shown to provide flow statistics in close agreement with flow statistics obtained from reference SGeMS models for a challenging case involving a small amount of conditioning data (O-PCA is not as well suited for cases that lack hard data). History matching for a conditional binary channelized system was also considered. A derivative-free optimization method, PSO--MADS,  was applied for the required minimization, and multiple posterior models were generated using a subspace RML procedure. Both CNN-PCA and O-PCA provided significant uncertainty reduction for production forecasts involving existing wells, but CNN-PCA was shown to provide more realistic results than O-PCA for forecasts involving new wells.

In this study, CNN-PCA was applied for 2D systems. In future work, extensions to treat 3D systems should be developed. For 3D cases, the direct use of the off-the-shelf convolutional neural network (the VGG-16 net), pre-trained with 2D images, is no longer viable. However, the general approach of using convolutional neural networks to characterize complex (multipoint) 3D geological models should still be valid. In fact, when correlations are weak in the vertical direction, straightforward extensions of the methodology presented in this paper may be sufficient. Other areas for future research include the evaluation of different approaches for history matching with CNN-PCA models. Additional minimization algorithms, such as adjoint-gradient-based methods, as well as ensemble-based techniques, should be assessed. Multilevel history matching treatments, along the lines of those presented in \cite{Liu2017}, should also be considered for use with CNN-PCA. Such methods enable history matching to be accomplished more efficiently in stages, using sequential sets of PCA basis components, and were shown to be effective when applied with O-PCA.

\begin{acknowledgements}
We thank the industrial affiliates of the Stanford Smart Fields Consortium for financial support. We are grateful to Hai Vo for providing O-PCA code and geological models, and to Hang Zhang and Abhishek Kadian for their open source PyTorch implementation of the neural style transfer and fast neural style transfer algorithms. We also acknowledge the teaching staff for the Stanford CS231N course for offering helpful guidance and providing computing resources on Google cloud.
\end{acknowledgements}

\newpage
\section*{Appendix: Model Transform Net Architecture}
\label{sec-app}

The architecture of the model transform net is summarized in Table~\ref{tab-cnn}. It is the same as that in \cite{Johnson2015} except for the first and last layers. In the table, `Conv' denotes a convolutional layer immediately followed by spatial batch normalization and a ReLU nonlinear activation. The last `Conv' layer is an exception as it only contains a convolutional layer. `Residual block' contains a stack of two convolutional layers, each with 128 filters of size $3 \times 3 \times 128$ and stride 1. Within each residual block, the first convolutional layer is followed by a spatial batch normalization and a ReLU nonlinear activation. The second convolutional layer is followed only by a spatial batch normalization. The final output of the residual block is the sum of the input to the first convolutional layer and the output from the second convolutional layer. 

\begin{table}[!htb]
\centering
\begin{tabular}{ c | c  }
  \textbf{Layer} & \textbf{Output size} \\
  \hline
  Input & ($\Nx$, $\Ny$, $1$)  \\
  \\[-1em]
  Conv, 32 filters of size $9 \times 9 \times 1$, stride 1 & ($\Nx$, $\Ny$, $32$)  \\
  \\[-1em]
  Conv, 64 filters of size $3 \times 3 \times32$, stride 2 & ($\Nx/2$, $\Ny/2$, $64$)  \\
  \\[-1em]
  Conv, 128 filters of size $3 \times 3 \times 64$, stride 2 & ($\Nx/4$, $\Ny/4$, $128$)  \\
    \\[-1em]
  Residual block, 128 filters & ($\Nx/4$, $\Ny/4$, $128$)  \\
      \\[-1em]
  Residual block, 128 filters & ($\Nx/4$, $\Ny/4$, $128$)  \\
      \\[-1em]
  Residual block, 128 filters & ($\Nx/4$, $\Ny/4$, $128$)  \\
      \\[-1em]
  Residual block, 128 filters & ($\Nx/4$, $\Ny/4$, $128$)  \\
      \\[-1em]
  Residual block, 128 filters & ($\Nx/4$, $\Ny/4$, $128$)  \\
    \\[-1em]
  Conv, 64 filters of size $3 \times 3 \times 128$, stride $1/2$ & ($\Nx/2$, $\Ny/2$, $64$)  \\
      \\[-1em]
  Conv, 32 filters of size $3 \times 3 \times 64$, stride $1/2$ & ($\Nx$, $\Ny$, $32$)  \\
        \\[-1em]
  Conv, 1 filter of size $9 \times 9 \times 64$, stride $1$ & ($\Nx$, $\Ny$, $1$)  \\
  \hline  
\end{tabular}

\caption{Network architecture used for the model transform net.}
\label{tab-cnn}
\end{table}

\bibliographystyle{spbasic}      
\bibliography{deep_rp,hm_master_thesis,manual,dsi}

\begin{thebibliography}{40}
\providecommand{\natexlab}[1]{#1}
\providecommand{\url}[1]{{#1}}
\providecommand{\urlprefix}{URL }
\expandafter\ifx\csname urlstyle\endcsname\relax
  \providecommand{\doi}[1]{DOI~\discretionary{}{}{}#1}\else
  \providecommand{\doi}{DOI~\discretionary{}{}{}\begingroup
  \urlstyle{rm}\Url}\fi
\providecommand{\eprint}[2][]{\url{#2}}

\bibitem[{Astrakova and Oliver(2015)}]{Astrakova2015}
Astrakova A, Oliver DS (2015) {Conditioning truncated pluri-Gaussian models to
  facies observations in ensemble-Kalman-based data assimilation}. Mathematical
  Geosciences 47(47):345--367

\bibitem[{Canchumuni et~al.(2017)Canchumuni, Emerick, and
  Pacheco}]{Canchumuni2017}
Canchumuni SA, Emerick AA, Pacheco MA (2017) {Integration of ensemble data
  assimilation and deep learning for history matching facies models. Paper
  OTC-28015-MS, presented at the OTC Brasil, Rio de Janeiro, Brazil, 24-26
  October.}

\bibitem[{Chang et~al.(2010)Chang, Zhang, and Lu}]{Chang2010}
Chang H, Zhang D, Lu Z (2010) {History matching of facies distribution with the
  EnKF and level set parameterization}. Journal of Computational Physics
  229(20):8011--8030

\bibitem[{Chen et~al.(2016)Chen, Gao, Ramirez, Vink, and Girardi}]{Chen2016}
Chen C, Gao G, Ramirez BA, Vink JC, Girardi AM (2016) {Assisted history
  matching of channelized models by use of pluri-principal-component analysis}.
  SPE Journal 21(05):1793--1812

\bibitem[{Deng et~al.(2009)Deng, Dong, Socher, Li, Li, and Li}]{JiaDeng2009}
Deng J, Dong W, Socher R, Li LJ, Li K, Li FF (2009) {ImageNet: A large-scale
  hierarchical image database}. In: 2009 IEEE Conference on Computer Vision and
  Pattern Recognition, IEEE, pp 248--255

\bibitem[{Dimitrakopoulos et~al.(2010)Dimitrakopoulos, Mustapha, and
  Gloaguen}]{Dimitrakopoulos2010}
Dimitrakopoulos R, Mustapha H, Gloaguen E (2010) {High-order statistics of
  spatial random fields: exploring spatial cumulants for modeling complex
  non-Gaussian and non-linear phenomena}. Mathematical Geosciences 42(1):65--99

\bibitem[{{Echeverr{\'{i}}a Ciaurri} et~al.(2009){Echeverr{\'{i}}a Ciaurri},
  Mukerji, and Santos}]{Echeverria2009}
{Echeverr{\'{i}}a Ciaurri} D, Mukerji T, Santos ET (2009) {Robust scheme for
  inversion of seismic and production data for reservoir facies modeling. Paper
  SEG-2009-2432, presented at the SEG Annual Meeting, Houston, Texas, 25-30
  October.}

\bibitem[{Emerick(2016)}]{Emerick2016}
Emerick AA (2016) {Investigation on principal component analysis
  parameterizations for history matching channelized facies models with
  ensemble-based data assimilation}. Mathematical Geosciences 49(1):85--120

\bibitem[{Gatys et~al.(2015)Gatys, Ecker, and Bethge}]{Gatys2015}
Gatys LA, Ecker AS, Bethge M (2015) {Texture synthesis using convolutional
  neural networks}. Advances in Neural Information Processing Systems pp
  262--270

\bibitem[{Gatys et~al.(2016)Gatys, Ecker, and Bethge}]{Gatys}
Gatys LA, Ecker AS, Bethge M (2016) {Image style transfer using convolutional
  neural networks}. Proceedings of the IEEE Conference on Computer Vision and
  Pattern Recognition pp 2414--2423

\bibitem[{Goodfellow et~al.(2016)Goodfellow, Bengio, and
  Courville}]{IanGoodfellowYoshuaBengio2016}
Goodfellow I, Bengio Y, Courville A (2016) {Deep Learning}. MIT Press Cambridge

\bibitem[{Hakim-Elahi and Jafarpour(2017)}]{Hakim-Elahi2017}
Hakim-Elahi S, Jafarpour B (2017) {A distance transform for continuous
  parameterization of discrete geologic facies for subsurface flow model
  calibration}. Water Resources Research 53(10):8226--8249

\bibitem[{Insuasty et~al.(2017)Insuasty, {Van den Hof}, Weiland, and
  Jansen}]{Insuasty2017}
Insuasty E, {Van den Hof} PMJ, Weiland S, Jansen JD (2017) {Low-dimensional
  tensor representations for the estimation of petrophysical reservoir
  parameters. Paper SPE-182707-MS presented at the SPE Reservoir Simulation
  Conference, Montgomery, Texas, 20-22 February}

\bibitem[{Isebor et~al.(2014)Isebor, {Echeverr{\'{i}}a Ciaurri}, and
  Durlofsky}]{Isebor2014}
Isebor OJ, {Echeverr{\'{i}}a Ciaurri} D, Durlofsky LJ (2014) {Generalized
  field-development optimization with derivative-free procedures}. SPE Journal
  19(05):891--908

\bibitem[{Jafarpour et~al.(2010)Jafarpour, Goyal, McLaughlin, and
  Freeman}]{Jafarpour2010}
Jafarpour B, Goyal VK, McLaughlin DB, Freeman WT (2010) {Compressed history
  matching: exploiting transform-domain sparsity for regularization of
  nonlinear dynamic data integration problems}. Mathematical Geosciences
  42(1):1--27

\bibitem[{Johnson(2015)}]{Johnson2015}
Johnson J (2015) Neural-style. \url{https://github.com/jcjohnson/neural-style}

\bibitem[{Johnson et~al.(2016)Johnson, Alahi, and Li}]{Johnson2016}
Johnson J, Alahi A, Li FF (2016) {Perceptual losses for real-time style
  transfer and super-resolution}. European Conference on Computer Vision pp
  694--711

\bibitem[{Kadian(2018)}]{Kadian2018}
Kadian A (2018) Pytorch implementation of an algorithm for artistic style
  transfer. \url{https://github.com/abhiskk/fast-neural-style/commits/master}

\bibitem[{Kingma and Ba(2014)}]{Kingma2014}
Kingma DP, Ba J (2014) {Adam: A method for stochastic optimization}. arXiv
  preprint arXiv:14126980 \eprint{1412.6980}

\bibitem[{Kitanidis(1986)}]{Kitanidis1986}
Kitanidis PK (1986) {Parameter uncertainty in estimation of spatial functions:
  Bayesian analysis}. Water Resources Research 22(4):499--507

\bibitem[{Laloy et~al.(2017{\natexlab{a}})Laloy, H{\'{e}}rault, Jacques, and
  Linde}]{Laloy2017b}
Laloy E, H{\'{e}}rault R, Jacques D, Linde N (2017{\natexlab{a}}) {Efficient
  training-image based geostatistical simulation and inversion using a spatial
  generative adversarial neural network}. arXiv preprint arXiv:170804975
  \eprint{1708.04975}

\bibitem[{Laloy et~al.(2017{\natexlab{b}})Laloy, H{\'{e}}rault, Lee, Jacques,
  and Linde}]{Laloy2017}
Laloy E, H{\'{e}}rault R, Lee J, Jacques D, Linde N (2017{\natexlab{b}})
  {Inversion using a new low-dimensional representation of complex binary
  geological media based on a deep neural network}. Advances in Water Resources
  110:387--405

\bibitem[{Liu(2017)}]{Liu2017}
Liu Y (2017) {Multilevel strategy for O-PCA-based history matching using mesh
  adaptive direct search}. Master's thesis, Stanford University

\bibitem[{Lu and Horne(2000)}]{LuPengbo2013}
Lu P, Horne RN (2000) {A multiresolution approach to reservoir parameter
  estimation using wavelet analysis. Paper SPE-62985-MS, presented at the SPE
  Annual Technical Conference and Exhibition, Dallas, Texas, 1-4 October.}

\bibitem[{Mosser et~al.(2017)Mosser, Dubrule, and Blunt}]{Mosser2017}
Mosser L, Dubrule O, Blunt MJ (2017) {Reconstruction of three-dimensional
  porous media using generative adversarial neural networks}. Physical Review E
  96(4):043,309

\bibitem[{Oliver(1996)}]{Oliver1996}
Oliver DS (1996) {Multiple realizations of the permeability field from well
  test data}. SPE Journal 1(2):145--154

\bibitem[{Paszke et~al.(2017)Paszke, Gross, Chintala, Chanan, and
  Yang}]{Paszke2017}
Paszke A, Gross S, Chintala S, Chanan G, Yang E (2017) {Automatic
  differentiation in PyTorch}. NIPS 2017 workshop

\bibitem[{Ping and Zhang(2013)}]{PingJing2013}
Ping J, Zhang D (2013) {History matching of fracture distributions by ensemble
  Kalman filter combined with vector based level set parameterization}. Journal
  of Petroleum Science and Engineering 108:288--303

\bibitem[{Remy et~al.(2009)Remy, Boucher, and Wu}]{Remy2009}
Remy N, Boucher A, Wu J (2009) {Applied Geostatistics with SGeMS: A User's
  Guide}. Cambridge University Press, Cambridge, UK

\bibitem[{Reynolds et~al.(1996)Reynolds, He, Chu, and Oliver}]{Reynolds1996}
Reynolds AC, He N, Chu L, Oliver DS (1996) {Reparameterization techniques for
  generating reservoir descriptions conditioned to variograms and well-test
  pressure data}. SPE Journal 1(4):413--426

\bibitem[{Rwechungura et~al.(2011)Rwechungura, Dadashpour, and
  Kleppe}]{Rwechungura2011}
Rwechungura R, Dadashpour M, Kleppe J (2011) {Application of particle swarm
  optimization for parameter estimation integrating production and time lapse
  seismic data. Paper SPE-146199-MS, presented at the SPE Offshore Europe
  Conference and Exhibition, Aberdeen, UK, 6-8 September.}

\bibitem[{Sarma et~al.(2006)Sarma, Durlofsky, Aziz, and Chen}]{Sarma2006}
Sarma P, Durlofsky LJ, Aziz K, Chen WH (2006) {Efficient real-time reservoir
  management using adjoint-based optimal control and model updating}.
  Computational Geosciences 10(1):3--36

\bibitem[{Sarma et~al.(2008)Sarma, Durlofsky, and Aziz}]{Sarma2008}
Sarma P, Durlofsky LJ, Aziz K (2008) {Kernel principal component analysis for
  efficient, differentiable parameterization of multipoint geostatistics}.
  Mathematical Geosciences 40(1):3--32

\bibitem[{Simonyan and Zisserman(2015)}]{Simonyan2015a}
Simonyan K, Zisserman A (2015) {Very deep convolutional networks for
  large-scale image recognition}. arXiv preprint arXiv:14091556 pp 1--14,
  \eprint{1409.1556}

\bibitem[{Strebelle(2002)}]{Strebelle2002a}
Strebelle S (2002) {Conditional simulation of complex geological structures
  using multiple-point statistics}. Mathematical Geology 34(1):1--21

\bibitem[{Tavakoli and Reynolds(2010)}]{Tavakoli2010}
Tavakoli R, Reynolds AC (2010) {History matching with parametrization based on
  the SVD of a dimensionless sensitivity matrix}. SPE Journal 15(02):495--508

\bibitem[{Vo and Durlofsky(2014)}]{Vo2014}
Vo HX, Durlofsky LJ (2014) {A new differentiable parameterization based on
  principal component analysis for the low-dimensional representation of
  complex geological models}. Mathematical Geosciences 46(7):775--813

\bibitem[{Vo and Durlofsky(2015)}]{Vo2015}
Vo HX, Durlofsky LJ (2015) {Data assimilation and uncertainty assessment for
  complex geological models using a new PCA-based parameterization}.
  Computational Geosciences 19(4):747--767

\bibitem[{Vo and Durlofsky(2016)}]{Vo2016}
Vo HX, Durlofsky LJ (2016) {Regularized kernel PCA for the efficient
  parameterization of complex geological models}. Journal of Computational
  Physics 322:859--881

\bibitem[{Zhang et~al.(2006)Zhang, Switzer, and Journel}]{Zhang2006}
Zhang T, Switzer P, Journel A (2006) {Filter-based classification of training
  image patterns for spatial simulation}. Mathematical Geology 38(1):63--80

\end{thebibliography}

\end{document}